%% file: main.tex
\documentclass[11pt]{article}

\input{acl/_preamble_default}

\input{acl/_preamble_custom}

\input{acl/_meta_info}

\begin{document}

\maketitle

\begin{abstract}
\input{sections/_abstract}
\end{abstract}

\input{sections/1_intro}
\input{sections/2_background}
\input{sections/3_preliminaries}
\input{sections/4_mechanistic_analysis}
\input{sections/5_conclusion}
\input{sections/6_limitations}
\input{sections/7_acknowledgements}

\bibliography{my_library}

\input{sections/8_appendix}

\end{document}

%% file: acl/_preamble_default.tex
\usepackage[final]{acl}

\usepackage{times}
\usepackage{latexsym}

\usepackage[T1]{fontenc}

\usepackage{microtype}

\usepackage{inconsolata}

\usepackage{graphicx}

\usepackage{fontawesome5}

%% file: acl/_preamble_custom.tex
\usepackage{booktabs}
\usepackage{multirow}

\newcommand{\ma}{MAs}
\newcommand{\as}{AS}
\newcommand{\asma}{\as{} and \ma{}}
\newcommand{\sinkRate}[2][\epsilon]{$\text{Sink}^{#1}_{#2}$}

\usepackage{subcaption}
\usepackage{amsmath}
\usepackage{amssymb}
\usepackage{bm}
\usepackage{cleveref}
\usepackage{xurl}

%% file: acl/_meta_info.tex
\title{It's Not RoPE that Creates Sinks:\\The Role of Self-Concentration and Value-Non-Mixing in Attention}

\author{
  \textbf{Raito Kiya\textsuperscript{1}}\quad
  \textbf{Satoki Ohashi\textsuperscript{1}}\quad
  \textbf{Kosuke Sato\textsuperscript{1}}\quad
  \textbf{Go Kamoda\textsuperscript{2,3}}\quad
  \textbf{Ryosuke Takahashi\textsuperscript{1,4,5}}\quad
\\
  \textbf{Yuji Yamamoto\textsuperscript{2,3}}\quad
  \textbf{Daiki Shiono\textsuperscript{1}}\quad
  \textbf{Keisuke Sakaguchi\textsuperscript{1,5}}\quad
  \textbf{Goro Kobayashi\textsuperscript{6,1}}
\\
\\
  \textsuperscript{1}Tohoku University\quad
  \textsuperscript{2}SOKENDAI\quad
  \textsuperscript{3}NINJAL\quad
    \textsuperscript{4}MBZUAI\quad
  \textsuperscript{5}RIKEN\quad
  \textsuperscript{6}Preferred Networks, Inc.
\\
  \small{
    \texttt{raito.kiya@dc.tohoku.ac.jp}
  }
}

%% file: sections/_abstract.tex
Large Language Models (LLMs) often exhibit ``Attention Sink'' (\as{}) and the accompanying ``Massive Activations'' (\ma{}) at the initial position of a sequence.
These phenomena frequently co-occur, and \ma{} can pose challenges for low-bit quantization.
In this study, we analyze the factors underlying \asma{} that emerge at the initial position regardless of the token occupying it. 
Our experiments suggest that self-concentration of attention, resulting from the causal mask, and the subsequent Value-non-mixing in attention outputs contribute to \asma{}.
These findings provide new empirical evidence on the internal dynamics of LLMs, offering insights that may inform future quantization strategies and advance our understanding of the internal mechanisms of attention layers.\footnote{\faGithub\ \href{https://github.com/kiya-raito/value-non-mixing}{github.com/kiya-raito/value-non-mixing}}

%% file: sections/1_intro.tex
\section{Introduction}
\label{sec:introduction}
\begin{figure}[t]
    \centering
    \includegraphics[width=\linewidth]{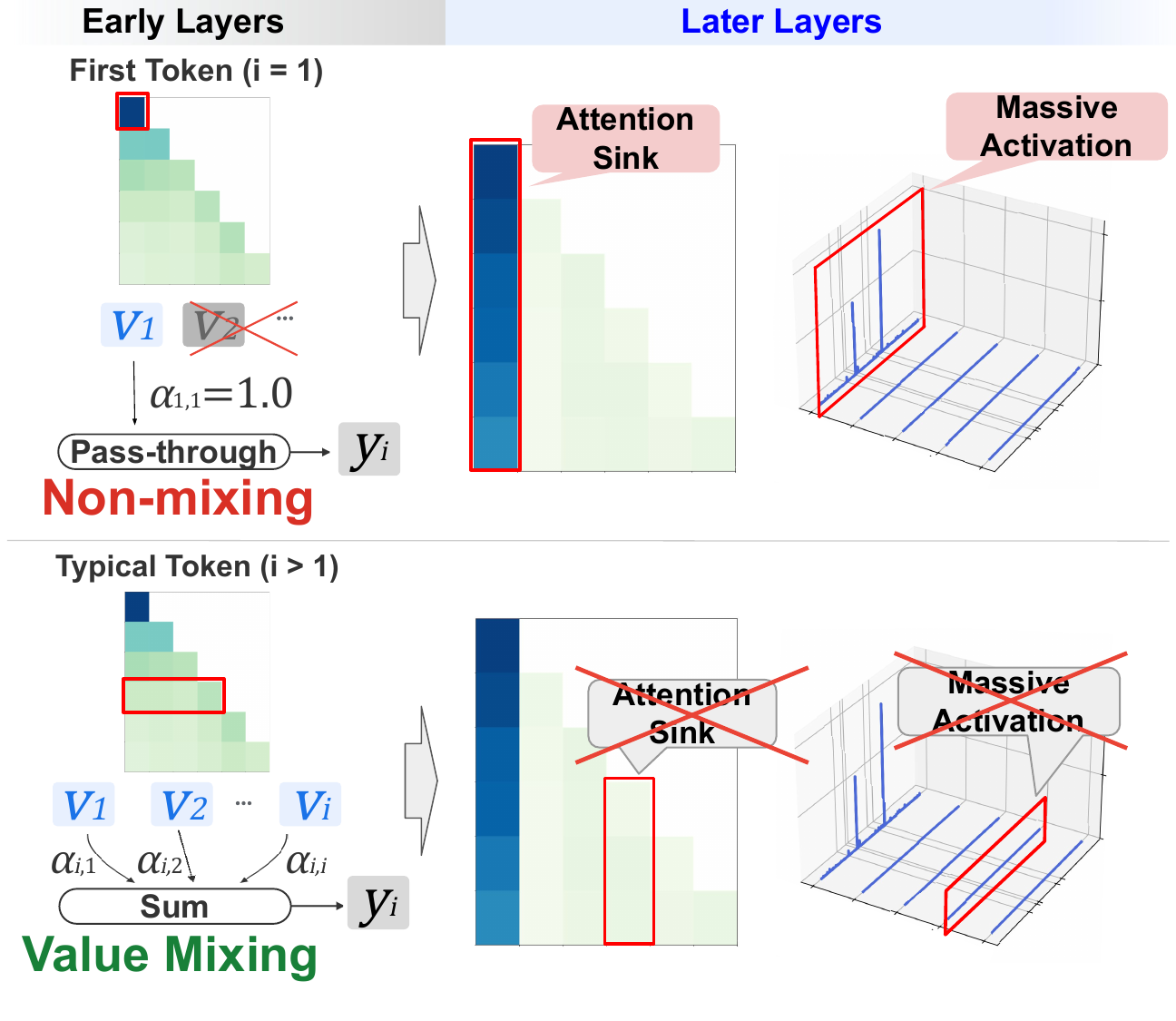}
    \caption{
        Main finding of this study. 
        While attention outputs for standard tokens ($i > 1$) are computed based on weighted averages (i.e., ``mixing'') of Value vectors, the output for the initial token is based solely on its own ``unmixed'' Value vector.
        This ``Value-non-mixing'' state in early layers is a contributing factor in \asma{} at the initial position of a sequence, regardless of the token occupying it.
    }
    \label{figure1}
\end{figure}

Large Language Models (LLMs) exhibit a phenomenon known as the ``Attention Sink'' (\as{}), where attention concentrates on specific tokens, particularly the first token of a sequence~\cite{xiaoEfficientStreamingLanguage2024, Gu-2025-whenAttentionSinkEmergesLanguageModelsEmpiricalView-qn}.
These same tokens also often exhibit a few extremely large hidden-state activation values in intermediate layers, termed ``Massive Activations'' (\ma{})~\cite{Sun-2024-massiveActivationsLargeLanguageModels-ii}.
Large-magnitude activations of this kind are known to pose a major challenge for low-precision quantization~\cite{dettmersLLMint88bitMatrix2022}.

Previous studies have offered various accounts of these phenomena, including their role in enabling no-op behavior under Softmax normalization constraints~\cite{Bondarenko-2023-quantizableTransformersRemovingOutliersHelpingAttentionHeadsNothing-eg}, the interpretation of the first token as a key bias~\cite{Gu-2025-whenAttentionSinkEmergesLanguageModelsEmpiricalView-qn}, and the influence of causal-mask asymmetry~\cite{wuEmergencePositionBias2025}.\footnote{Refer to \Cref{app:related-work} for more related works.}
Despite these different perspectives, how the properties specific to the initial position contribute to the emergence of \asma{} remains unclear.

To address this question, we decompose the properties of the initial position into three components: the typical presence of the BOS token, the first positional encoding, and forced self-concentration under causal masking.
We then disentangle their respective influences on \asma{} at the initial position of a sequence.
Our experiments provide evidence that self-concentration contributes to the emergence of these phenomena.
Under self-concentration, the attention output carries a single Value vector rather than a mixture of distinct ones, a state we call ``Value-non-mixing.''
We argue that this state contributes to \asma{} at the initial position regardless of the token occupying it (\Cref{figure1}).
Our findings provide new empirical evidence on how \asma{} form in LLMs, which may inform future quantization strategies and advance our understanding of the internal mechanisms of attention layers.

%% file: sections/2_background.tex
\section{Background: Attention Mechanism and Attention Sink Metric}
\label{sec:preliminaries}

An attention mechanism computes the output $\bm{y}_i \in \mathbb{R}^{d}$ for the $i$-th token from an input sequence $\{\bm{x}_j\}_{j=1}^T \in \mathbb{R}^{T \times d}$ as follows:
\begin{gather}
    \{\bm{q}_i,\bm{k}_i,\bm{v}_i\} = \bm{x}_i \{ \bm{W}_Q\bm{R}_{\Theta,i}^\top, \bm{W}_K \bm{R}_{\Theta,i}^\top, \bm{W}_V \} \nonumber \\
    \alpha_{i,j} = \frac{\exp\left(\bm{q}_i \bm{k}_j^\top / \sqrt{d'}\right)}{\sum_{m=1}^i \exp\left(\bm{q}_i \bm{k}_m^\top / \sqrt{d'}\right)} \nonumber \\
    \bm{y}_i = \sum_{j=1}^i \alpha_{i,j} \bm{v}_j \bm{W}_O\label{eq:attention}\text{,}
\end{gather}
where $\bm{W}_{Q,K,V} \in \mathbb{R}^{d \times d'}$ and $\bm{W}_O \in \mathbb{R}^{d' \times d}$ denote projection matrices for Query/Key/Value/Output, and $\bm{R}_{\Theta,i}$ denotes the RoPE \cite{rope2024} rotation matrix.

\as{} refers to the phenomenon where the token at position $j$ (often the first) receives disproportionately large attention weights $\alpha_{i,j}$ across many heads~\cite{xiaoEfficientStreamingLanguage2024}.
Following \citet{Gu-2025-whenAttentionSinkEmergesLanguageModelsEmpiricalView-qn}, we quantify the strength of \as{} at position $j$ using the \sinkRate{j} metric:
\begin{align}
    \bar{\alpha}_j^{l,h} &= \frac{1}{T-j+1} \sum_{i=j}^T \alpha_{i,j}^{l,h} \nonumber \\
    \text{Sink}^\epsilon_j &= \frac{1}{L \cdot H} \sum_{l=1}^L \sum_{h=1}^H \mathbb{I}(\bar{\alpha}_j^{l,h} > \epsilon)
    \label{eq:sink_definition}
    \text{,}
\end{align}
where $L$ is the number of layers, $H$ is the number of heads at each layer, $\epsilon$ is a threshold\footnote{
    Following \citet{Gu-2025-whenAttentionSinkEmergesLanguageModelsEmpiricalView-qn}, we use $\epsilon=0.3$.
    See \Cref{appendix:varying-epsilon-thresholds} for results confirming that our findings hold for $\epsilon \in \{0.2, 0.3, 0.4, 0.5\}$.
}, and $\mathbb{I}(\cdot)$ is the indicator function.

%% file: sections/3_preliminaries.tex
\section{Preliminaries: Influence of BOS Token}
\label{sec:exp-bos}
Before our main analysis, we examine whether \asma{} observed at the initial position of a sequence are attributable to the BOS embedding or to the positional characteristics of the initial position.
To disentangle these two factors, we swap the BOS token with the token at a non-initial position ($t=16$) in each sequence.
Our experiments use WikiText \citep{wikitext}\footnote{We use the \texttt{wikitext-2-raw-v1} subset from \url{https://huggingface.co/datasets/Salesforce/wikitext}.} and, among the models listed in \Cref{tab:hf-models}, the three whose tokenizer prepends a BOS token, since this swap is undefined for the other two.

\begin{figure}[t]
     \centering
     \includegraphics[width=0.42\linewidth]{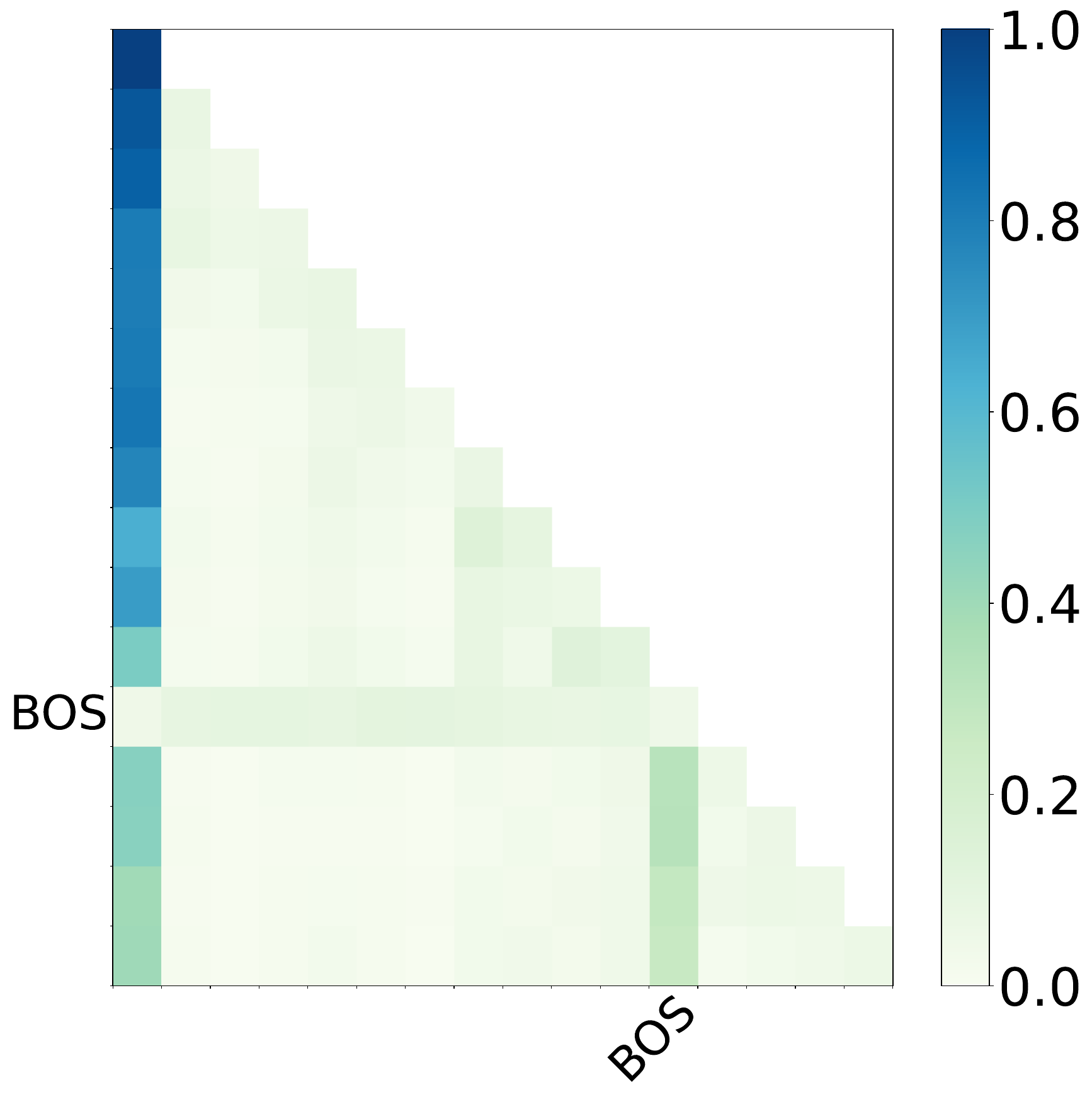}
     \includegraphics[width=0.48\linewidth]{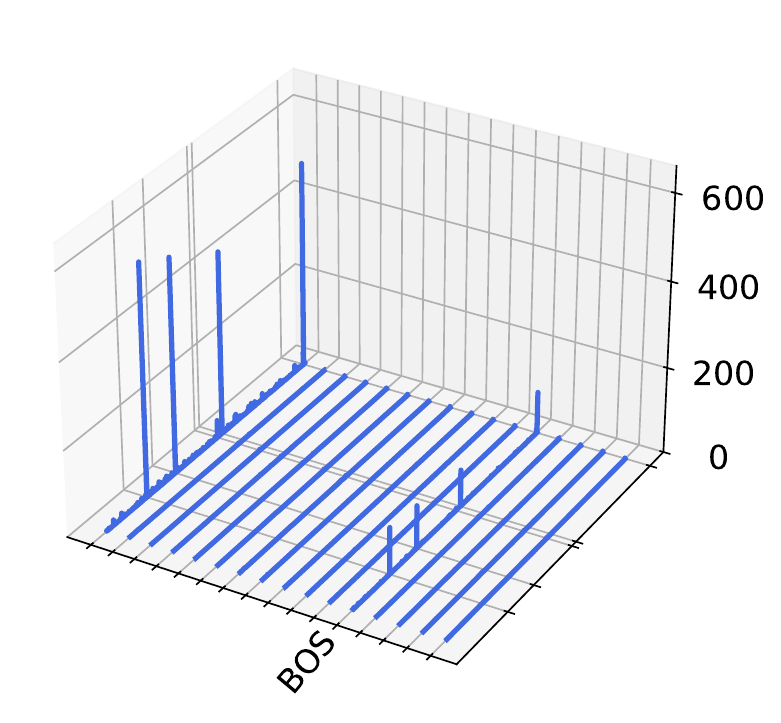}
     \caption{Attention heatmap (left) and activation magnitudes (right) from Layer 14 of Llama-3.2-3B with the BOS token swapped with the token at the 12th position of a 16-token sequence (shorter than the setting in \Cref{tab:move_bos} for readability).
     Both \asma{} emerge at the initial position, independent of the BOS token location.
     }
     \label{fig:move_bos}
 \end{figure}

\Cref{tab:move_bos} reports the mean \sinkRate{j} under this setting.
The results show that \as{} still occurs at the BOS token even when the token is relocated to the middle of each sequence.
Moreover, although its strength varies across models, \as{} consistently emerges at the initial position of the sequence even when the initial token is not the BOS token.
We observe a similar trend for \ma{}, as shown in \Cref{fig:move_bos}.
Taken together, these results indicate that both the BOS token and the initial position contribute to \asma{}, with the latter operating independently of token type.
These findings are consistent with prior reports that \ma{} occur at the initial position of a sequence~\cite{Sun-2024-massiveActivationsLargeLanguageModels-ii} and that the BOS token triggers \asma{}~\cite{oh2025House_cards_Massive_Weights}. 
Our results complement these works by providing a comprehensive quantitative evaluation of these phenomena using \sinkRate{j} as a unified metric across multiple samples and models.

\input{tables/table_set5_move_bos_random_eps0.3}

%% file: tables/table_set5_move_bos_random_eps0.3.tex
\begin{table}[t]
\centering
\small
\begin{tabular}{l l c c }
\toprule
\textbf{Model} & \textbf{Position} & \textbf{Vanilla} & \textbf{BOS at 16th} \\
\midrule
\multirow{2}{*}{Llama-2-7b-hf} & \textbf{1} & 0.9321 & 0.9178 \\
 & \textbf{16} & 0.0001 & \textbf{0.7597} \\
\midrule
\multirow{2}{*}{Llama-3.2-3B} & \textbf{1} & 0.9879 & 0.8856 \\
 & \textbf{16} & 0.0000 & \textbf{0.4312} \\
\midrule
\multirow{2}{*}{Mistral-7B-v0.3} & \textbf{1} & 0.9833 & 0.1675 \\
 & \textbf{16} & 0.0000 & \textbf{0.6376} \\
\bottomrule
\end{tabular}
\caption{\sinkRate{j} when the BOS token is moved to the 16th position. Unless otherwise noted, all subsequent tables in this analysis use $T=64$ over $N=100$ WikiText sequences.}
\label{tab:move_bos}
\end{table}

%% file: sections/4_mechanistic_analysis.tex
\input{tables/table_set1_eps0.3}

\section{Mechanistic Analysis of the Initial-Position Effect on \asma{}}

In \Cref{sec:exp-bos}, we showed that \asma{} are associated not only with the BOS token but also with properties specific to the sequence-initial position.
In this section, we focus on the initial-position effect by examining two factors: positional encoding (\Cref{sec:exp-rope}) and the self-concentration of attention (\Cref{sec:exp-self100}).
We then use these observations to further investigate the underlying mechanism that gives rise to \asma{} (\Cref{sec:repeat-token}).

\input{tables/table_set3_intervention_eps0.3}

\begin{figure}[t]
  \centering
    \centering
    \includegraphics[width=\linewidth]{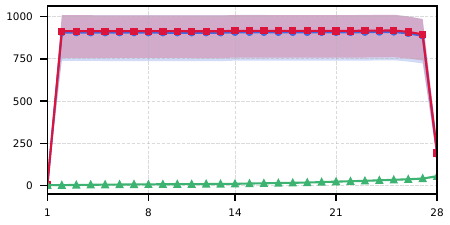}
  \caption{Layer-wise averaged hidden-state norms for Llama-3.2-3B under the self-concentration intervention (w/o BOS token).
  Blue circles, red squares, and green triangles denote the first token, the intervention token, and the mean over all other input tokens, respectively.
  The horizontal axis shows the layer index.
  Shaded bands show the sample min--max range.}
  \label{fig:activation_norms_remove_bos_self_only_all_models}
\end{figure}

\subsection{Influence of Positional Encoding}
\label{sec:exp-rope}

We investigate whether the positional information encoded by Rotary Position Embedding \citep[RoPE;][]{rope2024} contributes to the emergence of \as{} at the initial position.
\footnote{We focus on RoPE as it is adopted by most recent open-weight LLMs.
Our models also differ in their RoPE configuration, such as the fraction of head dimensions to which the rotation is applied (\Cref{appendix:models}).
Other positional-encoding schemes are outside our scope (see the Limitations section).}
To this end, we intervene on the RoPE index of the Key vector for the initial token.
Specifically, following \Cref{eq:attention}, we replace the Key-side
rotation matrix $\bm{R}_{\Theta,1}$ with $\bm{R}_{\Theta,r}$, where
$r$ is sampled uniformly from $\{2,\ldots,T\}$, which excludes the initial token's original RoPE index.
This yields the following intervened Key vector:
\begin{equation}
\bm{k}_1^{\mathrm{interv}} = \bm{x}_1 \bm{W}_K \bm{R}_{\Theta,r}^{\top} \;\; \text{.}
\end{equation}
If RoPE were the primary driver of \as{}, this intervention would be expected to reduce \sinkRate{1} substantially.
To isolate the effect of the initial position from the BOS token, we conduct this experiment using sequences without a BOS token.

As shown in \Cref{tab:rope_ablation_sink_metrics_epsilon_0.3_only}, the RoPE intervention does not substantially reduce \sinkRate{1} in any of the five models.
This result suggests that the initial token's Key-side RoPE encoding has only a limited effect on \as{} under this intervention.

\begin{figure*}[t!]
  \centering
  \begin{subfigure}[t]{0.22\linewidth}
    \centering
    \includegraphics[width=\linewidth]{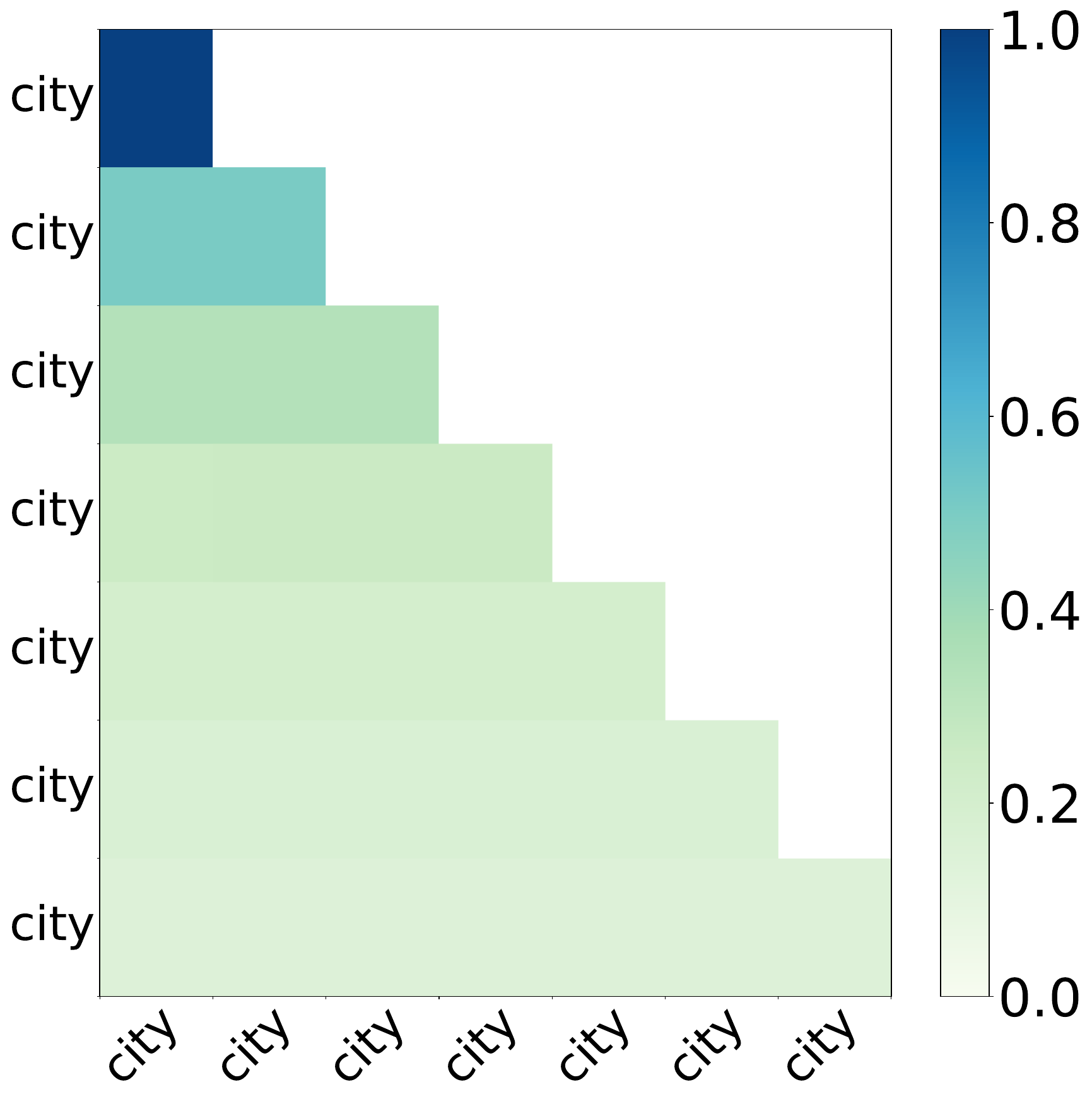}
    \caption{Heatmap: Uniform}
    \label{fig:heatmap_all_same_llama3}
  \end{subfigure}
  \hfill
  \begin{subfigure}[t]{0.22\linewidth}
    \centering
    \includegraphics[width=\linewidth]{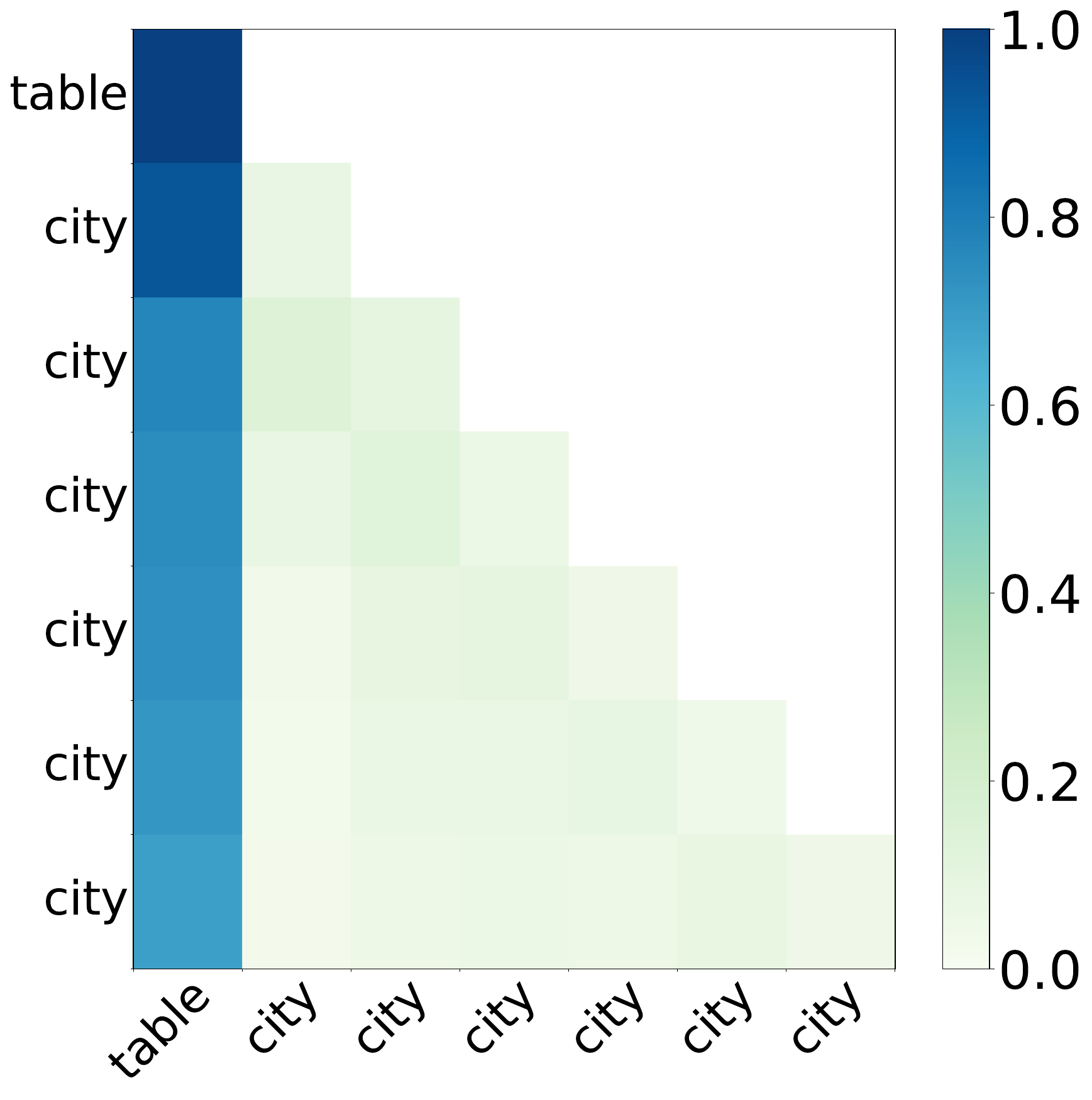}
    \caption{Heatmap: Pos. 1}
    \label{fig:heatmap_diff_first_llama3}
  \end{subfigure}
  \hfill
  \begin{subfigure}[t]{0.22\linewidth}
    \centering
    \includegraphics[width=\linewidth]{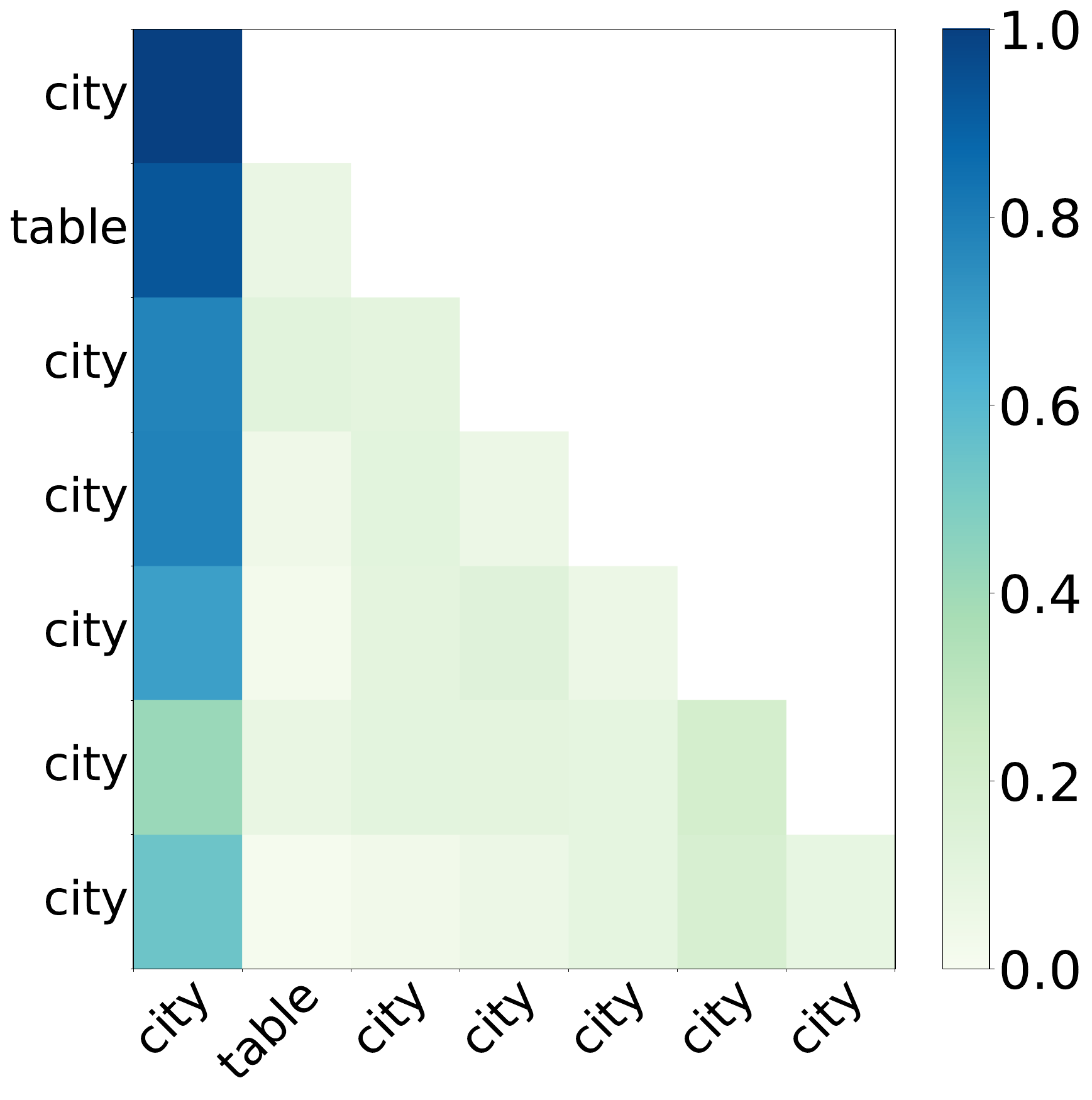}
    \caption{Heatmap: Pos. 2}
    \label{fig:heatmap_diff_second_llama3}
  \end{subfigure}
  \hfill
  \begin{subfigure}[t]{0.22\linewidth}
    \centering
    \includegraphics[width=\linewidth]{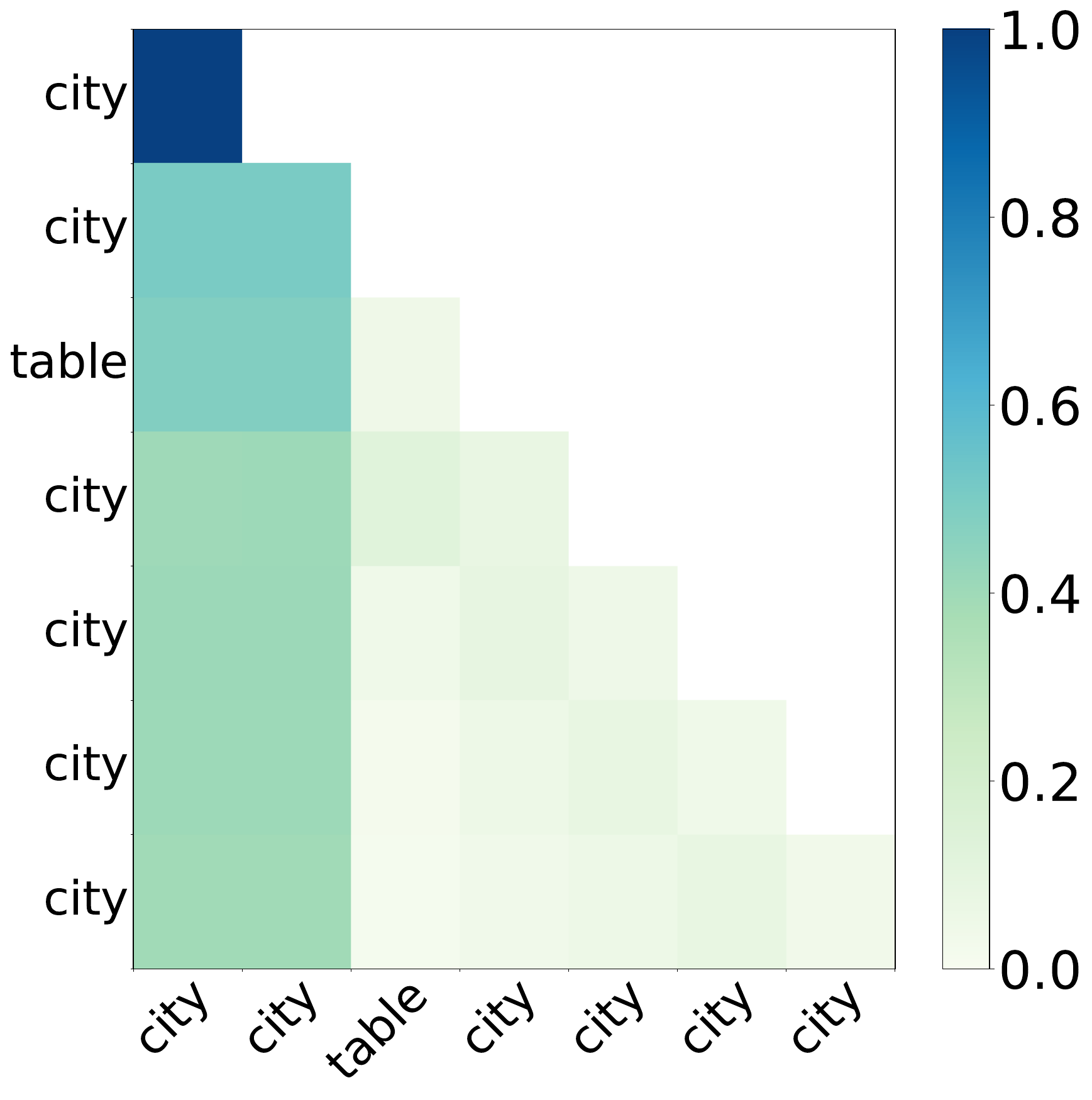}
    \caption{Heatmap: Pos. 3}
    \label{fig:heatmap_diff_third_llama3}
  \end{subfigure}

  \begin{subfigure}[t]{0.24\linewidth}
    \centering
    \includegraphics[width=\linewidth]{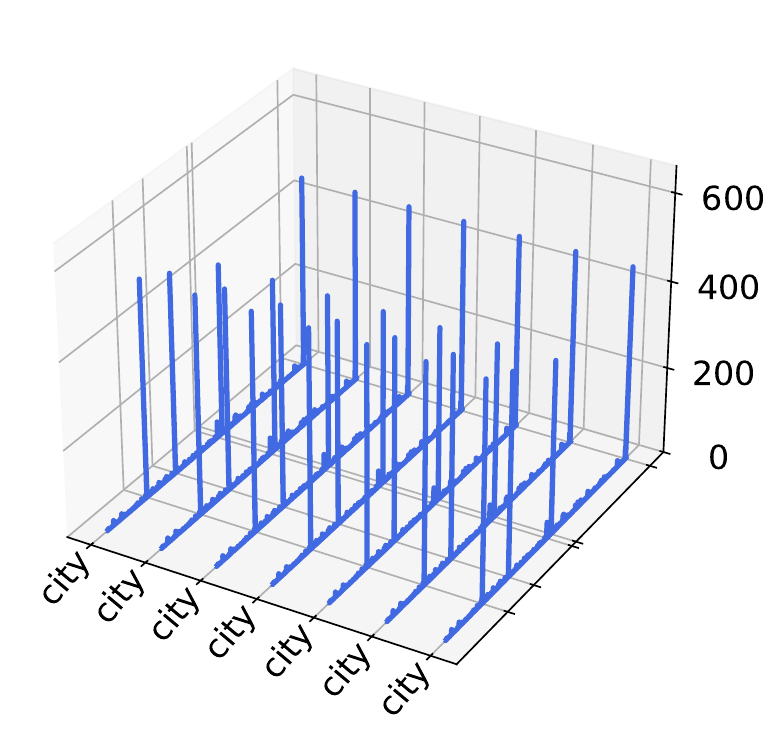}
    \caption{Activation: Uniform}
    \label{fig:act_all_same_llama3}
  \end{subfigure}
  \hfill
  \begin{subfigure}[t]{0.24\linewidth}
    \centering
    \includegraphics[width=\linewidth]{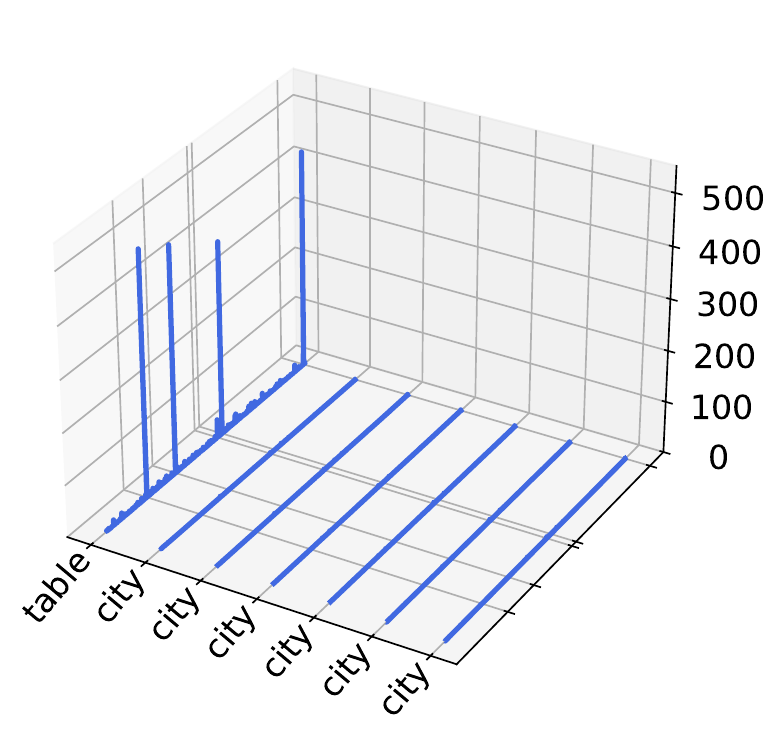}
    \caption{Activation: Pos. 1}
    \label{fig:act_diff_first_llama3}
  \end{subfigure}
  \hfill
  \begin{subfigure}[t]{0.24\linewidth}
    \centering
    \includegraphics[width=\linewidth]{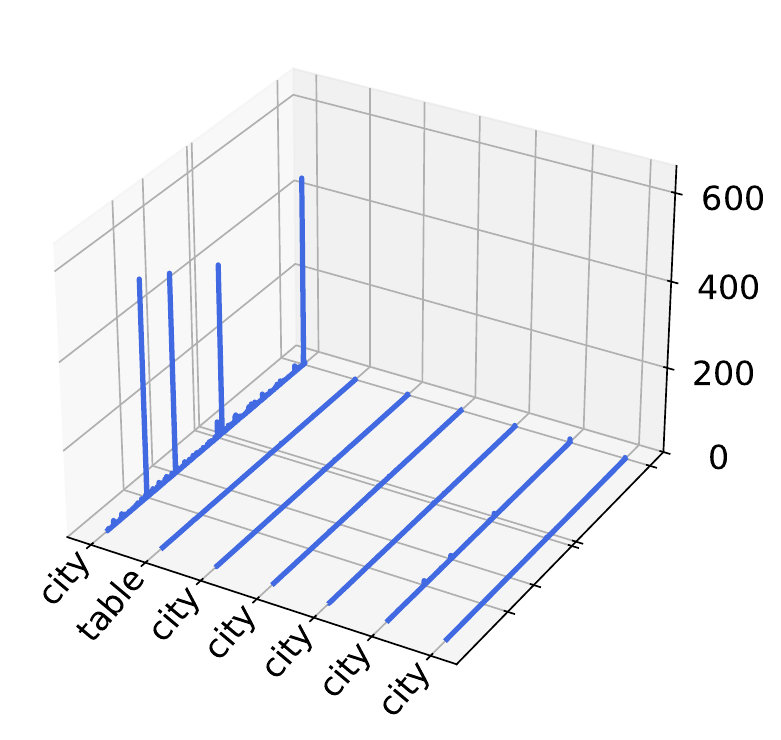}
    \caption{Activation: Pos. 2}
    \label{fig:act_diff_second_llama3}
  \end{subfigure}
  \hfill
  \begin{subfigure}[t]{0.24\linewidth}
    \centering
    \includegraphics[width=\linewidth]{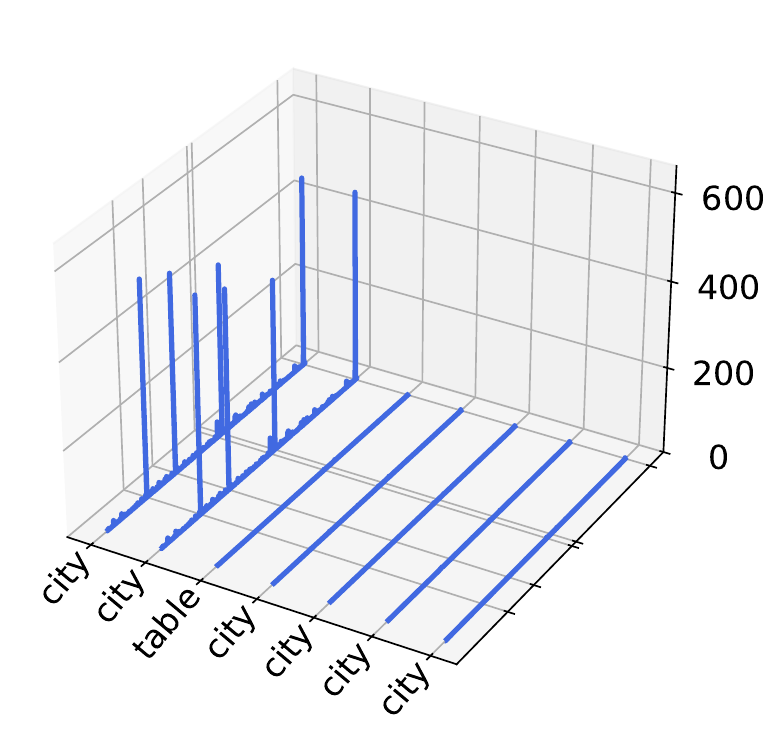}
    \caption{Activation: Pos. 3}
    \label{fig:act_diff_third_llama3}
  \end{subfigure}

  \caption{
    Attention heatmap (top row) and hidden-state activations (bottom row) at Layer 14 of Llama-3.2-3B for repeated-token sequences with varying distinct token positions.
}
  \label{fig:repeated_tokens_results_llama3}
\end{figure*}

\begin{figure*}[t!]
  \centering
  \begin{subfigure}[t]{0.24\linewidth}
    \centering
    \includegraphics[width=\linewidth]{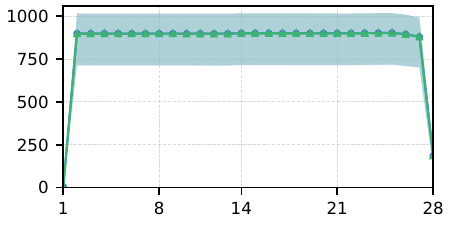}
    \caption{Norms: Uniform}
    \label{fig:norm_all_same_llama3}
  \end{subfigure}
  \hfill
  \begin{subfigure}[t]{0.24\linewidth}
    \centering
    \includegraphics[width=\linewidth]{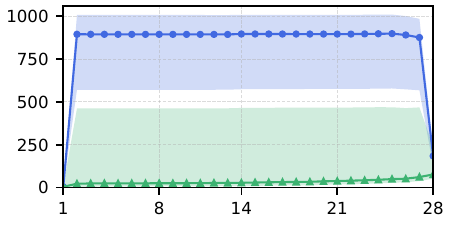}
    \caption{Norms: Pos. 1}
    \label{fig:norm_diff_first_llama3}
  \end{subfigure}
  \hfill
  \begin{subfigure}[t]{0.24\linewidth}
    \centering
    \includegraphics[width=\linewidth]{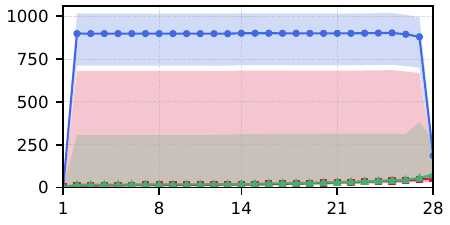}
    \caption{Norms: Pos. 2}
    \label{fig:norm_diff_second_llama3}
  \end{subfigure}
  \hfill
  \begin{subfigure}[t]{0.24\linewidth}
    \centering
    \includegraphics[width=\linewidth]{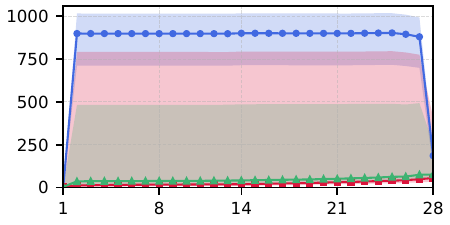}
    \caption{Norms: Pos. 3}
    \label{fig:norm_diff_third_llama3}
  \end{subfigure}
  \caption{Layer-wise averaged hidden-state norms in Llama-3.2-3B for repeated-token sequences ($T=64$, without a BOS token and without any attention intervention), averaged over $N=100$ sequences per panel.
  Panels correspond to the four input patterns: a uniform repeated-token sequence (Uniform) and sequences with a single distinct token inserted at the first, second, or third position (Pos.\ 1--3).
  Blue circles and green triangles denote the first token and the mean over all remaining input tokens, respectively.
  Red squares denote the distinct token, and therefore appear only in the Pos.\ 2 and Pos.\ 3 panels: the Uniform panel contains no distinct token, and in Pos.\ 1 the distinct token is the first token itself.
  The horizontal axis shows the layer index in every panel.
  Shaded bands show the sample min--max range.}
  \label{fig:activation_norms_repeated_tokens_llama3}
\end{figure*}

\subsection{Influence of Self-Concentration of Attention}
\label{sec:exp-self100}

Given the limited impact of positional encoding, we then focus on the property that the initial token is forced to direct all attention to itself ($\alpha_{1,1}=1.0$).
We refer to this property as ``self-concentration.''
To investigate whether this property contributes to \asma{}, we intervene on the attention weights across all layers at a non-initial position ($t=16$) to force self-concentration ($\alpha_{t,t} = 1.0, \alpha_{t,j \neq t}=0.0$).
We conduct this experiment using sequences without a BOS token as well.

\Cref{tab:self_const_interv_sink_metrics_epsilon_0.3} reports \sinkRate{t} before and after forcing self-concentration at $t=16$ across five models, and \Cref{fig:activation_norms_remove_bos_self_only_all_models} presents the corresponding layer-wise hidden-state norms for Llama-3.2-3B.
These results show that forcing self-concentration induces both \as{} and \ma{} at non-initial positions in all five models,\footnote{
    \Cref{appendix:varying-intervened-position} shows similar trends at $t=8$ and $t=32$.
    Hidden-state norms for the other four models are shown in \Cref{fig:activation_norms_remove_bos_self_only_other_models}.
} although both effects are weaker for Mistral-7B-v0.3.
This suggests that self-concentration is one factor that can trigger the emergence of \asma{}.

\subsection{Value-non-mixing as a Driver}
\label{sec:repeat-token}

\paragraph{Hypothesis.}
We hypothesize that the emergence of \asma{} under self-concentration is driven by the resulting ``Value-non-mixing'' state, rather than by self-concentration itself.
As shown in \Cref{eq:attention}, an attention mechanism computes each output $\bm y_i$ as a weighted average of the Value vectors corresponding to context tokens.
Under self-concentration, however, the output consists solely of the token's own Value vector, without being mixed with those of other tokens (\Cref{figure1}).
We formalize this Value-non-mixing state at position $i$ as
\begin{equation}
    {\bm v_j \mid j \le i, ; \alpha_{i,j} > 0} = {\bm v_i}
    \label{eq:value_non_mixing}
    \text{,}
\end{equation}
meaning that all Value vectors receiving nonzero attention from the query at position $i$ are identical to $\bm v_i$.
Because the attention weights sum to one, this implies $\bm y_i = \bm v_i \bm{W}_O$.

\paragraph{Experimental Design.}
To test whether Value-non-mixing can induce \asma{}, we reproduce this state without forcing self-concentration by using sequences composed of repeated identical tokens without a BOS token.
Because all tokens share the same Value vector, the attention output remains a pass-through of that single Value vector even without self-concentration.
By inserting a distinct token into the sequence, we can control the range of positions over which the Value-non-mixing state is maintained.
We evaluate four conditions: a uniform repeated-token sequence and sequences with a distinct token inserted at the first, second, or third position.
For each of the five models, we generate $N=100$ sequences per condition, each of length $T=64$, randomly sampling the repeated token and, where applicable, the distinct token for each sequence.
Although repeated-token inputs deviate from natural language, this manipulation complements the self-concentration intervention in \Cref{sec:exp-self100}, which induces the same state on natural text. 

\paragraph{Results.}

\Cref{fig:repeated_tokens_results_llama3} shows the results for Llama-3.2-3B.
When the sequence consists entirely of identical tokens,
\ma{} are observed across all positions (\Cref{fig:act_all_same_llama3}), even though the attention weights are distributed across positions (\Cref{fig:heatmap_all_same_llama3}).
For sequences containing a distinct token, \Crefrange{fig:heatmap_diff_first_llama3}{fig:heatmap_diff_third_llama3} and \Crefrange{fig:act_diff_first_llama3}{fig:act_diff_third_llama3} show that \asma{} occur at the initial token or at the identical-token positions preceding the distinct token, but subside at subsequent positions.
This positional pattern is also visible in the layer-wise hidden-state norms shown in \Cref{fig:activation_norms_repeated_tokens_llama3}.
These observations for Llama-3.2-3B indicate that \asma{} emerge at positions that remain in the Value-non-mixing state, supporting our hypothesis.
See Appendices~\ref{appendix:repeated-token-across-model} and~\ref{appendix:repeat-length-varied} for results across all models, quantitative analyses, and additional experiments under varied settings.

For three of the five models examined, including Llama-3.2-3B, inserting a distinct token consistently produces \asma{} across the sampled token combinations.
For the remaining two, Mistral-7B-v0.3 and pythia-1b, the emergence of \asma{} is more input-dependent, and the effect is also weaker on average (\Cref{appendix:repeated-token-across-model}).
What drives this input dependence---for example, whether it relates to the similarity between the Value vectors of the repeated and inserted tokens---is left for future work.

Our findings provide further insight into the mechanisms underlying \asma{}.
Recent architectures proposed by major LLM developers, including Qwen3-Next~\cite{qiu2025qwengatedattention} and gpt-oss~\cite{openai2025gptoss120bgptoss20bmodel}, highlight the growing importance of reducing these phenomena.
These findings may also be relevant to the design of quantization-robust architectures.

%% file: tables/table_set1_eps0.3.tex
\begin{table}[t]
\centering
\small
\begin{tabular}{l c c }
\toprule
\textbf{Model} & \textbf{Vanilla} & \textbf{RoPE Interv} \\
\midrule
Llama-2-7b-hf & 0.9305 & 0.9273 \\
Llama-3.2-3B & 0.9218 & 0.9133 \\
Mistral-7B-v0.3 & 0.0343 & 0.1525 \\
Qwen2-7B & 0.8173 & 0.7931 \\
pythia-1b & 0.7337 & 0.7547 \\
\bottomrule
\end{tabular}
\caption{\sinkRate{1} with and without RoPE intervention (both w/o BOS).}
\label{tab:rope_ablation_sink_metrics_epsilon_0.3_only}
\end{table}

%% file: tables/table_set3_intervention_eps0.3.tex
\begin{table}[t]
\centering
\small
\begin{tabular}{l l c c }
\toprule
\textbf{Model} & \textbf{Position} & \textbf{Vanilla} & \textbf{Interv} \\
\midrule
\multirow{2}{*}{Llama-2-7b-hf} & 1 & 0.9305 & 0.9164 \\
 & 16 & 0.0073 & \textbf{0.8397} \\
\midrule
\multirow{2}{*}{Llama-3.2-3B} & 1 & 0.9218 & 0.8785 \\
 & 16 & 0.0012 & \textbf{0.7424} \\
\midrule
\multirow{2}{*}{Mistral-7B-v0.3} & 1 & 0.0343 & 0.0198 \\
 & 16 & 0.0373 & \textbf{0.1509} \\
\midrule
\multirow{2}{*}{Qwen2-7B} & 1 & 0.8173 & 0.7278 \\
 & 16 & 0.0000 & \textbf{0.5085} \\
\midrule
\multirow{2}{*}{pythia-1b} & 1 & 0.7337 & 0.6702 \\
 & 16 & 0.0000 & \textbf{0.4703} \\
\bottomrule
\end{tabular}
\caption{Comparison of \sinkRate{j} when intervention forces self-concentration of attention at position 16 (both w/o BOS).}
\label{tab:self_const_interv_sink_metrics_epsilon_0.3}
\end{table}

%% file: sections/5_conclusion.tex
\section{Conclusion}
In this study, we investigated the mechanism underlying \asma{} in LLMs, with a particular focus on the sequence-initial position. 
At the sequence-initial position, causal masking inherently enforces self-concentration of attention ($\alpha_{1,1}=1.0$).
Our experiments show that forcing such self-concentration at a non-initial position induces \asma{} across all five models.
We further investigated the ``Value-non-mixing'' state resulting from self-concentration as a potential mechanism through which self-concentration induces \asma{}. 
Repeated-token experiments provide additional support for this hypothesis in three of the five models, while the emergence of \asma{} is more input-dependent and weaker on average in the remaining two.
Together, these findings support a mechanistic link between self-concentration, Value-non-mixing, and the emergence of \asma{}.
By clarifying this mechanistic link, our findings provide empirical evidence that may inform future quantization strategies and advance our understanding of the internal mechanisms of attention layers.
Future work should directly quantify Value-vector similarity and identify the contributing head-level pathways.

%% file: sections/6_limitations.tex
\section*{Limitations}
\label{sec:limitations}

Our study has several limitations.

First, the five models we evaluate span four families (Llama, Mistral, Qwen, and Pythia; \Cref{appendix:models}), but whether our findings extend to other architectures (e.g., Mixture-of-Experts models~\citep{shazeer2017}), languages, and domains beyond the English datasets evaluated here remains open.
Among the five models, Mistral-7B-v0.3 responds more weakly to our interventions than the other models, and pythia-1b shows the weakest and most input-dependent effect in the repeated-token experiments (\Cref{sec:repeat-token}).
Identifying the architectural differences behind this variance is left for future work.
All five models are RoPE-based, though they vary structurally within RoPE (e.g., partial RoPE in pythia-1b), and \citet{Gu-2025-whenAttentionSinkEmergesLanguageModelsEmpiricalView-qn} report that positional embeddings did not affect the emergence of \as{}.
How positional-encoding schemes influence the \asma{} mechanism is outside our scope and an important future direction.

Second, we constrain our scope to the sequence-initial position to isolate its structural drivers.
Our intervention induces \as{} at non-initial positions (\Cref{sec:exp-self100} and \Cref{appendix:varying-intervened-position}), but whether naturally occurring intermediate \as{}~\citep{Sun-2024-massiveActivationsLargeLanguageModels-ii} can be explained within the same Value-non-mixing framework remains an open question.

Third, our interventions cannot exclude the contribution of other emergent factors.
The RoPE intervention manipulates only the key-side positional index of the initial token, so isolating query-side effects and query--key interactions, and rigorously excluding RoPE's entire contribution, are left for future work.
Likewise, restricting the intervention to early layers reproduces most of the all-layer effect (\Cref{appendix:layer-restricted-intervention}), but pinpointing the head-level causal pathway is challenging and architecture-dependent, and we leave head-selective analyses for future work.

Fourth, our definition of Value-non-mixing (\Cref{eq:value_non_mixing}) characterizes a discrete state; extending it to a continuous, predictive ``Value Mixing Index'' requires modeling the geometry of Value vectors and validating that it predicts both \sinkRate{j} and \ma{} magnitude across models and natural inputs.
The threshold-based \sinkRate{j} is also sensitive to sequence length, because attention weights for a query at position $i$ average $1/i$ over its accessible keys and thus more readily exceed a fixed $\epsilon$ in short sequences (e.g., repeat length 16 at $\epsilon=0.2$ in \Cref{tab:repeat_input_sensitivity_analysis_part1}), which limits fixed-threshold comparisons and motivates length-aware metrics.

Finally, although prior work distinguishes \ma{} from the activation outlier features that complicate low-bit quantization~\citep{Sun-2024-massiveActivationsLargeLanguageModels-ii,dettmersLLMint88bitMatrix2022}, we do not evaluate whether \ma{} or Value-non-mixing affects quantization error, perplexity, or downstream accuracy; establishing such a connection requires dedicated experiments.

%% file: sections/7_acknowledgements.tex
\section*{Acknowledgements}
This work was supported by
Google Research Grant; 
JST BOOST
    JPMJBY24D2,
    JPMJBS2412,
    JPMJBS2421;
JST SPRING
    JPMJSP2104;
JSPS KAKENHI 25K03175;
and
the Nakajima Foundation.
We used generative AI tools to assist with coding and manuscript preparation.
We thank the members of the Tohoku NLP Group for their cooperation in this research.

%% file: sections/8_appendix.tex
\appendix
\setlength{\abovecaptionskip}{2pt}
\setlength{\belowcaptionskip}{0pt}
\setlength{\floatsep}{6pt plus 2pt minus 2pt}
\setlength{\textfloatsep}{6pt plus 2pt minus 2pt}
\setlength{\intextsep}{6pt plus 2pt minus 2pt}
\captionsetup[figure]{font=footnotesize}
\captionsetup[sub]{font=scriptsize, skip=1pt}

\section{Related Work}
\label{app:related-work}
\asma{} are phenomena of practical importance from an engineering perspective.
Specific tokens function as aggregation points for attention, enabling the reuse and compression of KV caches \cite{xiaoEfficientStreamingLanguage2024, xiaoDuoAttentionEfficientLongContext2025, zhangH2OHeavyHitterOracle2023}.
Furthermore, the presence of activations with large magnitudes has been widely recognized as a major challenge for effectively quantizing LLMs~\citep{dettmersLLMint88bitMatrix2022, Sun-2024-massiveActivationsLargeLanguageModels-ii}.
Note that \ma{} are distinct from the outlier features studied in the quantization literature~\citep[\S2.3]{Sun-2024-massiveActivationsLargeLanguageModels-ii}.

Existing literature analyzing \asma{} has documented their layer-wise formation profiles and functional significance.
Empirical observations indicate that these phenomena typically emerge in the initial layers, stabilize across the middle layers, and weaken near the final layers \cite{Sun-2024-massiveActivationsLargeLanguageModels-ii, Gu-2025-whenAttentionSinkEmergesLanguageModelsEmpiricalView-qn}.
To explain this vertical persistence, \citet{queipo-de-llanoAttentionSinksCompression2026} proposed the Mix-Compress-Refine Theory, arguing that \as{} in the middle layers prevents the over-mixing of contextual information.
Additionally, \citet{Gu-2025-whenAttentionSinkEmergesLanguageModelsEmpiricalView-qn} provided causal evidence that \as{} can be replaced by explicit key biases.
\citet{Barbero-2025-whyLlmsAttendFirstToken-yq} instead argue that \as{} prevents over-mixing, with the first token anchoring the residual stream against representational collapse (the Anti-Overmixing Theory).

From a mechanistic standpoint, several studies have investigated why specific tokens attract disproportionate attention.
\citet{kobayashi-etal-2020-attention} showed, via a norm-based analysis of attention outputs, that BERT's attention heads assign large weights to uninformative delimiter tokens (e.g., [SEP]) whose value vectors are nonetheless small, so that the attention output leaves the residual stream almost unchanged.
\citet{Bondarenko-2023-quantizableTransformersRemovingOutliersHelpingAttentionHeadsNothing-eg} relate this behavior to activation outliers observed more broadly in trained transformers (BERT, OPT, and ViT), framing such heads as having learned a no-op that leaves the residual stream unchanged or only partially updated.
They report an analogous pattern for uninformative tokens beyond BERT (e.g., background patches in ViT).
They further argue that a strict no-update requires near-exact zeros in the attention matrix, and that producing them drives the softmax inputs to ever larger values during training, which in turn creates outliers elsewhere in the network.
Architectural factors have also been related to position bias more broadly.
\citet{wuEmergencePositionBias2025} analyze multi-layer attention with a graph-theoretic framework and show that causal masking biases attention toward earlier positions, since tokens in deeper layers attend to increasingly contextualized representations of earlier tokens.
They treat \as{} as one manifestation of the position bias covered by this framework.
A complementary account attributes part of this bias to the training data.
\citet{salvatore2025lostmiddleemergentproperty} train GPT-2 and Llama variants from scratch on tasks that simulate long-term and short-term memory demands, and report that the primacy effect is induced by the uniform long-term demand and is further influenced by the autoregressive architecture and the formation of \as{}.

Our study differs from these accounts in its research question and direction of causality.
Prior work primarily investigates the effects, functional roles, or alternative formulations of existing \as{} (e.g., anti-overmixing, key bias), whereas we investigate how \asma{} emerge at the sequence-initial position.
The notion of ``mixing'' also differs in meaning.
For \citet{Barbero-2025-whyLlmsAttendFirstToken-yq}, mixing refers to the propagation of contextual information across positions and layers, whereas our ``Value-non-mixing'' concerns the composition of Value vectors within a single attention output (\Cref{eq:value_non_mixing}).
The two accounts are therefore complementary.
Our contribution is to control the Value-non-mixing state artificially, using attention interventions and repeated-token inputs into which a distinct token is inserted, and to show that \asma{} emerge largely within the range of positions over which this state holds in the models we evaluate.

\section{Experimental Setup}
\label{appendix:models}
To evaluate the generality of our findings regarding the mechanism of \asma{}, we conduct our main experiments on the WikiText dataset \citep{wikitext}\footnote{We use the \texttt{wikitext-2-raw-v1} subset from \url{https://huggingface.co/datasets/Salesforce/wikitext}.}.
For the robustness analyses in \Cref{appendix:varied-configurations}, we additionally use the GSM8K \citep{cobbe2021gsm8k} and SlimPajama\footnote{We use \texttt{DKYoon/SlimPajama-6B} from \url{https://huggingface.co/datasets/DKYoon/SlimPajama-6B}.} datasets.
We evaluate our hypothesis across a diverse set of prominent open-weight LLMs, including Llama-3.2-3B \cite{Grattafiori-2024-llama3HerdModels-vs}, Llama-2-7b-hf \cite{touvron2023llama2openfoundation}, Mistral-7B-v0.3 \cite{jiang2023mistral7b}, pythia-1b \cite{Biderman-2023-pythia}, and Qwen2-7B \cite{yang2024qwen2technicalreport}.
The details of the models and their Hugging Face identifiers are summarized in \Cref{tab:hf-models}.

\begin{table}[t]
\centering
\begin{tabular}{ll}
\toprule
\textbf{Model} & \textbf{HF Identifier} \\ \midrule
Llama-2-7b-hf & \href{https://huggingface.co/meta-llama/Llama-2-7b-hf}{meta-llama/Llama-2-7b-hf} \\
Llama-3.2-3B & \href{https://huggingface.co/meta-llama/Llama-3.2-3B}{meta-llama/Llama-3.2-3B} \\
Mistral-7B-v0.3 & \href{https://huggingface.co/mistralai/Mistral-7B-v0.3}{mistralai/Mistral-7B-v0.3} \\
Qwen2-7B & \href{https://huggingface.co/Qwen/Qwen2-7B}{Qwen/Qwen2-7B} \\
pythia-1b & \href{https://huggingface.co/EleutherAI/pythia-1b}{EleutherAI/pythia-1b} \\
\bottomrule
\end{tabular}
\caption{Open-weight LMs used in this study, with their Hugging Face identifiers as reported by the respective sources.
Parameter counts are indicated in the model names.}
\label{tab:hf-models}
\end{table}

This selection covers several architectural variations in the attention and normalization components:
\begin{enumerate}
    \item \textbf{Attention and Normalization Formats:} Spanning standard Multi-Head Attention (Llama-2-7b-hf, pythia-1b) and Grouped-Query Attention (Llama-3.2-3B, Mistral-7B-v0.3, Qwen2-7B), as well as contrasting standard Layer Normalization (pythia-1b) with RMSNorm (Llama families, Mistral-7B-v0.3, Qwen2-7B).
    \item \textbf{Positional Encoding and Projection Biases:} pythia-1b applies RoPE to only 25\% of each head's dimensions, whereas the other models apply it to the full head dimension.
    Qwen2-7B and pythia-1b use biases in their QKV projections, whereas the Llama and Mistral models do not.
\end{enumerate}

We used an NVIDIA RTX 6000 Ada Generation GPU for our experiments, which took approximately 4 hours for the main experiments, with additional compute for the position-sweep experiments in \Cref{appendix:varying-intervened-position}.
All experiments were implemented with \texttt{PyTorch} 2.9.0 and \texttt{transformers} 4.57.1.

\begin{figure*}[t]
  \centering
  \begin{subfigure}[t]{0.3\linewidth}
    \centering
    \includegraphics[width=\linewidth]{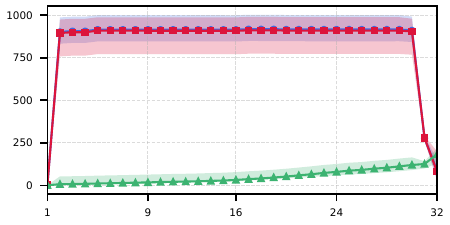}
    \caption{Llama-2-7b-hf}
  \end{subfigure}
  \hspace{0.02\linewidth}
  \begin{subfigure}[t]{0.3\linewidth}
    \centering
    \includegraphics[width=\linewidth]{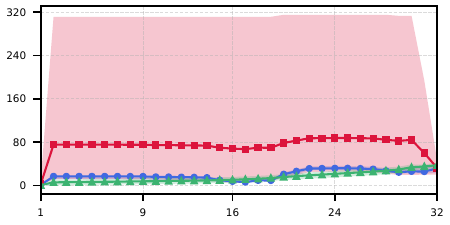}
    \caption{Mistral-7B-v0.3}
  \end{subfigure}

  \begin{subfigure}[t]{0.3\linewidth}
    \centering
    \includegraphics[width=\linewidth]{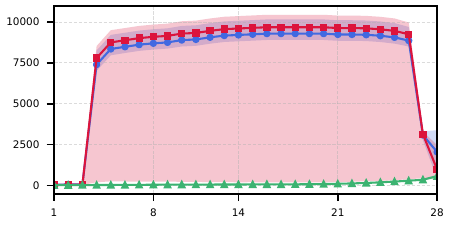}
    \caption{Qwen2-7B}
  \end{subfigure}
  \hspace{0.02\linewidth}
  \begin{subfigure}[t]{0.3\linewidth}
    \centering
    \includegraphics[width=\linewidth]{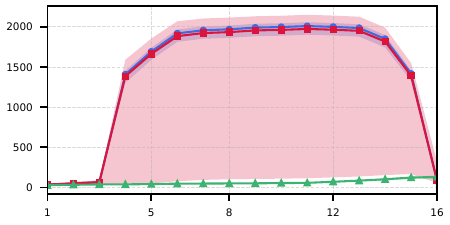}
    \caption{pythia-1b}
  \end{subfigure}

  \caption{Layer-wise averaged hidden-state norms under the self-concentration intervention (w/o BOS token) for the four models other than Llama-3.2-3B.
  Blue circles, red squares, and green triangles denote the first token, the intervention token, and the mean over all other input tokens, respectively.
  The horizontal axis shows the layer index in every panel.
  Shaded bands show the sample min--max range.}
  \label{fig:activation_norms_remove_bos_self_only_other_models}
\end{figure*}

\section{Transition of \texorpdfstring{\sinkRate{j}}{Sink Rate} with relocated BOS}
\label{sec:sink_rate_with_moved_bos}
\begin{figure}[t]
    \centering
    \begin{subfigure}[b]{\linewidth}
        \includegraphics[width=\linewidth]{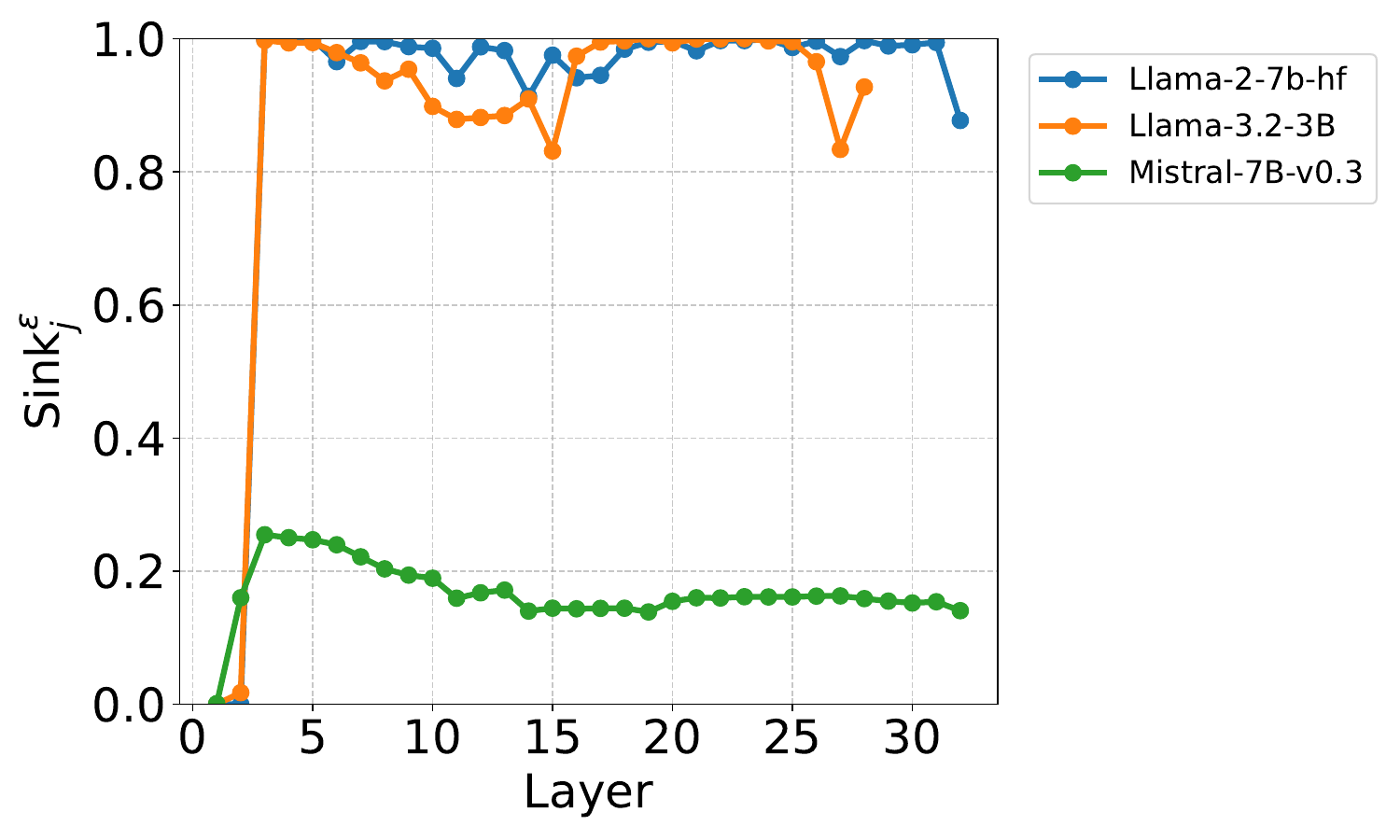}
        \caption{\sinkRate{1}}
        \label{fig:sink_rate_idx0}
    \end{subfigure}
    \\
    \begin{subfigure}[b]{\linewidth}
        \includegraphics[width=\linewidth]{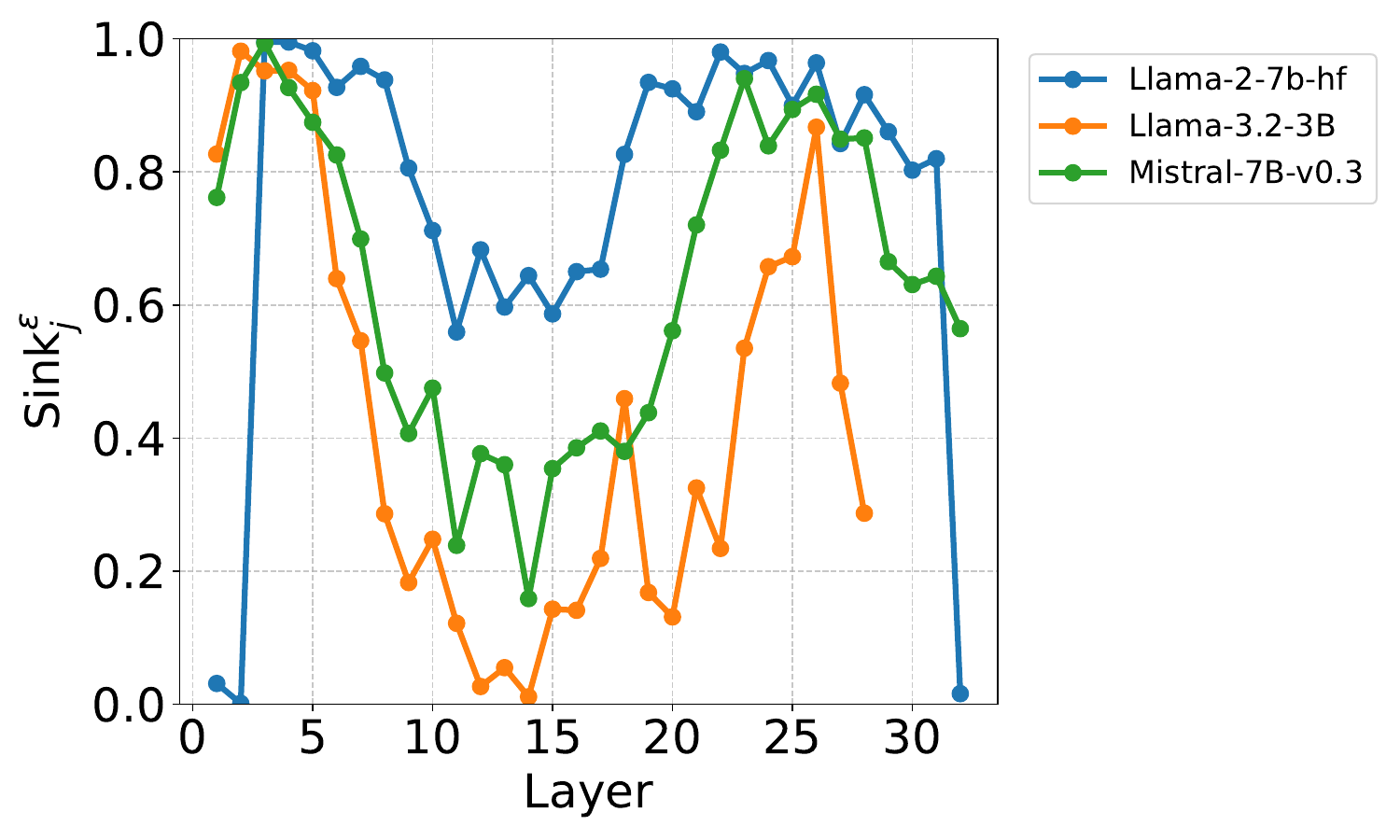}
        \caption{\sinkRate{16} (moved BOS)}
        \label{fig:sink_rate_idx15}
    \end{subfigure}
    \caption{Transition of layer-wise \sinkRate{j} when the BOS token is swapped with the token at position 16 (the setting of \Cref{tab:move_bos}).}
    \label{fig:sink_rate_transition_when_move_bos}
\end{figure}

\Cref{fig:sink_rate_transition_when_move_bos} shows the layer-wise transition of \sinkRate{j} under the setting of \Cref{sec:exp-bos}, at the sequence-initial position ($j=1$) and at the position to which the BOS token is relocated ($j=16$).
As in \Cref{tab:move_bos}, we report the three models whose tokenizer prepends a BOS token.
Below we summarize the layers at which \asma{} emerge for the relocated BOS token in each model:

\begin{itemize}
    \item \textbf{Llama-3.2-3B}: \asma{} are observed for the relocated BOS token from Layer 1 on.
    \item \textbf{Mistral-7B-v0.3}: \as{} emerges from Layer 1, whereas \ma{} emerge from Layer 2.
    \as{} therefore precedes \ma{} here, reversing the order we observe for Llama-2-7b-hf below.
    \item \textbf{Llama-2-7b-hf}: \ma{} emerge from Layer 2 and \as{} from Layer 3, so both appear in later layers than in the other two models.
\end{itemize}

\section{Results for Repeated Token Sequences across Models}
\label{appendix:repeated-token-across-model}
\begin{figure}[t]
  \centering
  \begin{subfigure}[t]{0.48\linewidth}
    \centering
    \includegraphics[width=\linewidth]{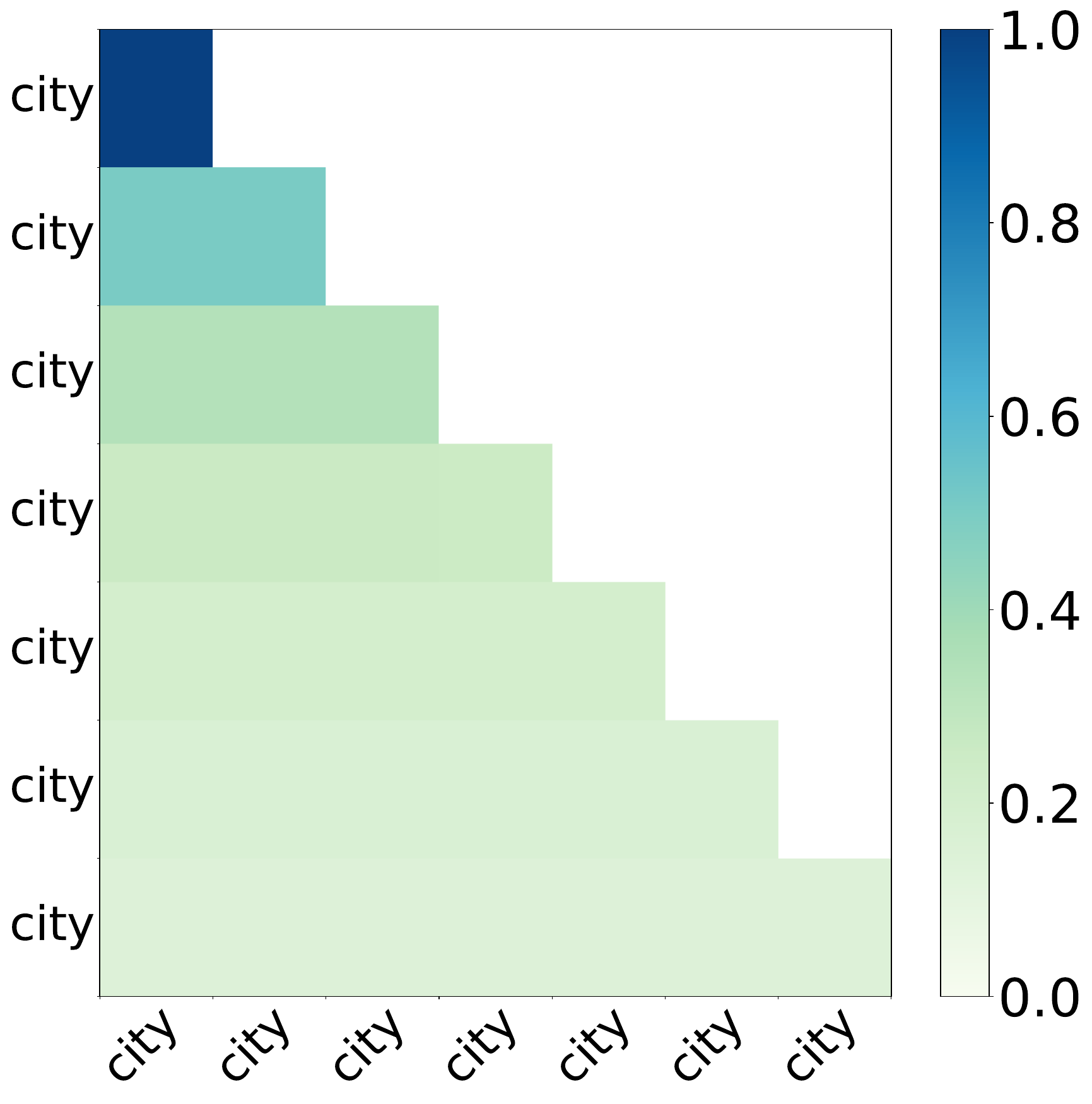}
    \caption{Heatmap: Uniform sequence}
    \label{fig:heatmap_all_same_llama2_s2}
  \end{subfigure}
  \begin{subfigure}[t]{0.48\linewidth}
      \centering
      \includegraphics[width=\linewidth]{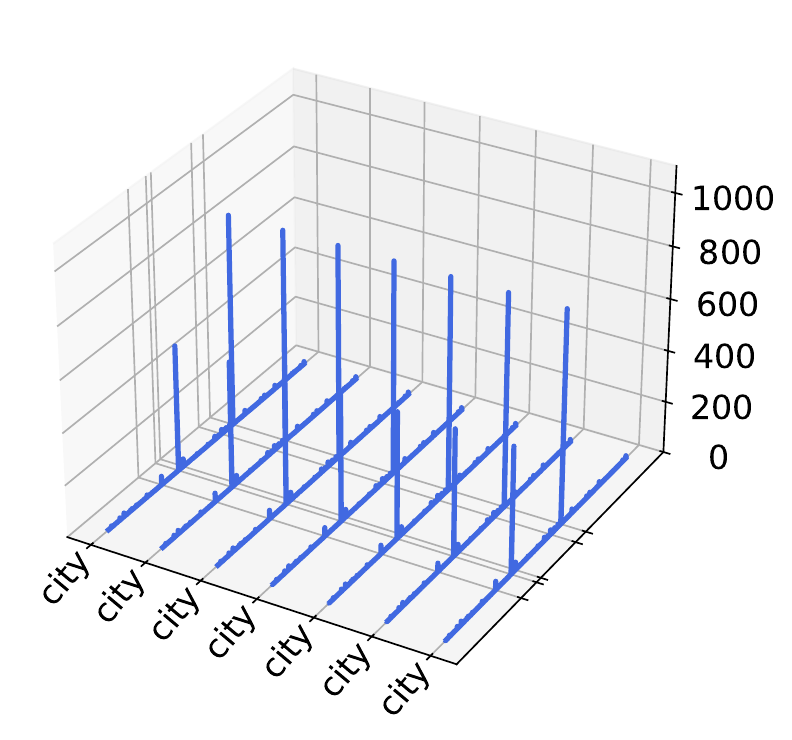}
      \caption{Activation: Uniform sequence}
      \label{fig:act_all_same_llama2_s2}
  \end{subfigure}

  \begin{subfigure}[t]{0.48\linewidth}
    \centering
    \includegraphics[width=\linewidth]{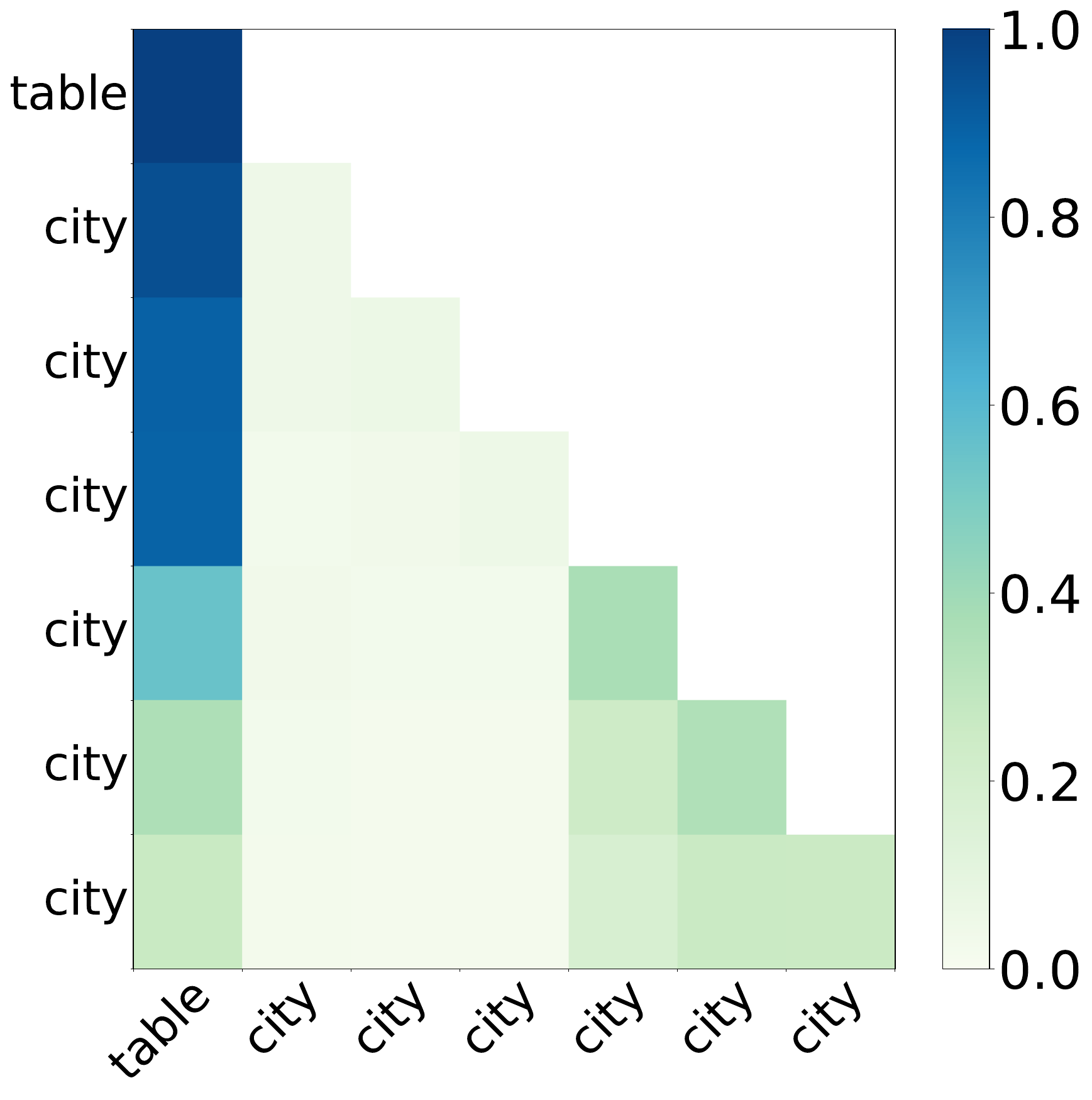}
    \caption{Heatmap: Distinct token at pos. 1}
    \label{fig:heatmap_diff_first_llama2_s2}
  \end{subfigure}
  \begin{subfigure}[t]{0.48\linewidth}
      \centering
      \includegraphics[width=\linewidth]{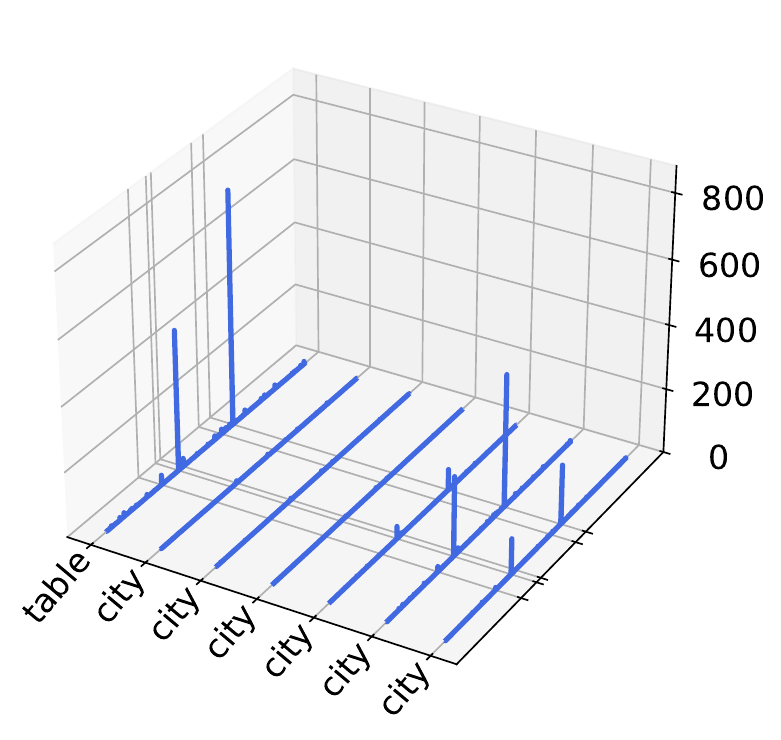}
      \caption{Activation: Distinct token at pos. 1}
      \label{fig:act_diff_first_llama2_s2}
  \end{subfigure}

  \begin{subfigure}[t]{0.48\linewidth}
    \centering
    \includegraphics[width=\linewidth]{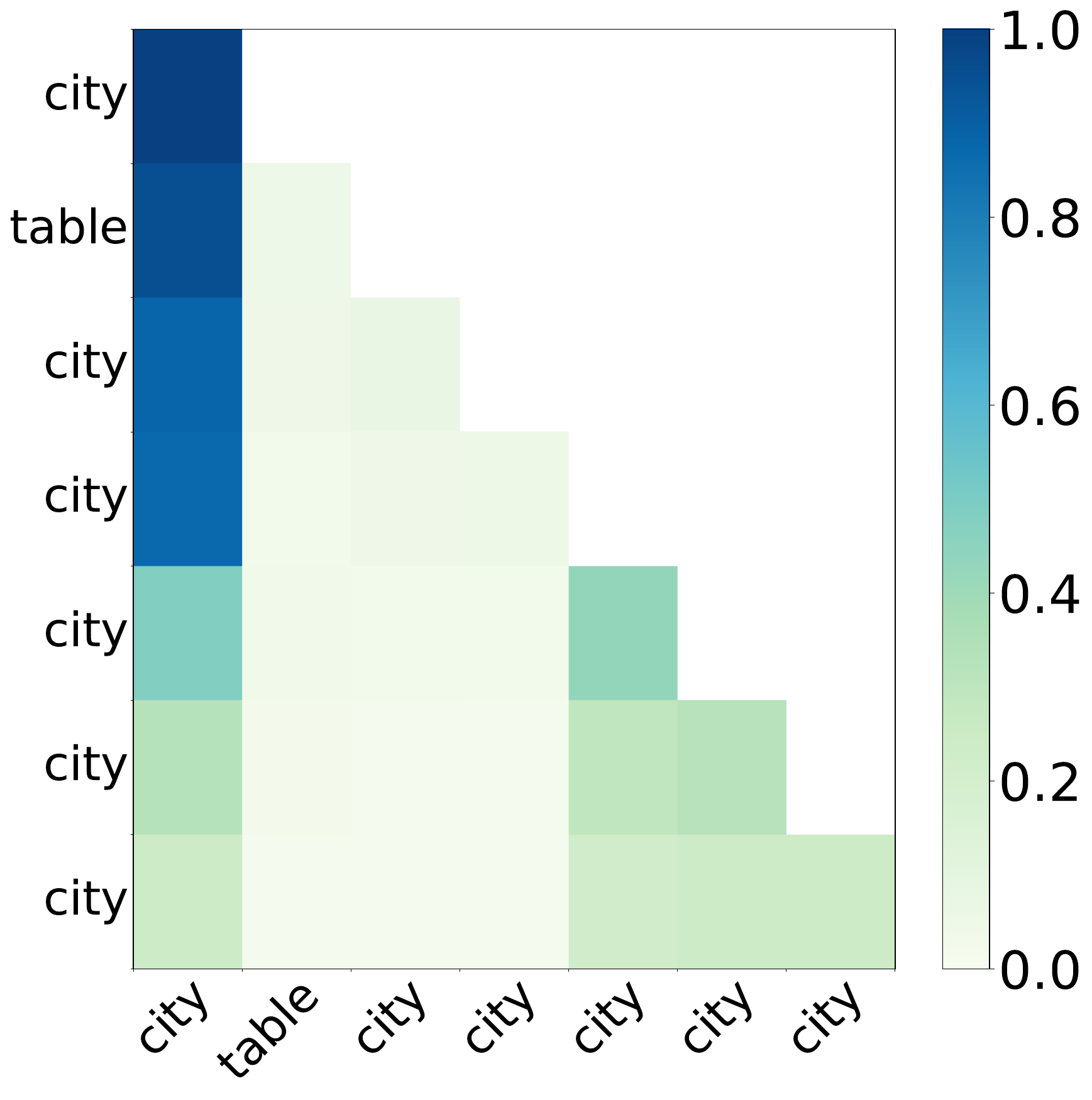}
    \caption{Heatmap: Distinct token at pos. 2}
    \label{fig:heatmap_diff_second_llama2_s2}
  \end{subfigure}
  \begin{subfigure}[t]{0.48\linewidth}
      \centering
      \includegraphics[width=\linewidth]{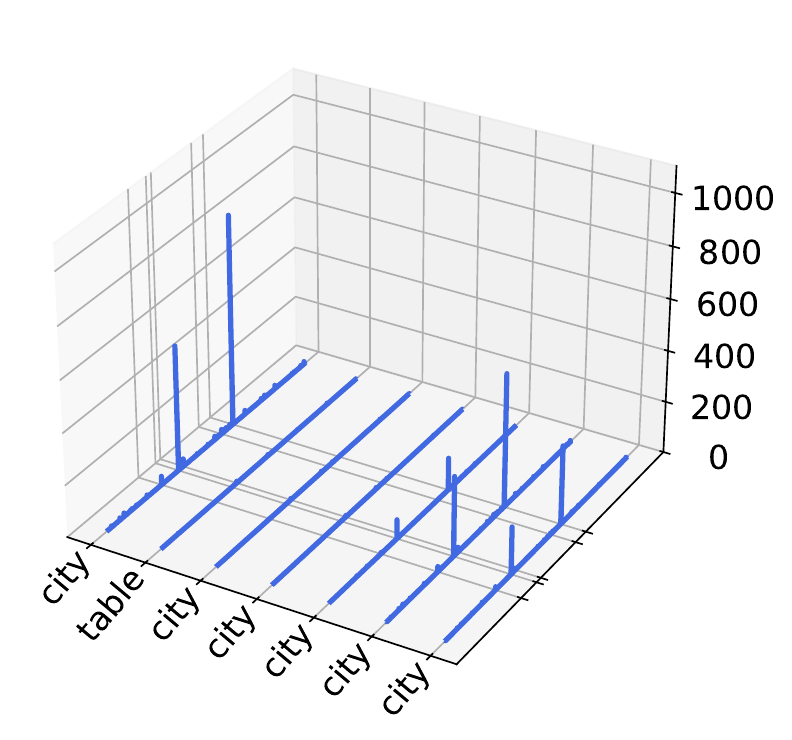}
      \caption{Activation: Distinct token at pos. 2}
      \label{fig:act_diff_second_llama2_s2}
  \end{subfigure}

  \begin{subfigure}[t]{0.48\linewidth}
      \centering
      \includegraphics[width=\linewidth]{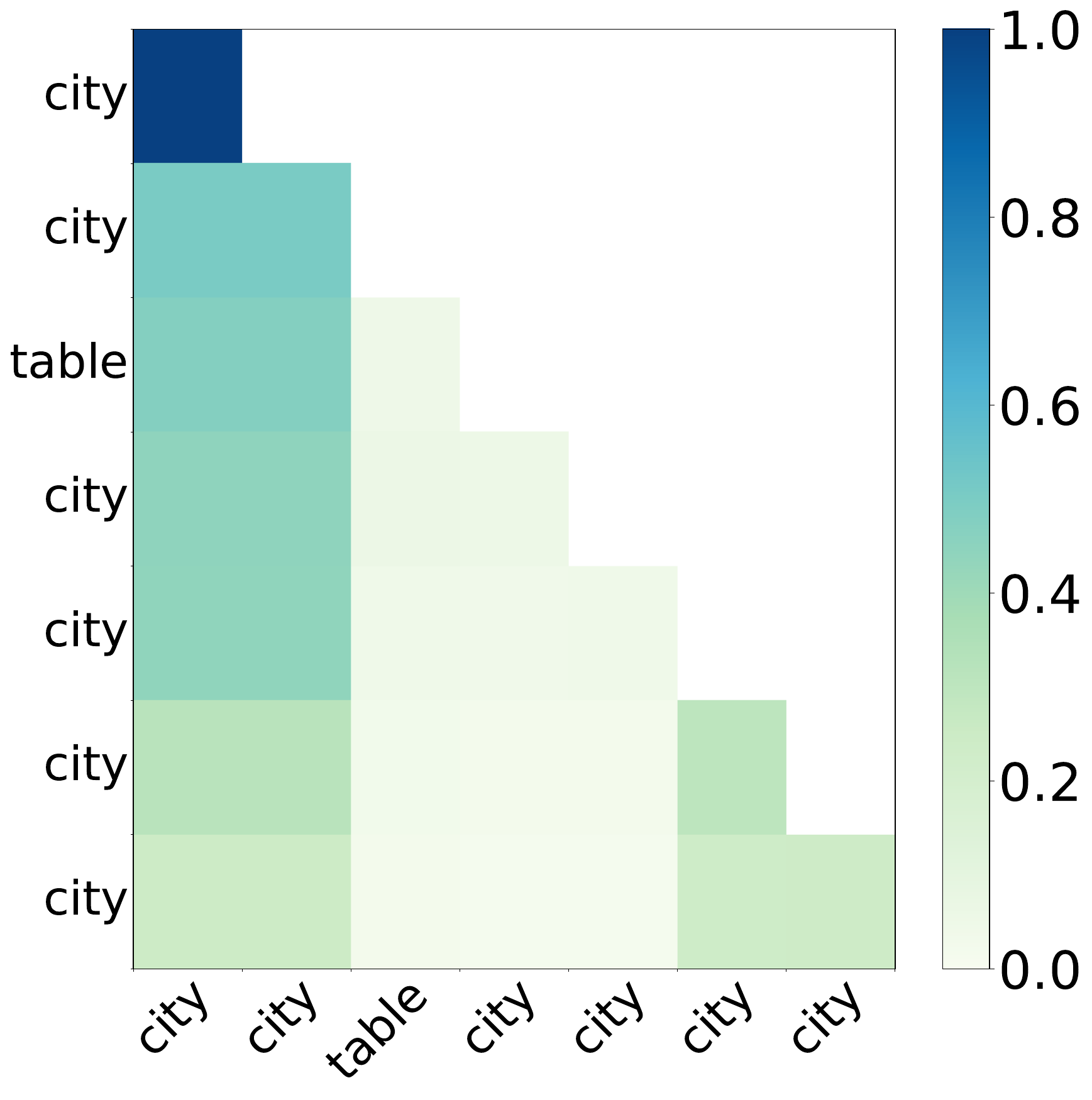}
      \caption{Heatmap: Distinct token at pos. 3}
      \label{fig:heatmap_diff_third_llama2_s2}
  \end{subfigure}
  \begin{subfigure}[t]{0.48\linewidth}
      \centering
      \includegraphics[width=\linewidth]{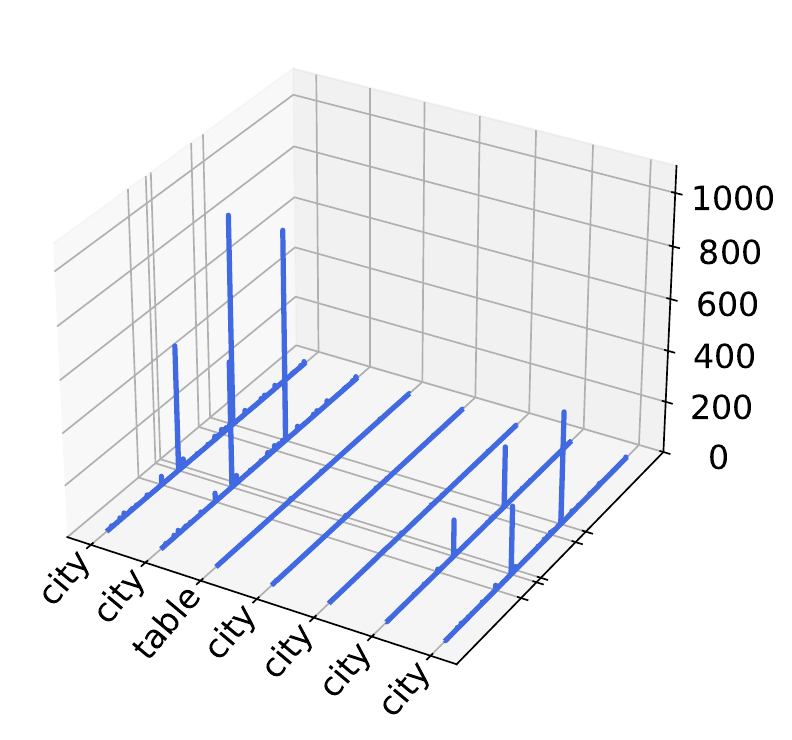}
      \caption{Activation: Distinct token at pos. 3}
      \label{fig:act_diff_third_llama2_s2}
  \end{subfigure}

  \caption{Results for sequences with repeated tokens.
  Each row compares the attention heatmap (left) and hidden-state activations (right) under the same token distribution pattern for Llama-2-7b-hf (Sample 2).}
  \label{fig:repeated_tokens_results_llama2_s2}
\end{figure}

\begin{figure}[t]
  \centering
  \begin{subfigure}[t]{0.48\linewidth}
    \centering
    \includegraphics[width=\linewidth]{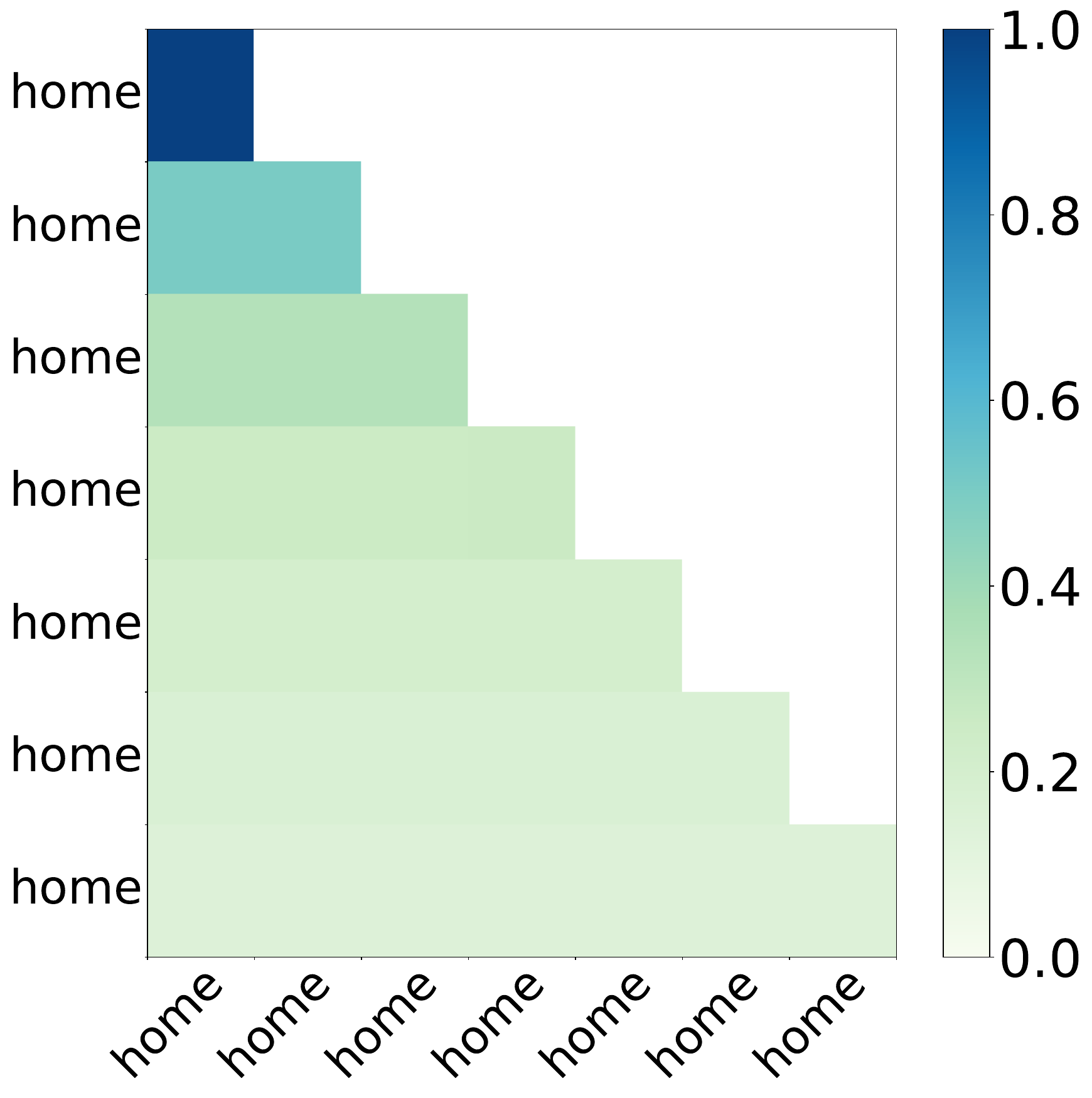}
    \caption{Heatmap: Uniform sequence}
    \label{fig:heatmap_all_same_llama2_s1}
  \end{subfigure}
  \begin{subfigure}[t]{0.48\linewidth}
      \centering
      \includegraphics[width=\linewidth]{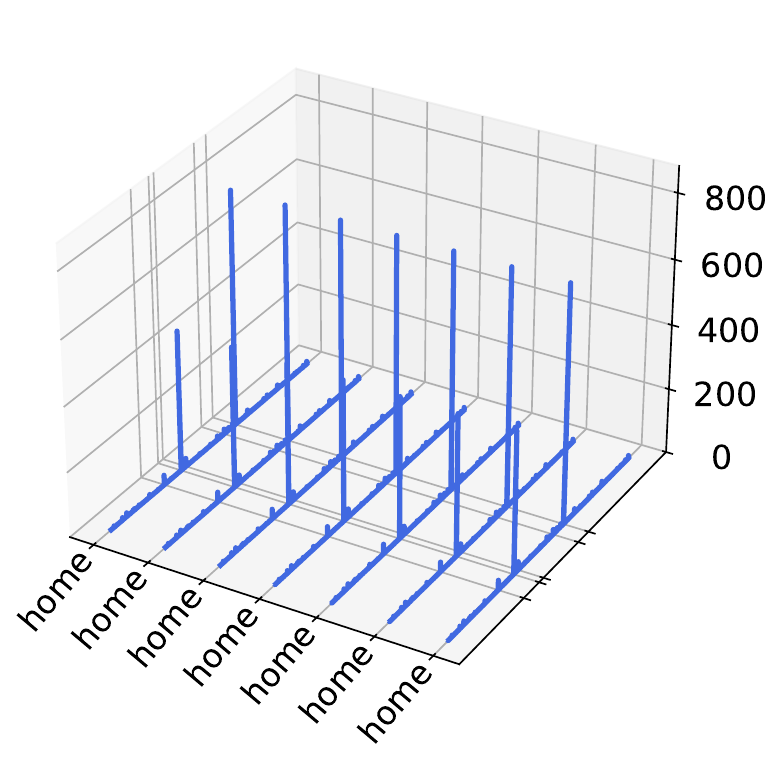}
      \caption{Activation: Uniform sequence}
      \label{fig:act_all_same_llama2_s1}
  \end{subfigure}

  \begin{subfigure}[t]{0.48\linewidth}
    \centering
    \includegraphics[width=\linewidth]{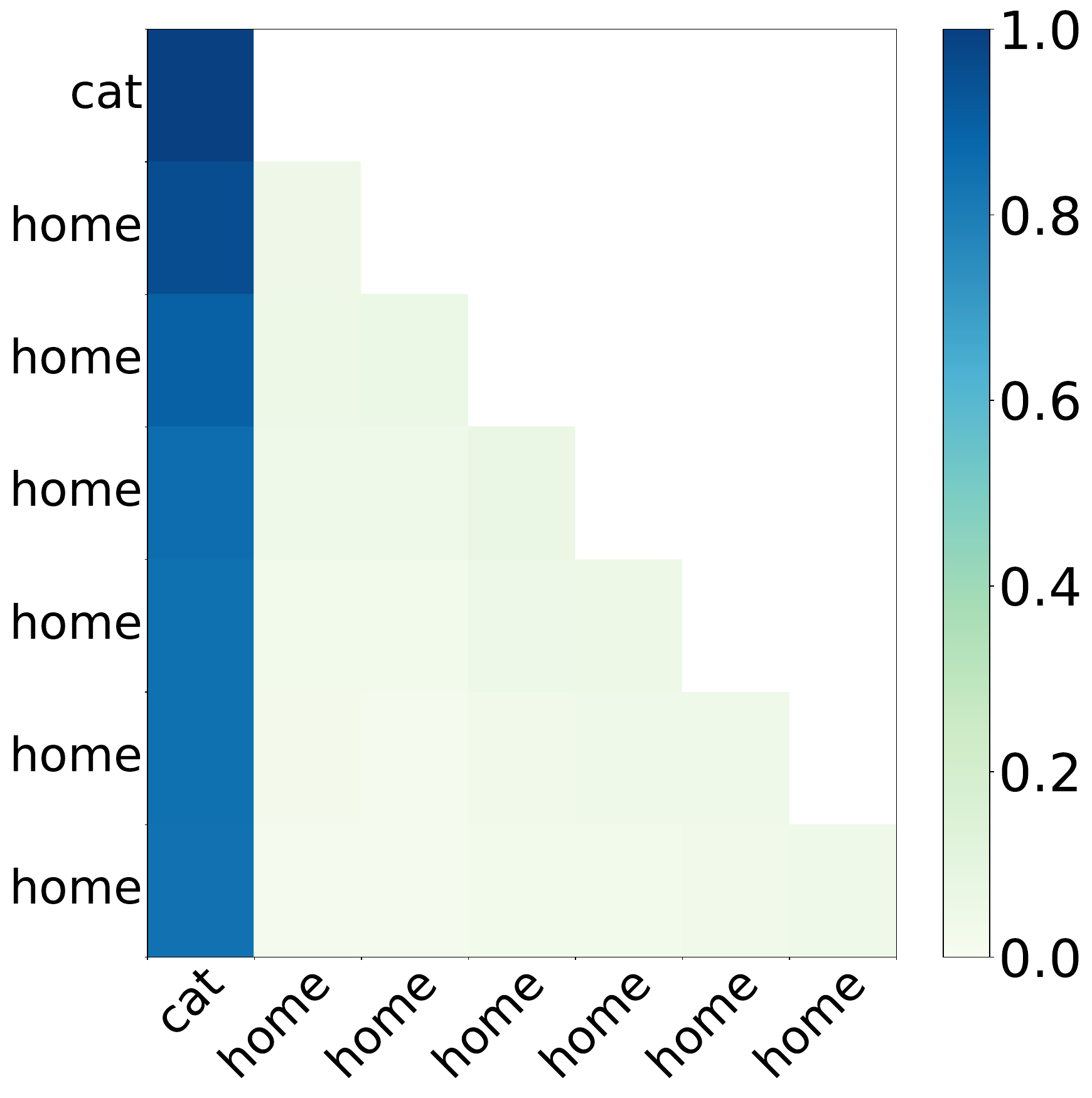}
    \caption{Heatmap: Distinct token at pos. 1}
    \label{fig:heatmap_diff_first_llama2_s1}
  \end{subfigure}
  \begin{subfigure}[t]{0.48\linewidth}
      \centering
      \includegraphics[width=\linewidth]{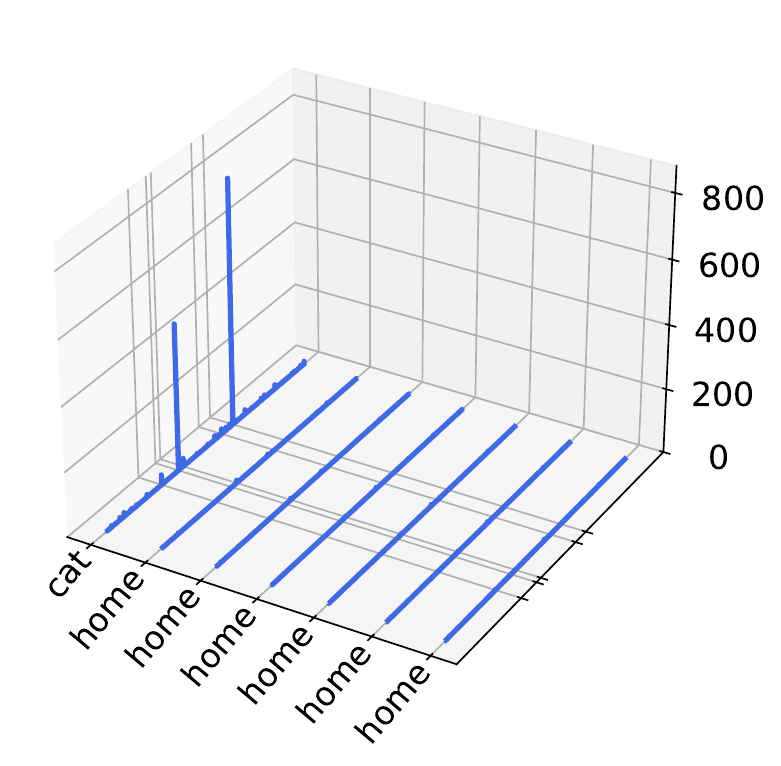}
      \caption{Activation: Distinct token at pos. 1}
      \label{fig:act_diff_first_llama2_s1}
  \end{subfigure}

  \begin{subfigure}[t]{0.48\linewidth}
    \centering
    \includegraphics[width=\linewidth]{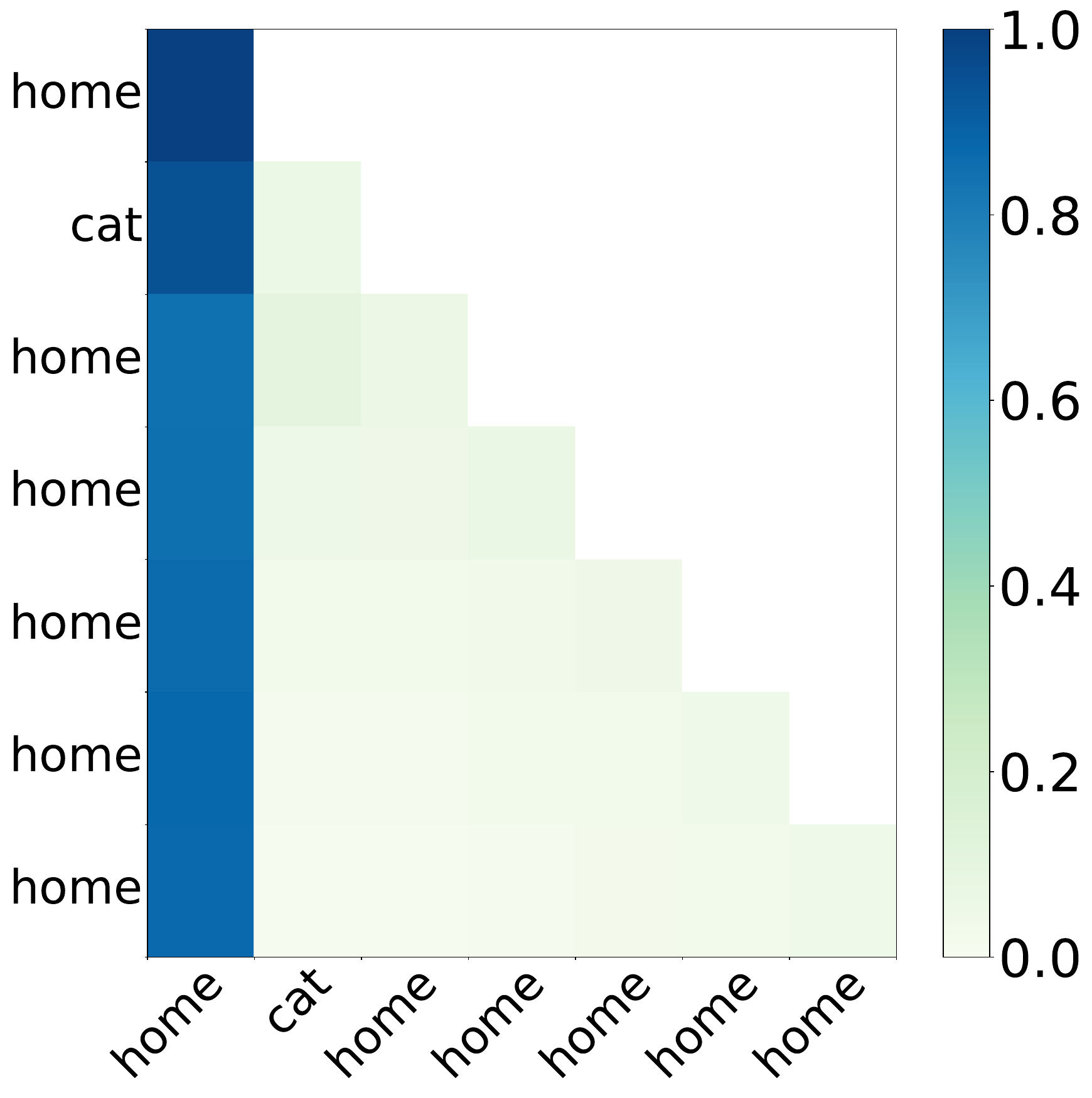}
    \caption{Heatmap: Distinct token at pos. 2}
    \label{fig:heatmap_diff_second_llama2_s1}
  \end{subfigure}
  \begin{subfigure}[t]{0.48\linewidth}
      \centering
      \includegraphics[width=\linewidth]{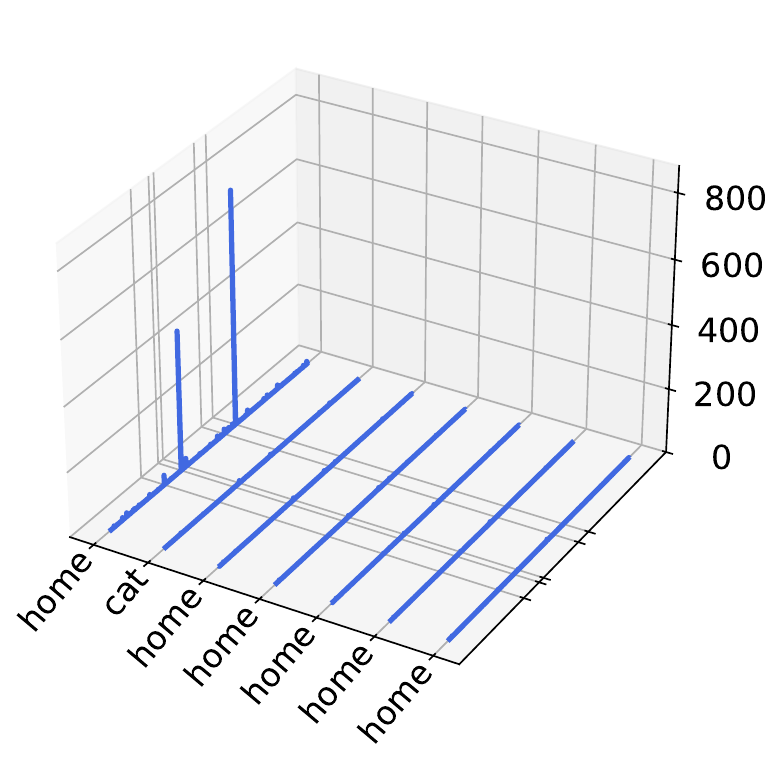}
      \caption{Activation: Distinct token at pos. 2}
      \label{fig:act_diff_second_llama2_s1}
  \end{subfigure}

  \begin{subfigure}[t]{0.48\linewidth}
      \centering
      \includegraphics[width=\linewidth]{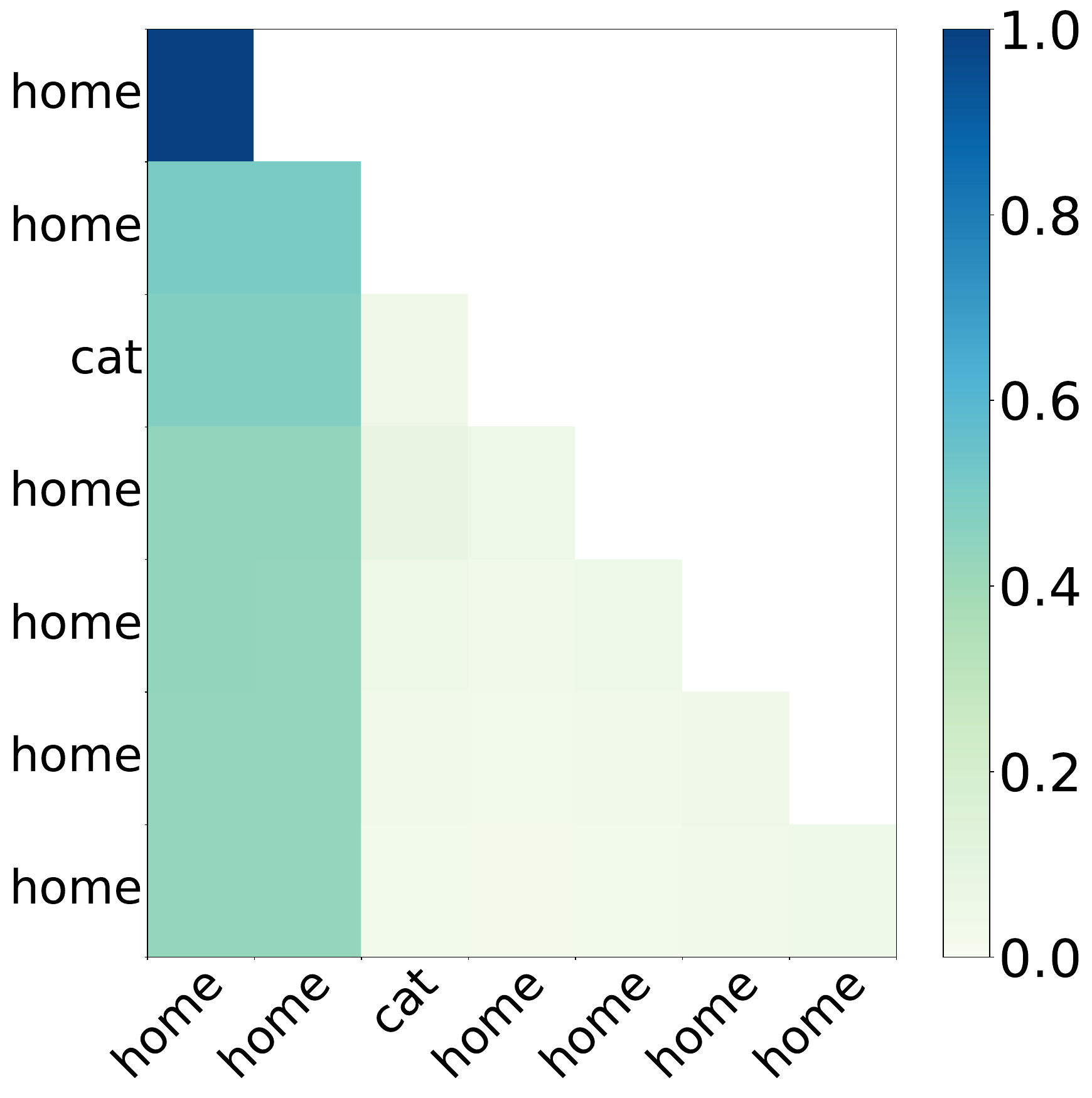}
      \caption{Heatmap: Distinct token at pos. 3}
      \label{fig:heatmap_diff_third_llama2_s1}
  \end{subfigure}
  \begin{subfigure}[t]{0.48\linewidth}
      \centering
      \includegraphics[width=\linewidth]{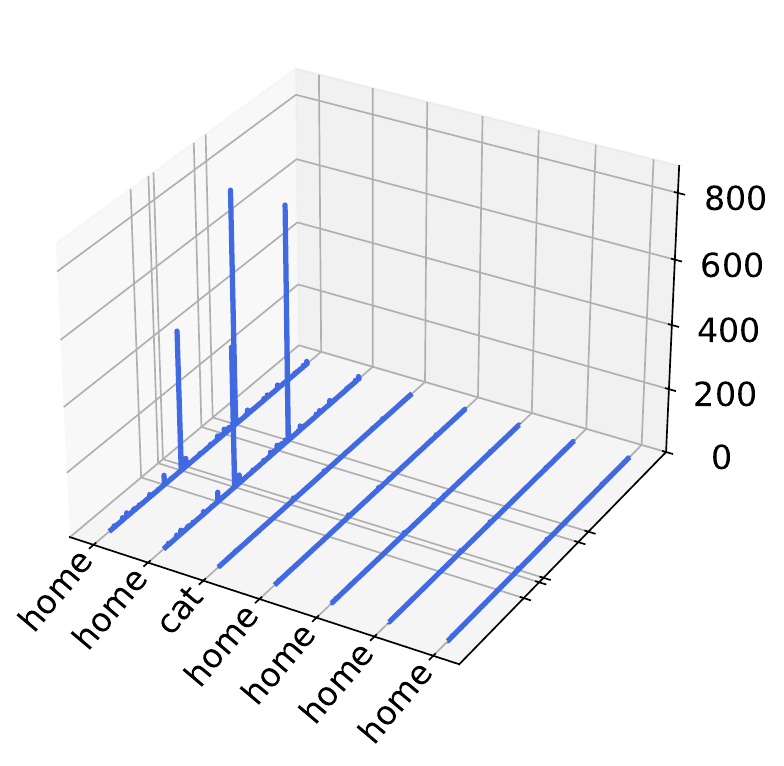}
      \caption{Activation: Distinct token at pos. 3}
      \label{fig:act_diff_third_llama2_s1}
  \end{subfigure}

  \caption{Results for sequences with repeated tokens.
  Each row compares the attention heatmap (left) and hidden-state activations (right) under the same token distribution pattern for Llama-2-7b-hf (Sample 1).}
  \label{fig:repeated_tokens_results_llama2_s1}
\end{figure}

\begin{figure*}[t]
  \centering
  \begin{subfigure}[t]{0.48\linewidth}
    \centering
    \includegraphics[width=\linewidth]{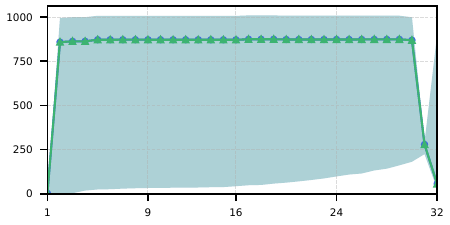}
    \caption{Norms: Uniform}
  \end{subfigure}
  \hfill
  \begin{subfigure}[t]{0.48\linewidth}
    \centering
    \includegraphics[width=\linewidth]{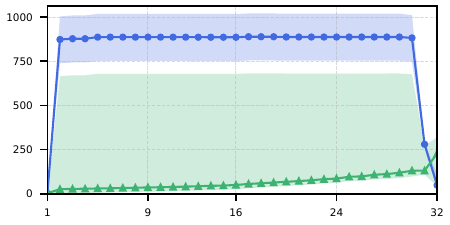}
    \caption{Norms: Pos. 1}
  \end{subfigure}

  \begin{subfigure}[t]{0.48\linewidth}
    \centering
    \includegraphics[width=\linewidth]{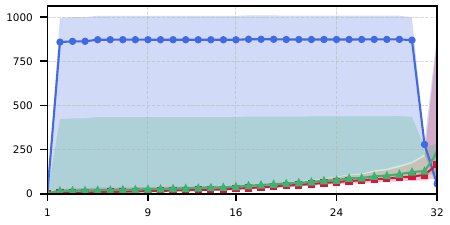}
    \caption{Norms: Pos. 2}
  \end{subfigure}
  \hfill
  \begin{subfigure}[t]{0.48\linewidth}
    \centering
    \includegraphics[width=\linewidth]{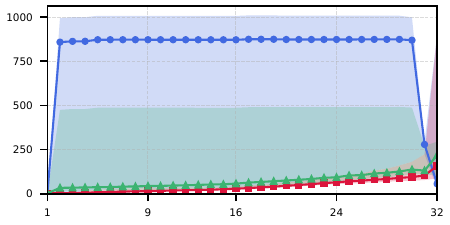}
    \caption{Norms: Pos. 3}
  \end{subfigure}
  \caption{Layer-wise averaged hidden-state norms in Llama-2-7b-hf for repeated-token sequences under different distinct-token positions (colors and markers as in \Cref{fig:activation_norms_repeated_tokens_llama3}).
  The horizontal axis shows the layer index in every panel.}
  \label{fig:activation_norms_repeated_tokens_llama2}
\end{figure*}

\begin{figure}[t]
  \centering
  \begin{subfigure}[t]{0.48\linewidth}
    \centering
    \includegraphics[width=\linewidth]{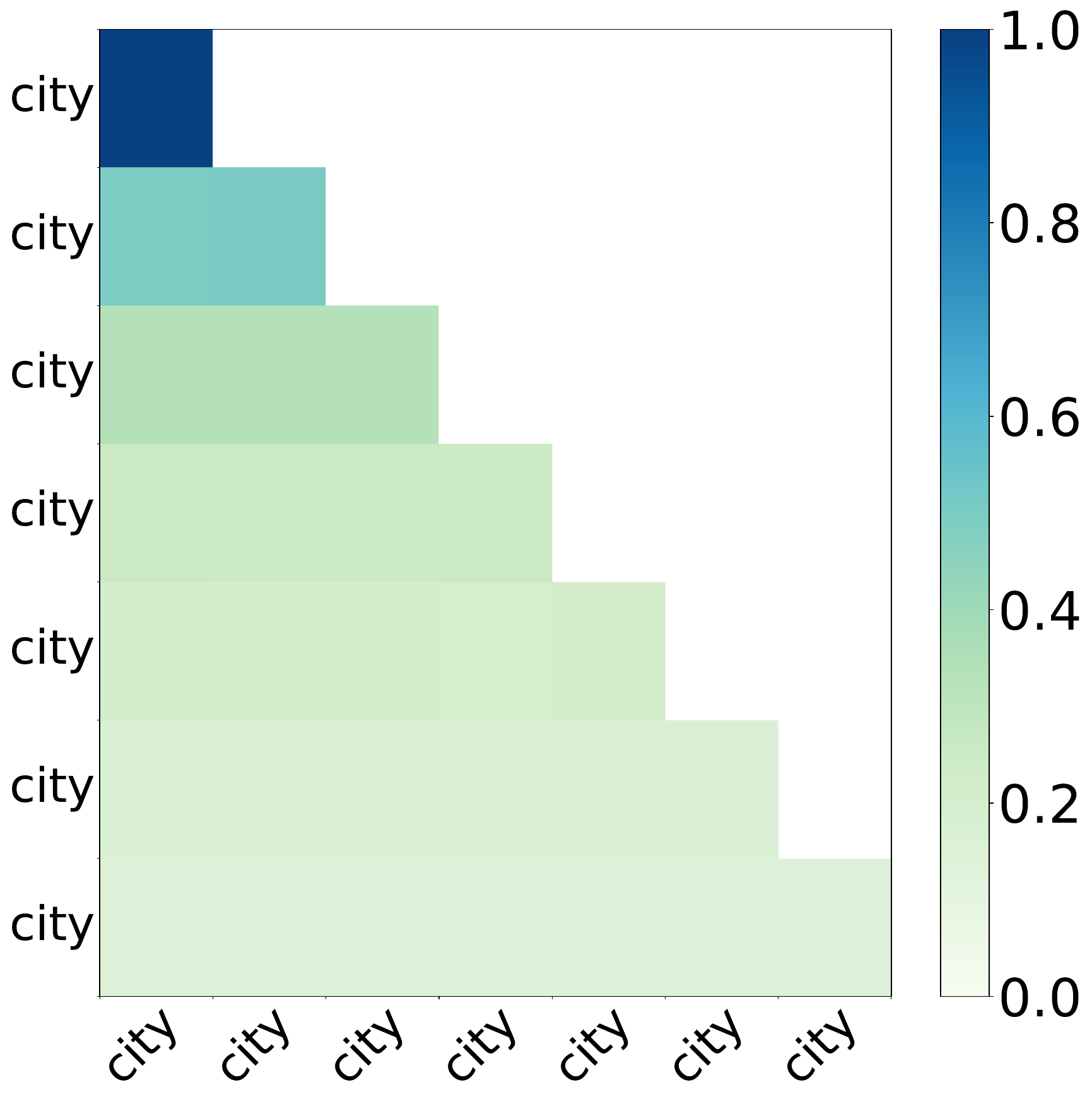}
    \caption{Heatmap: Uniform sequence}
    \label{fig:heatmap_all_same_mistral_s2}
  \end{subfigure}
  \begin{subfigure}[t]{0.48\linewidth}
      \centering
      \includegraphics[width=\linewidth]{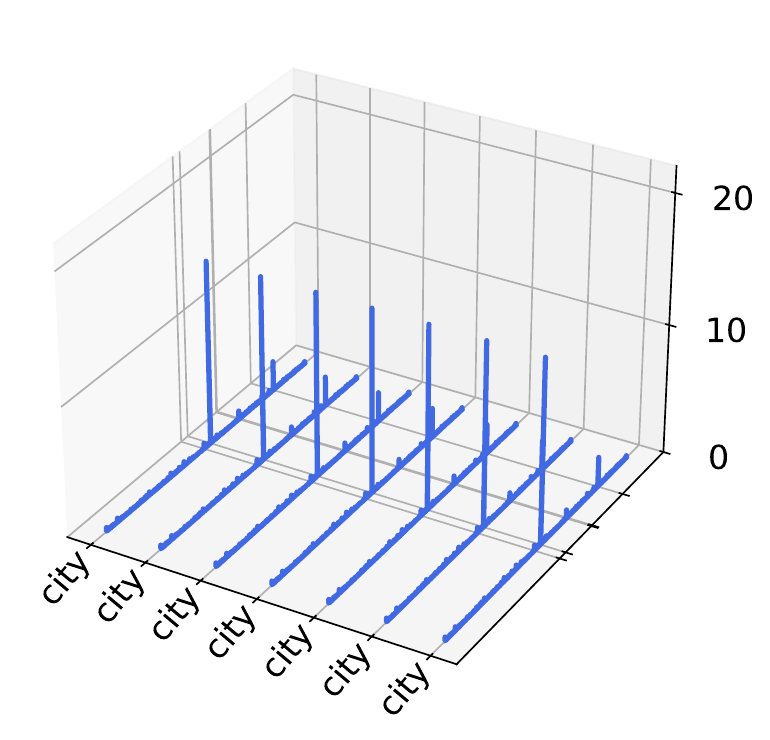}
      \caption{Activation: Uniform sequence}
      \label{fig:act_all_same_mistral_s2}
  \end{subfigure}

  \begin{subfigure}[t]{0.48\linewidth}
    \centering
    \includegraphics[width=\linewidth]{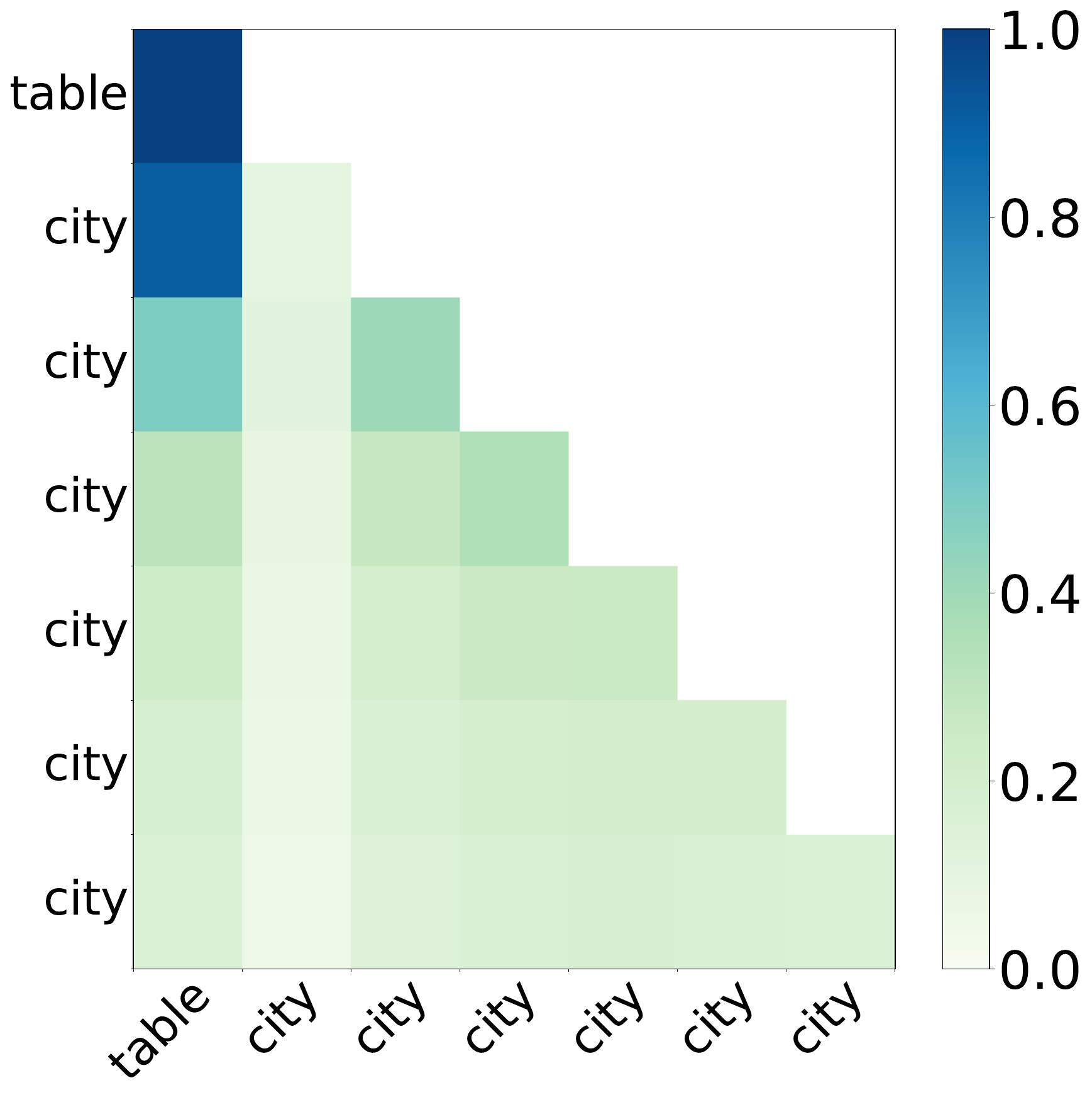}
    \caption{Heatmap: Distinct token at pos. 1}
    \label{fig:heatmap_diff_first_mistral_s2}
  \end{subfigure}
  \begin{subfigure}[t]{0.48\linewidth}
      \centering
      \includegraphics[width=\linewidth]{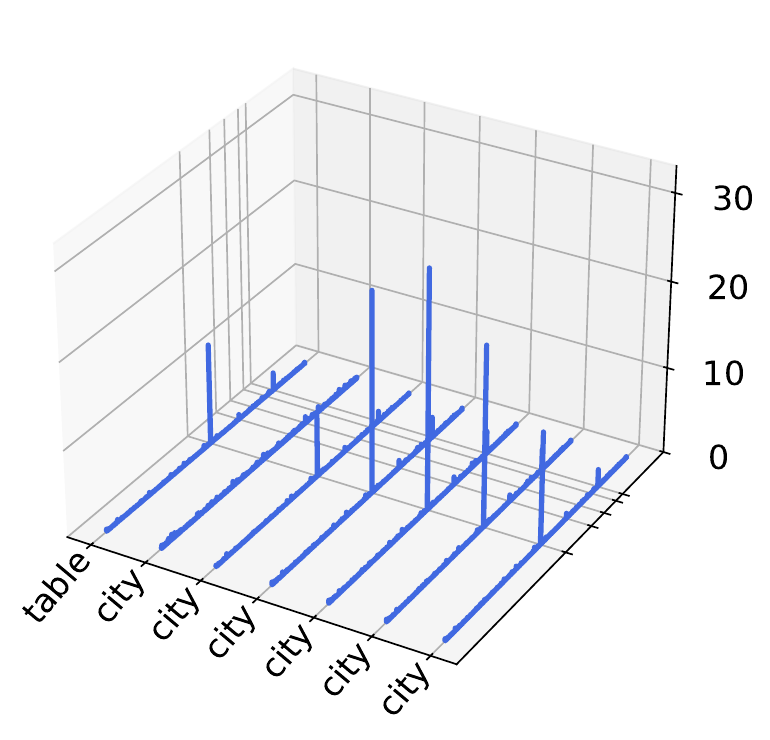}
      \caption{Activation: Distinct token at pos. 1}
      \label{fig:act_diff_first_mistral_s2}
  \end{subfigure}

  \begin{subfigure}[t]{0.48\linewidth}
    \centering
    \includegraphics[width=\linewidth]{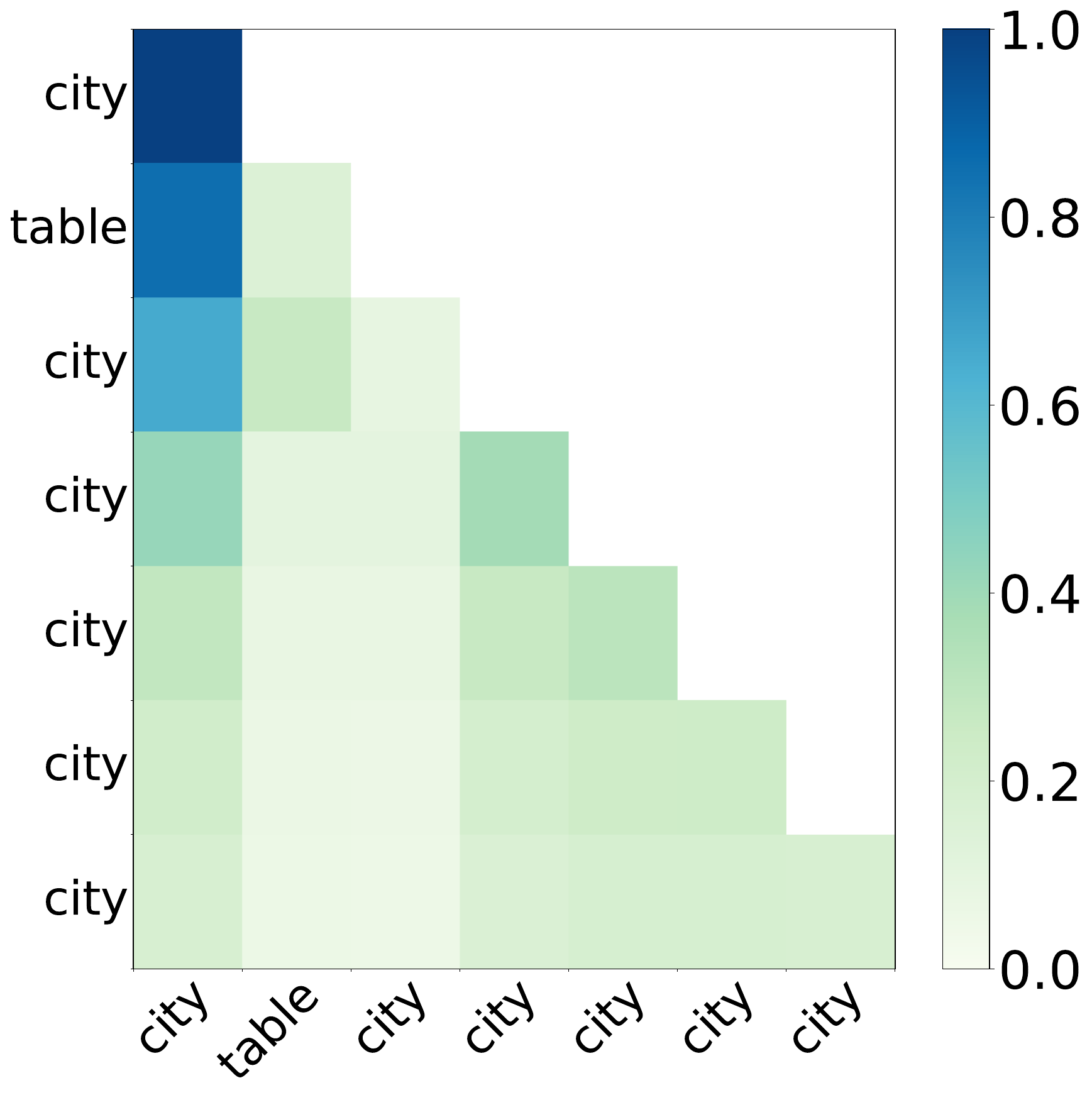}
    \caption{Heatmap: Distinct token at pos. 2}
    \label{fig:heatmap_diff_second_mistral_s2}
  \end{subfigure}
  \begin{subfigure}[t]{0.48\linewidth}
      \centering
      \includegraphics[width=\linewidth]{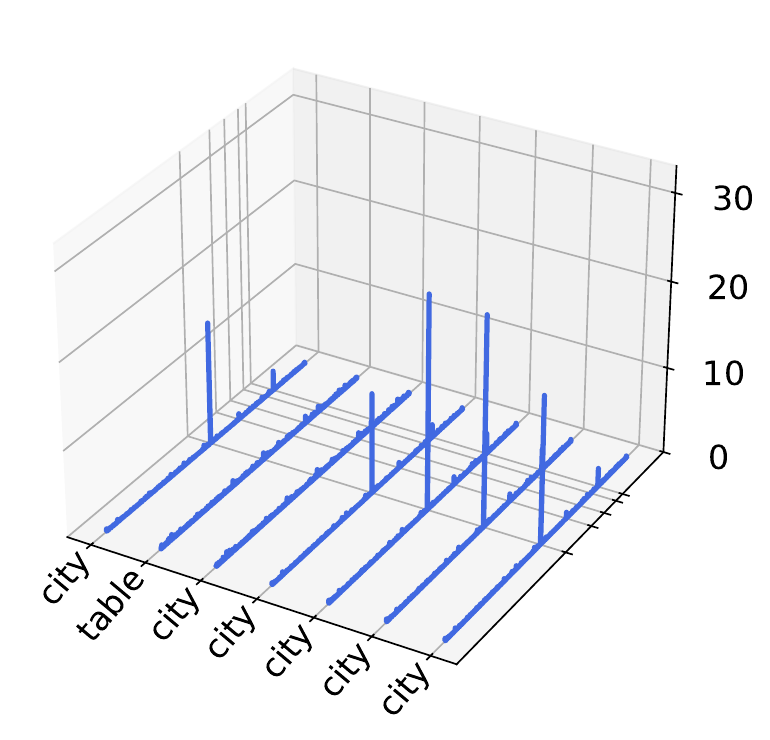}
      \caption{Activation: Distinct token at pos. 2}
      \label{fig:act_diff_second_mistral_s2}
  \end{subfigure}

  \begin{subfigure}[t]{0.48\linewidth}
      \centering
      \includegraphics[width=\linewidth]{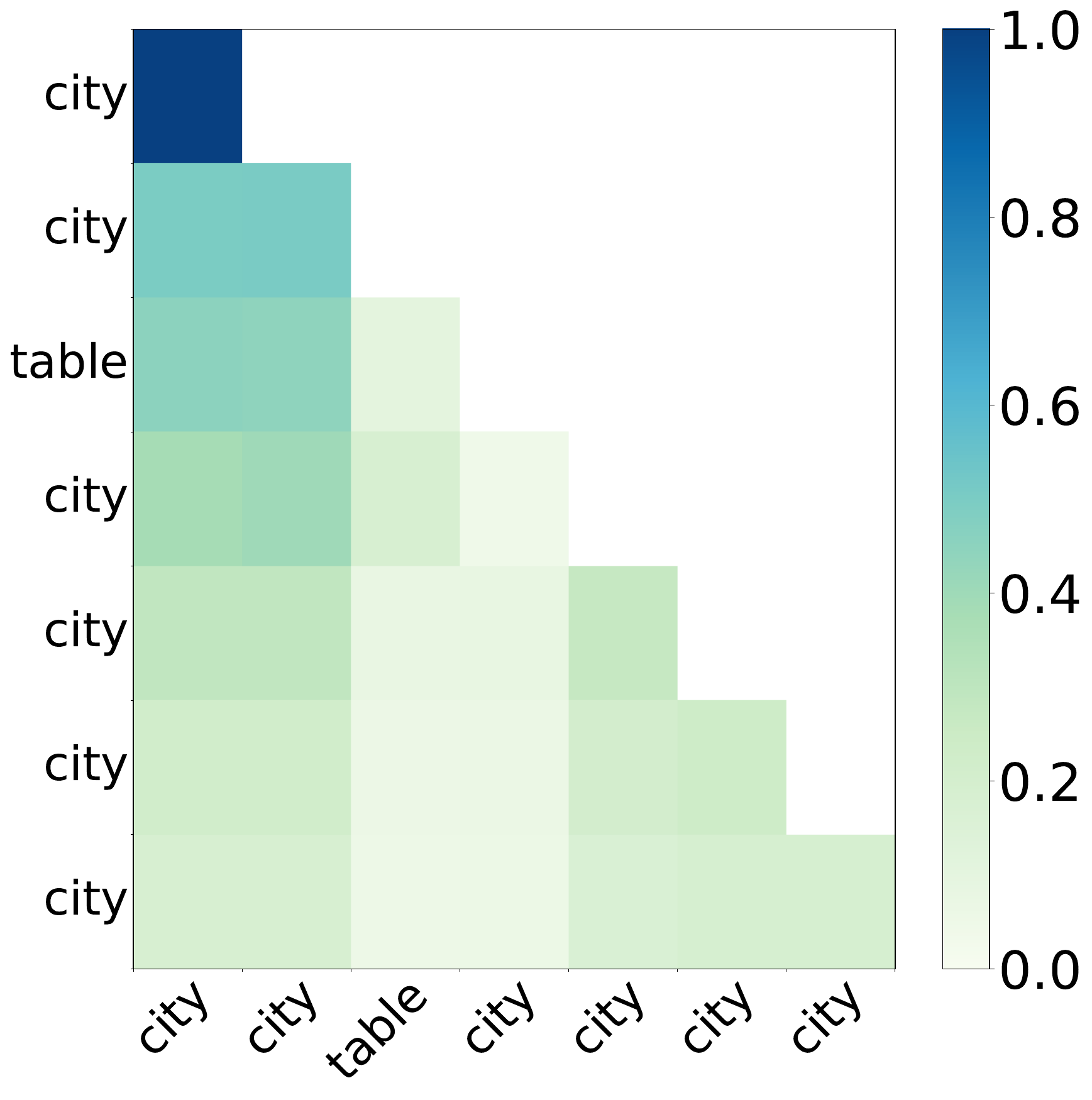}
      \caption{Heatmap: Distinct token at pos. 3}
      \label{fig:heatmap_diff_third_mistral_s2}
  \end{subfigure}
  \begin{subfigure}[t]{0.48\linewidth}
      \centering
      \includegraphics[width=\linewidth]{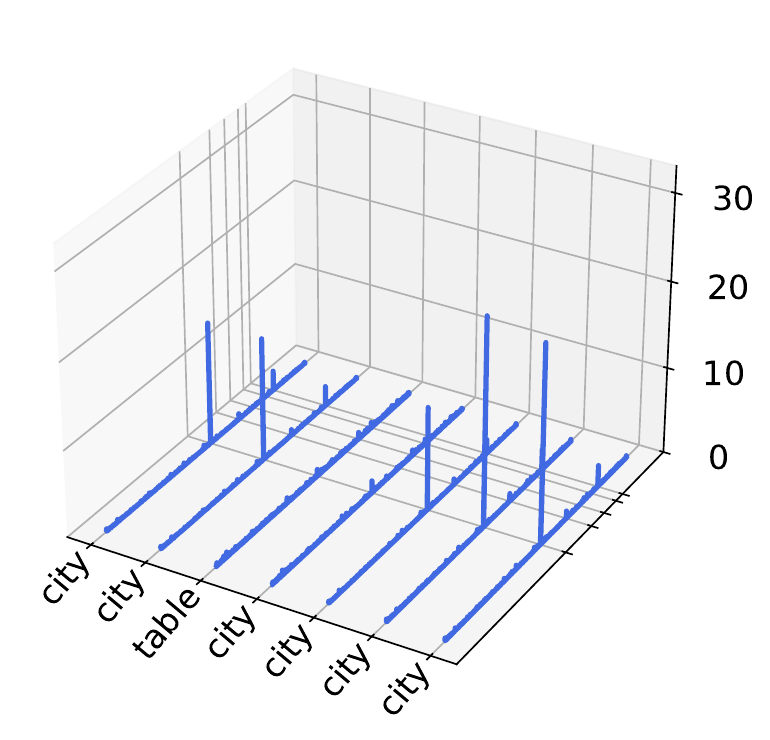}
      \caption{Activation: Distinct token at pos. 3}
      \label{fig:act_diff_third_mistral_s2}
  \end{subfigure}

  \caption{Results for sequences with repeated tokens.
  Each row compares the attention heatmap (left) and hidden-state activations (right) under the same token distribution pattern for Mistral-7B-v0.3 (Sample 2).}
  \label{fig:repeated_tokens_results_mistral_s2}
\end{figure}

\begin{figure}[t]
  \centering
  \begin{subfigure}[t]{0.48\linewidth}
    \centering
    \includegraphics[width=\linewidth]{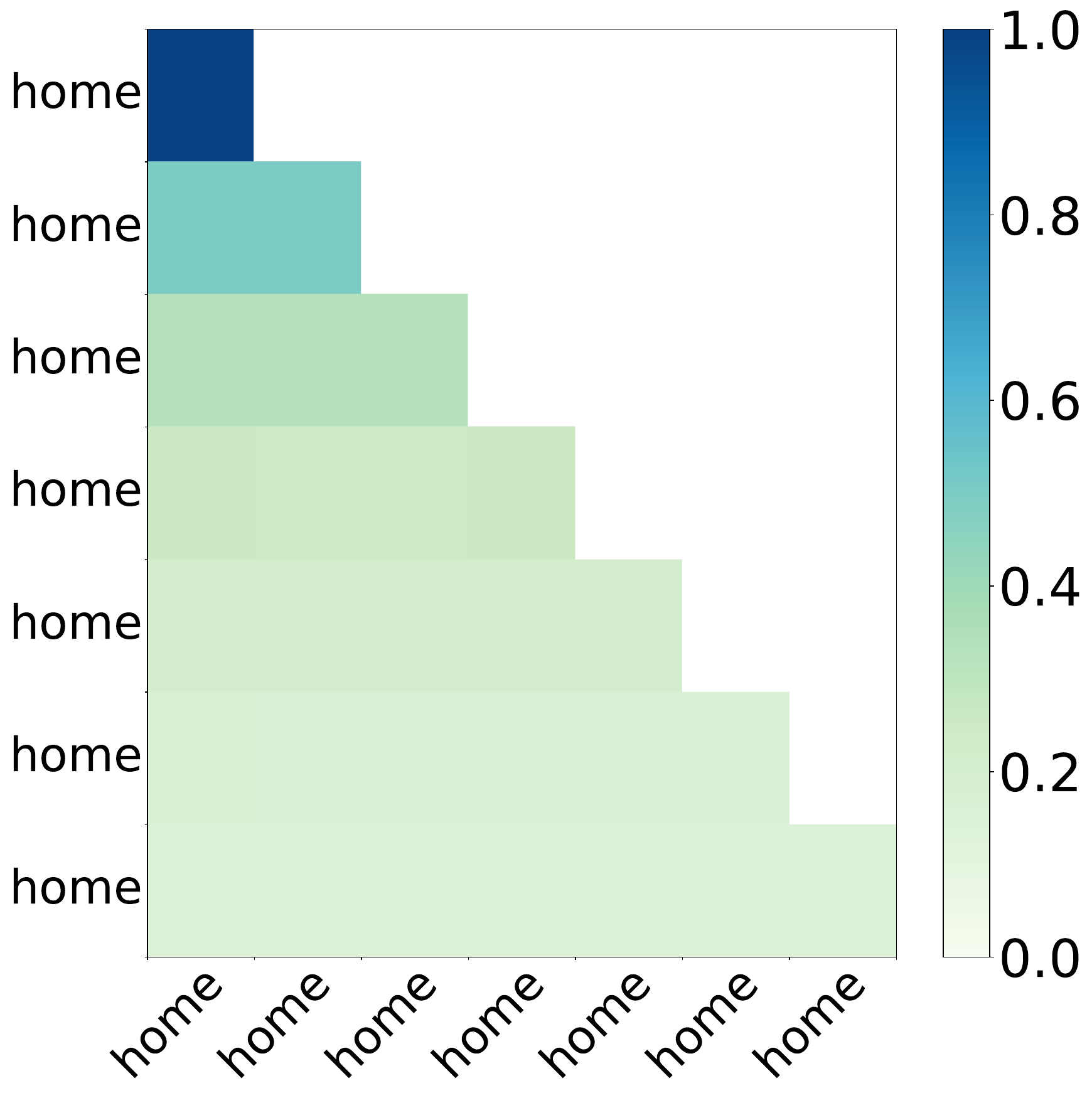}
    \caption{Heatmap: Uniform sequence}
    \label{fig:heatmap_all_same_mistral_s1}
  \end{subfigure}
  \begin{subfigure}[t]{0.48\linewidth}
      \centering
      \includegraphics[width=\linewidth]{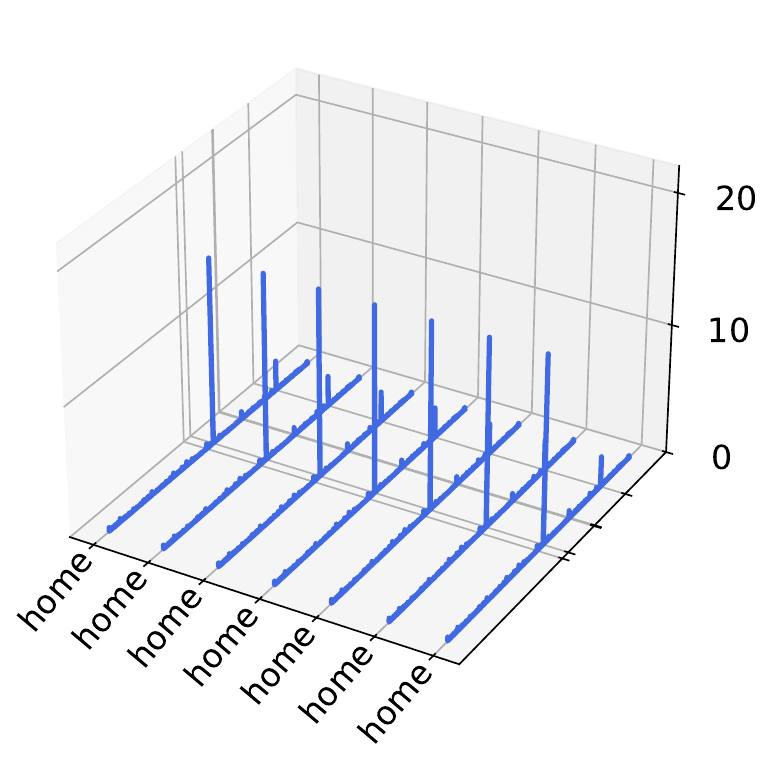}
      \caption{Activation: Uniform sequence}
      \label{fig:act_all_same_mistral_s1}
  \end{subfigure}

  \begin{subfigure}[t]{0.48\linewidth}
    \centering
    \includegraphics[width=\linewidth]{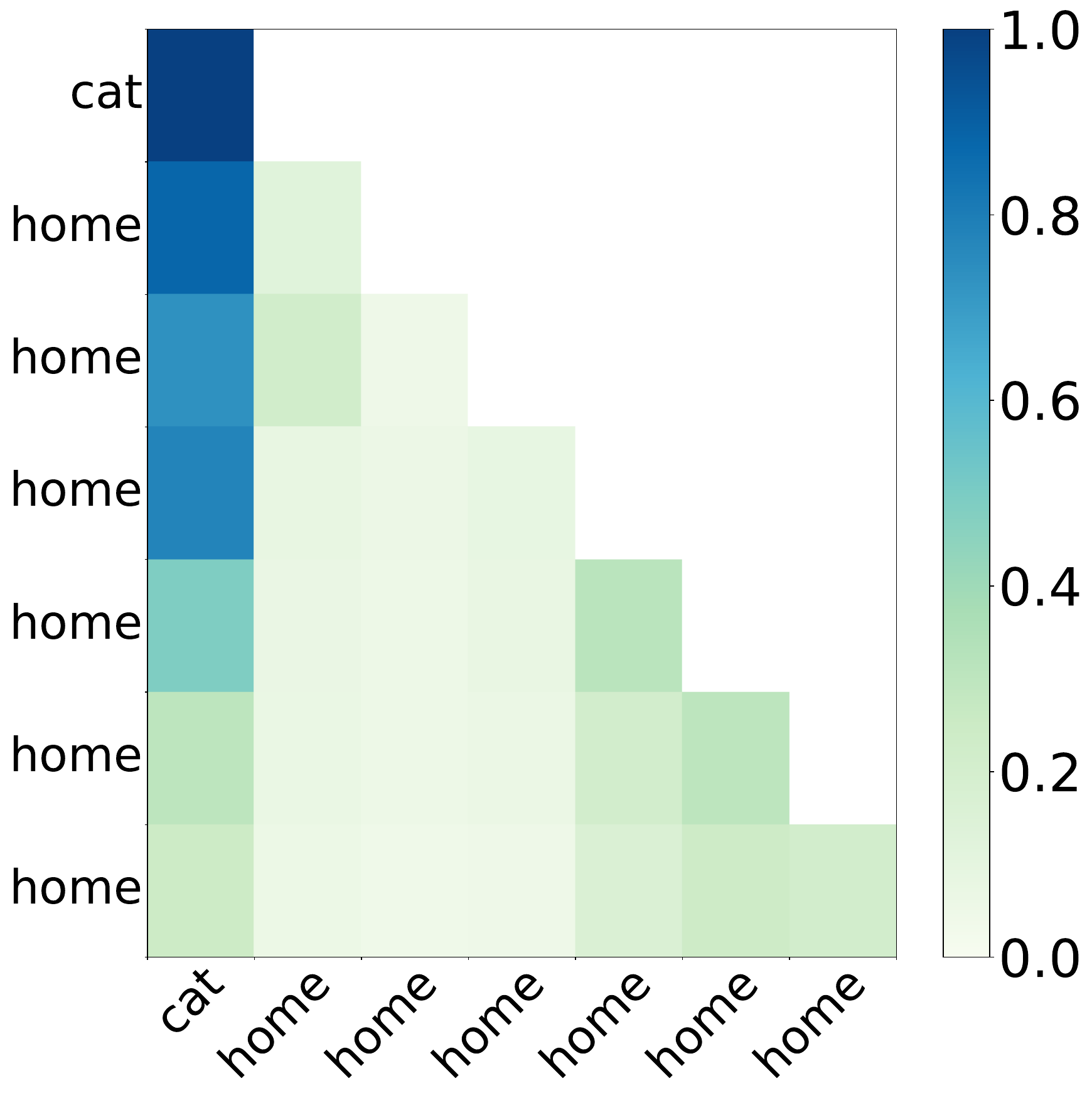}
    \caption{Heatmap: Distinct token at pos. 1}
    \label{fig:heatmap_diff_first_mistral_s1}
  \end{subfigure}
  \begin{subfigure}[t]{0.48\linewidth}
      \centering
      \includegraphics[width=\linewidth]{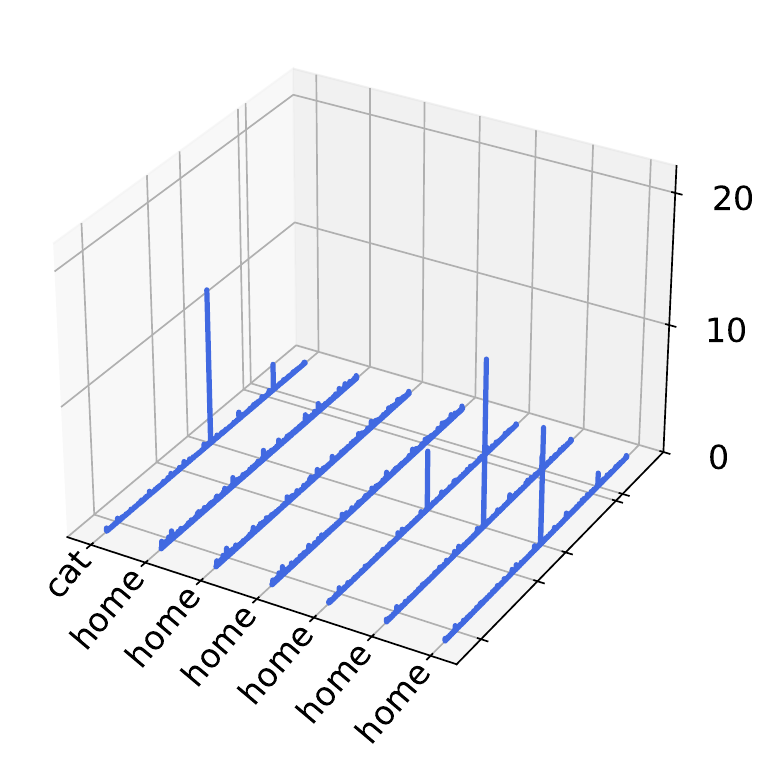}
      \caption{Activation: Distinct token at pos. 1}
      \label{fig:act_diff_first_mistral_s1}
  \end{subfigure}

  \begin{subfigure}[t]{0.48\linewidth}
    \centering
    \includegraphics[width=\linewidth]{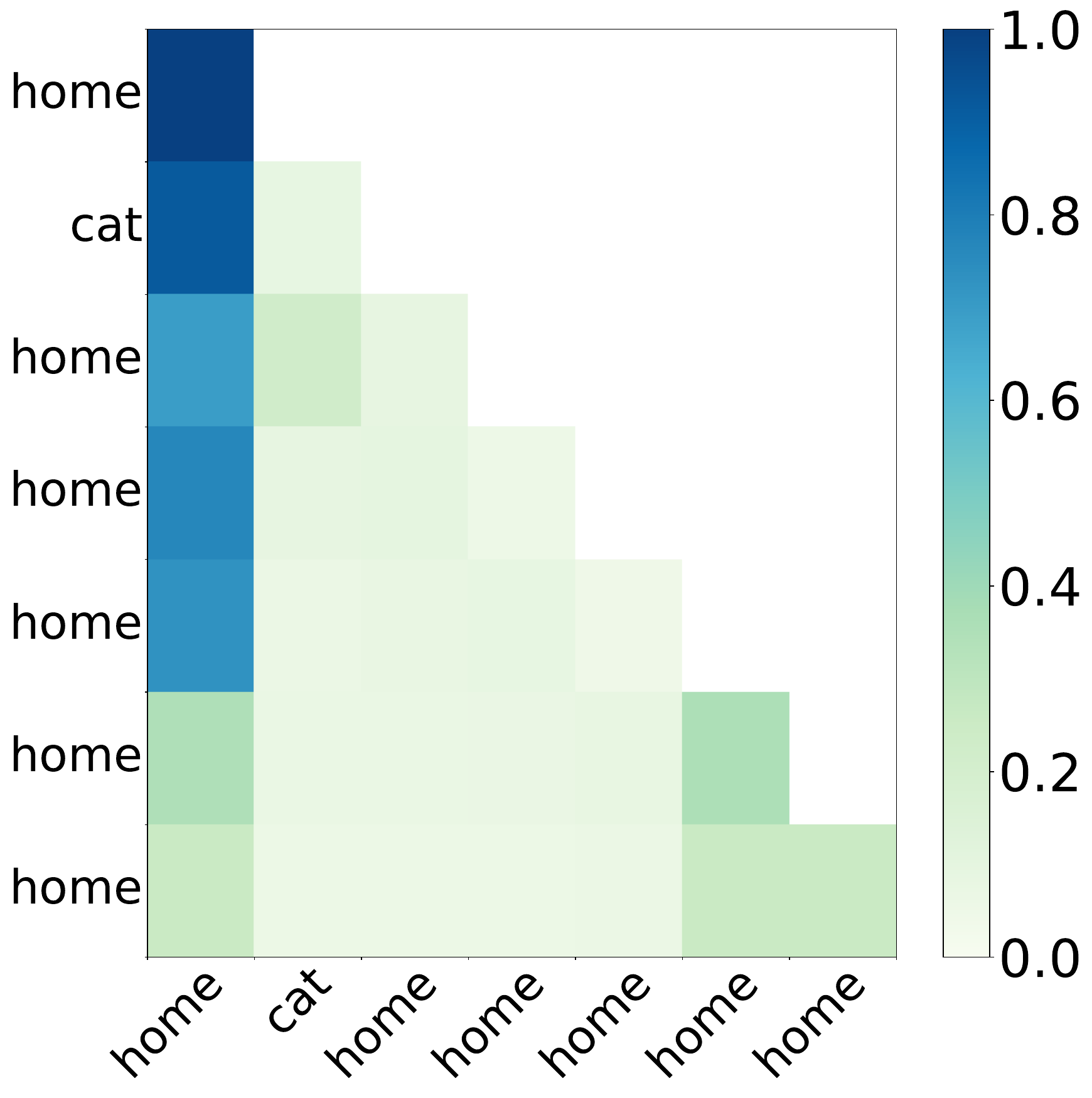}
    \caption{Heatmap: Distinct token at pos. 2}
    \label{fig:heatmap_diff_second_mistral_s1}
  \end{subfigure}
  \begin{subfigure}[t]{0.48\linewidth}
      \centering
      \includegraphics[width=\linewidth]{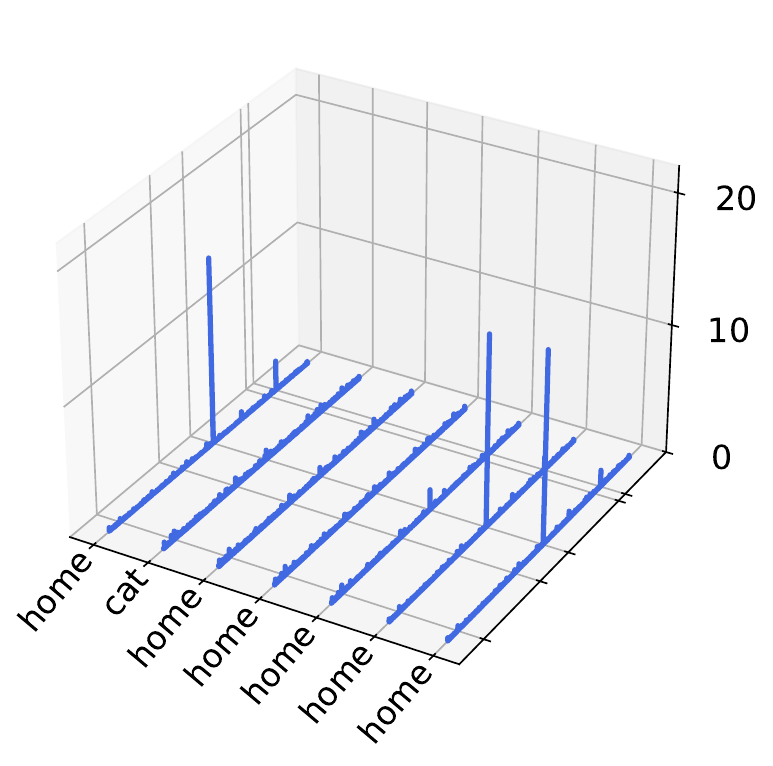}
      \caption{Activation: Distinct token at pos. 2}
      \label{fig:act_diff_second_mistral_s1}
  \end{subfigure}

  \begin{subfigure}[t]{0.48\linewidth}
      \centering
      \includegraphics[width=\linewidth]{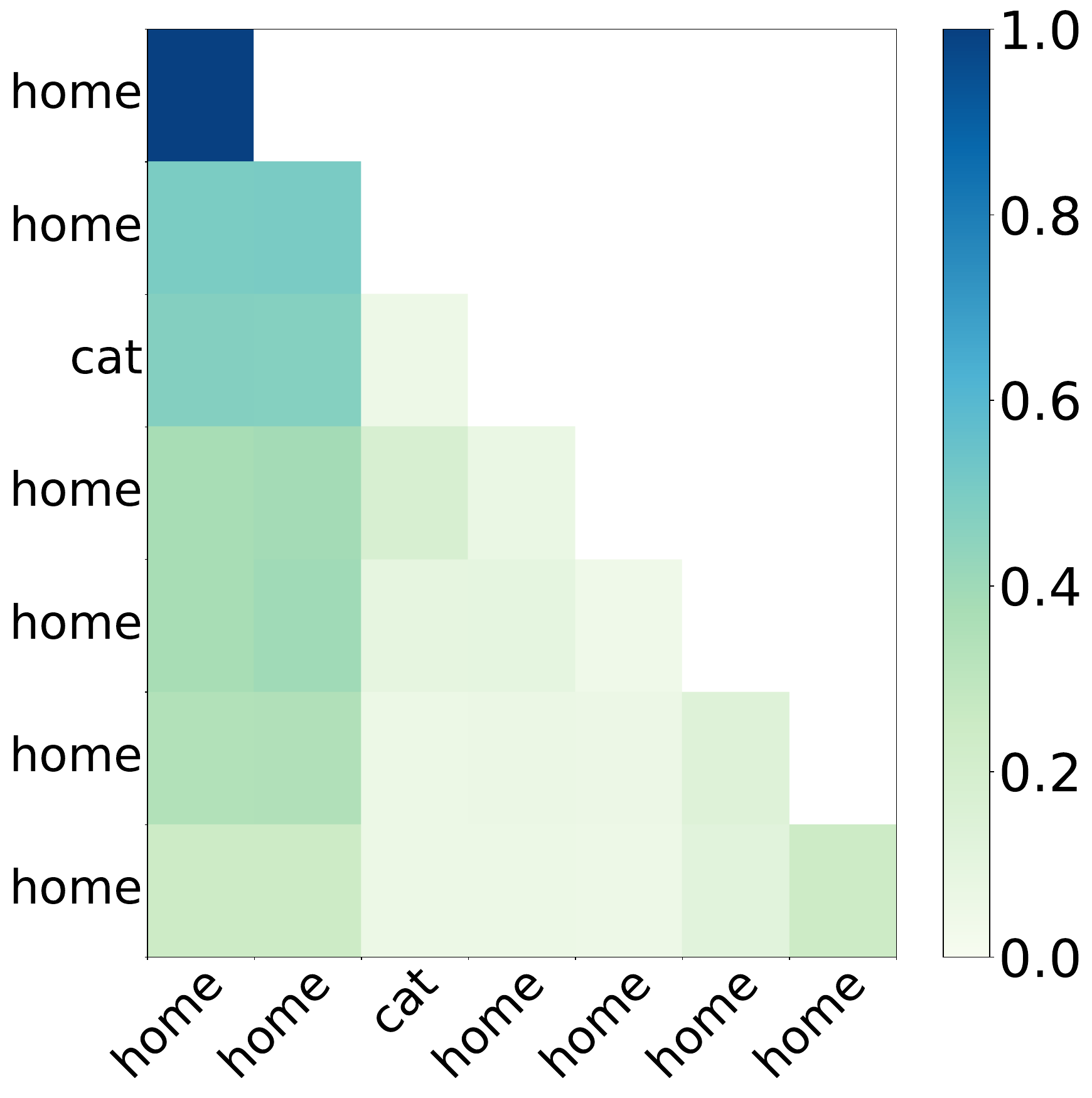}
      \caption{Heatmap: Distinct token at pos. 3}
      \label{fig:heatmap_diff_third_mistral_s1}
  \end{subfigure}
  \begin{subfigure}[t]{0.48\linewidth}
      \centering
      \includegraphics[width=\linewidth]{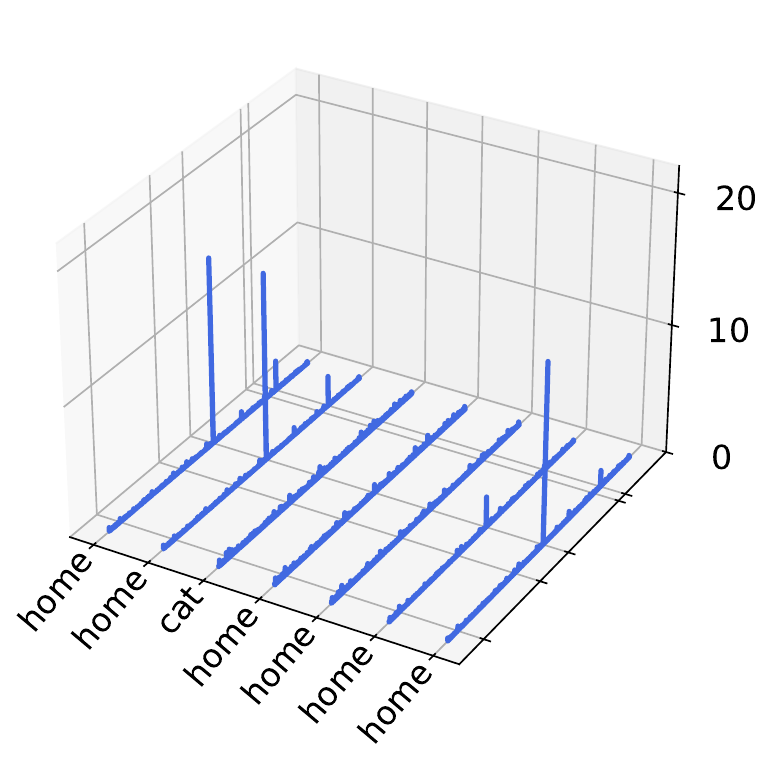}
      \caption{Activation: Distinct token at pos. 3}
      \label{fig:act_diff_third_mistral_s1}
  \end{subfigure}

  \caption{Results for sequences with repeated tokens.
  Each row compares the attention heatmap (left) and hidden-state activations (right) under the same token distribution pattern for Mistral-7B-v0.3 (Sample 1).}
  \label{fig:repeated_tokens_results_mistral_s1}
\end{figure}

\begin{figure*}[t]
  \centering
  \begin{subfigure}[t]{0.48\linewidth}
    \centering
    \includegraphics[width=\linewidth]{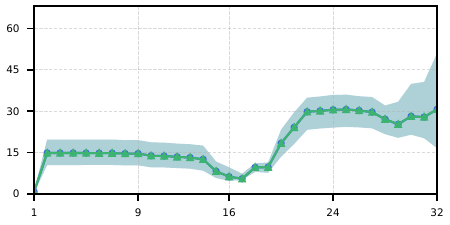}
    \caption{Norms: Uniform}
  \end{subfigure}
  \hfill
  \begin{subfigure}[t]{0.48\linewidth}
    \centering
    \includegraphics[width=\linewidth]{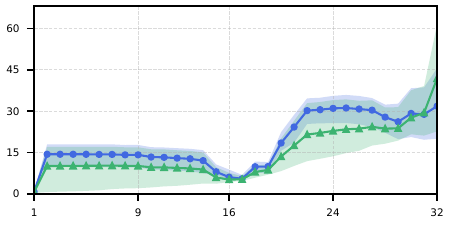}
    \caption{Norms: Pos. 1}
  \end{subfigure}

  \begin{subfigure}[t]{0.48\linewidth}
    \centering
    \includegraphics[width=\linewidth]{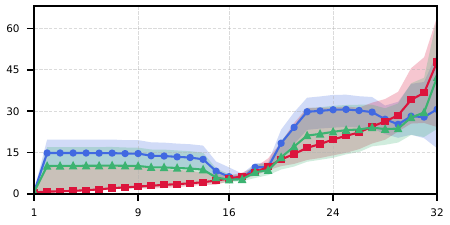}
    \caption{Norms: Pos. 2}
  \end{subfigure}
  \hfill
  \begin{subfigure}[t]{0.48\linewidth}
    \centering
    \includegraphics[width=\linewidth]{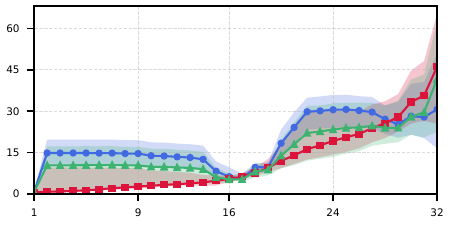}
    \caption{Norms: Pos. 3}
  \end{subfigure}
  \caption{Layer-wise averaged hidden-state norms in Mistral-7B-v0.3 for repeated-token sequences under different distinct-token positions (colors and markers as in \Cref{fig:activation_norms_repeated_tokens_llama3}).
  The horizontal axis shows the layer index in every panel.}
  \label{fig:activation_norms_repeated_tokens_mistral}
\end{figure*}

\begin{figure}[t]
  \centering
  \begin{subfigure}[t]{0.48\linewidth}
    \centering
    \includegraphics[width=\linewidth]{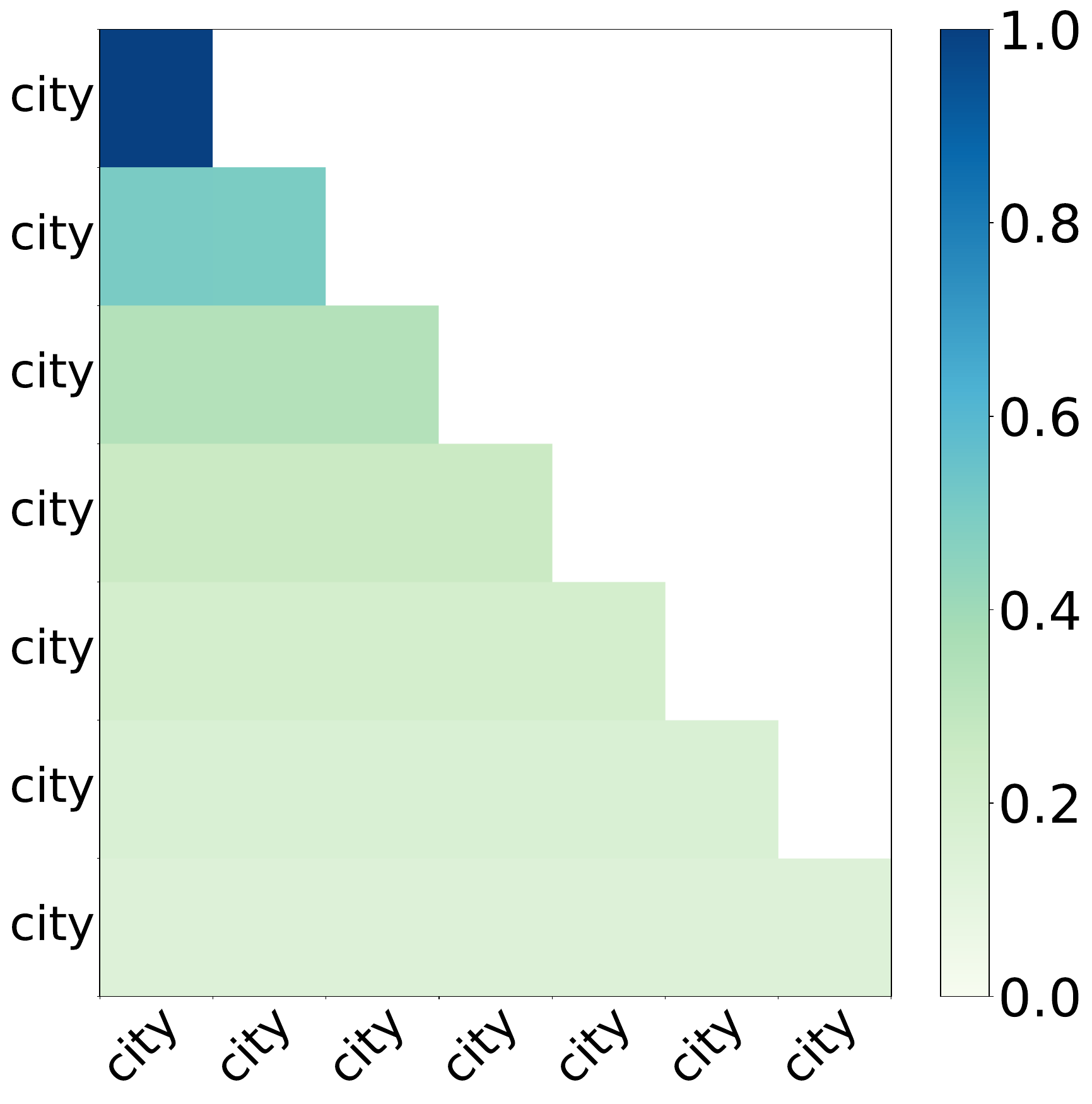}
    \caption{Heatmap: Uniform sequence}
    \label{fig:heatmap_all_same_qwen_s2}
  \end{subfigure}
  \begin{subfigure}[t]{0.48\linewidth}
      \centering
      \includegraphics[width=\linewidth]{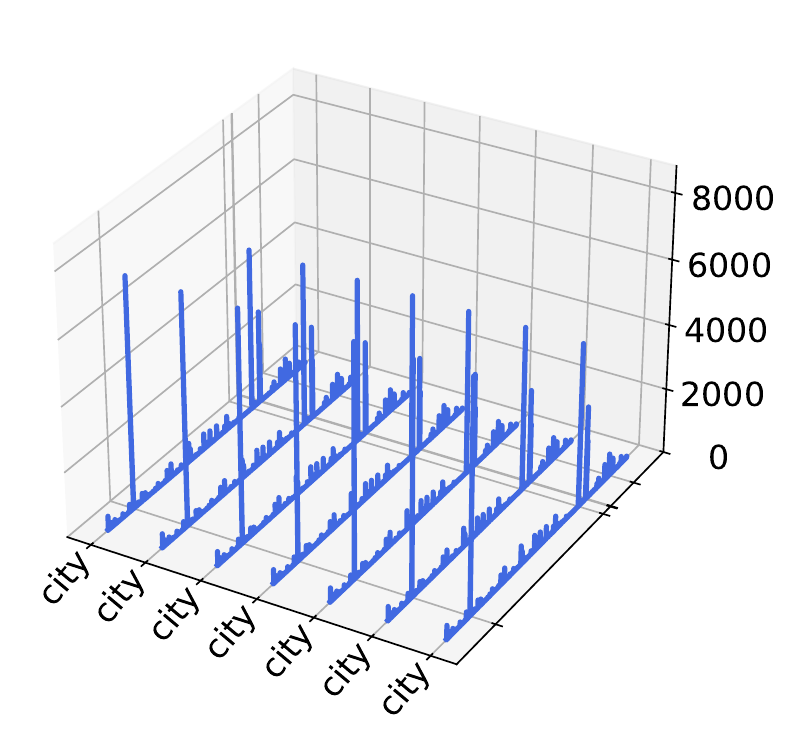}
      \caption{Activation: Uniform sequence}
      \label{fig:act_all_same_qwen_s2}
  \end{subfigure}

  \begin{subfigure}[t]{0.48\linewidth}
    \centering
    \includegraphics[width=\linewidth]{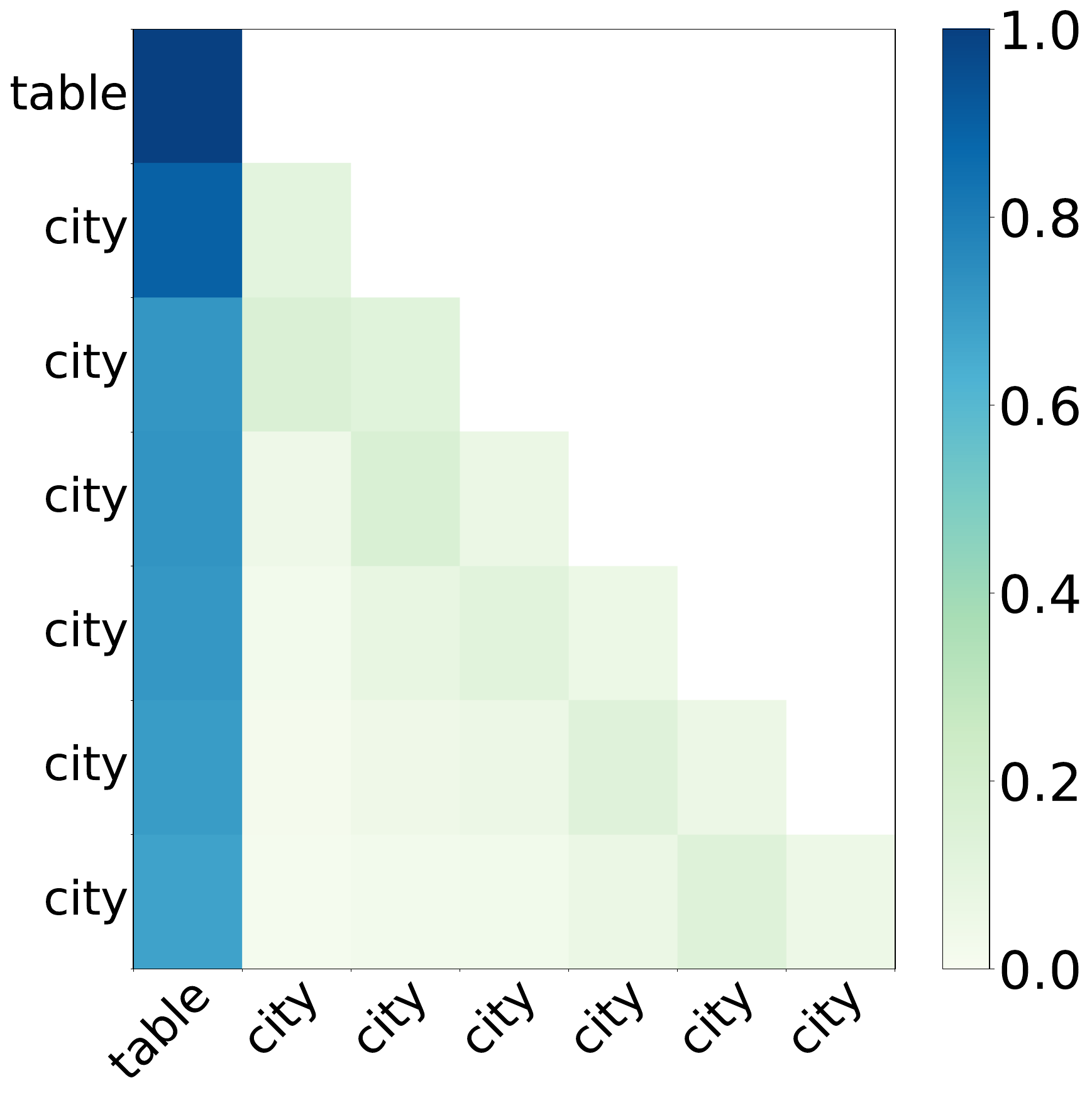}
    \caption{Heatmap: Distinct token at pos. 1}
    \label{fig:heatmap_diff_first_qwen_s2}
  \end{subfigure}
  \begin{subfigure}[t]{0.48\linewidth}
      \centering
      \includegraphics[width=\linewidth]{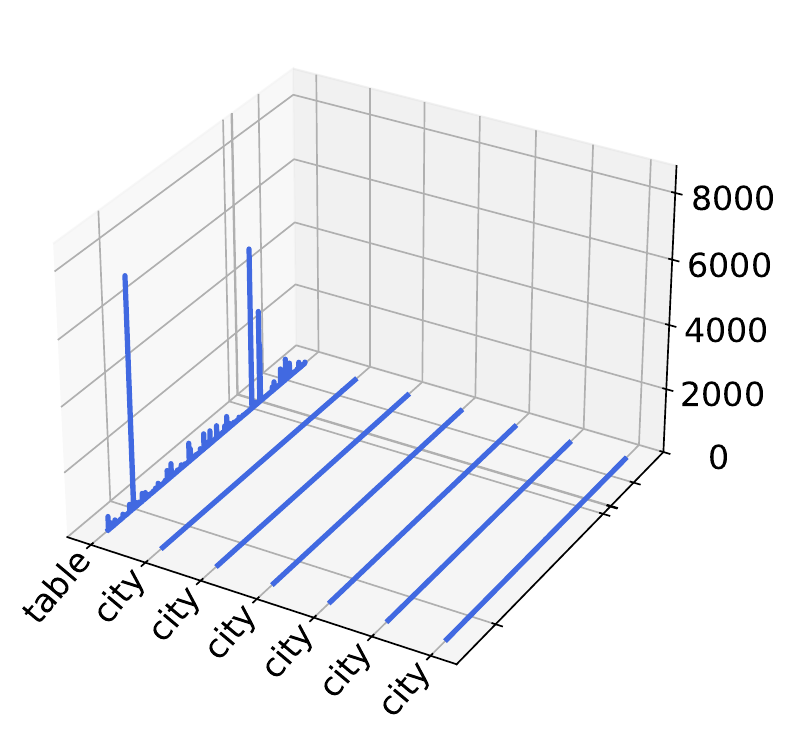}
      \caption{Activation: Distinct token at pos. 1}
      \label{fig:act_diff_first_qwen_s2}
  \end{subfigure}

  \begin{subfigure}[t]{0.48\linewidth}
    \centering
    \includegraphics[width=\linewidth]{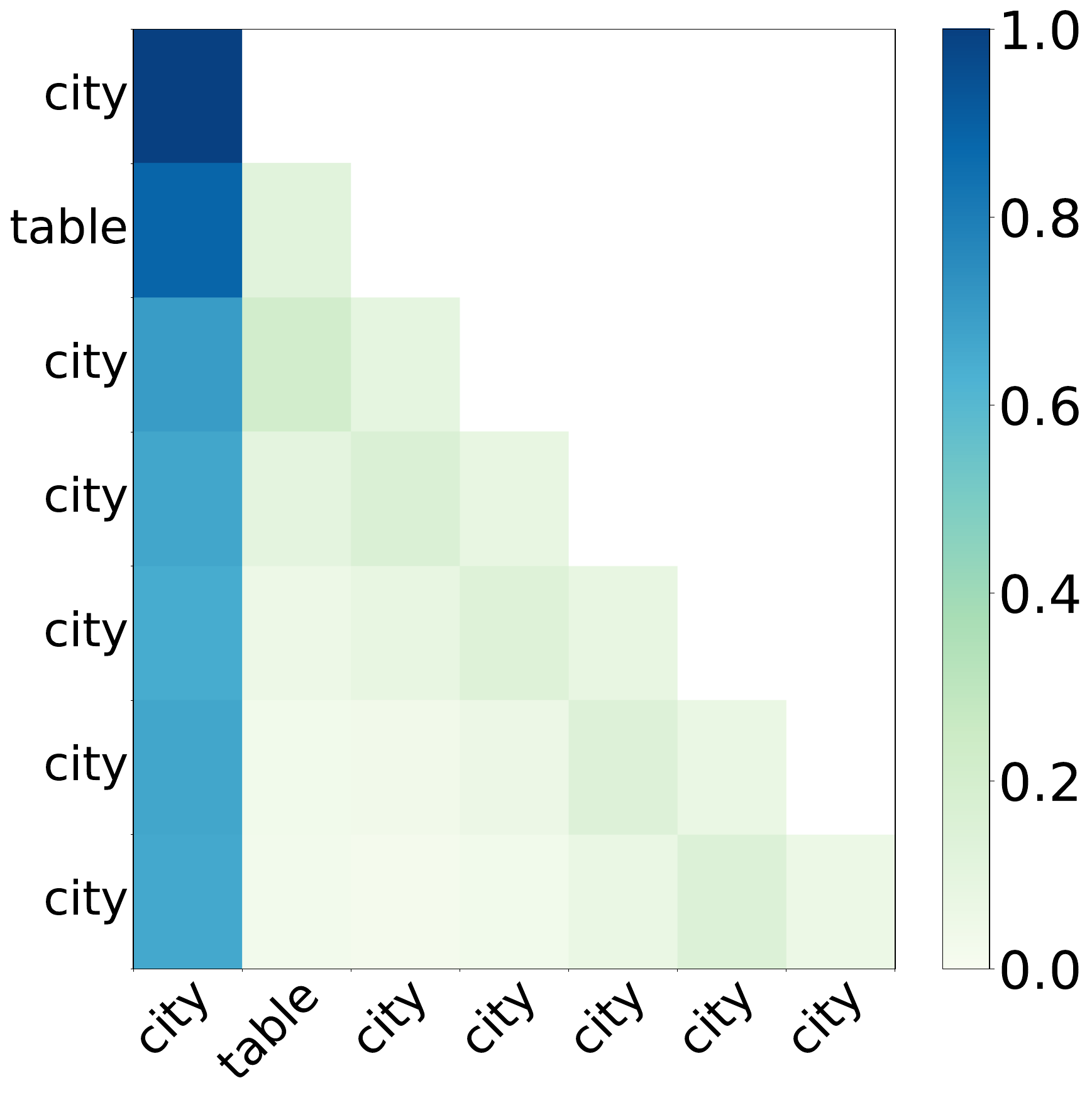}
    \caption{Heatmap: Distinct token at pos. 2}
    \label{fig:heatmap_diff_second_qwen_s2}
  \end{subfigure}
  \begin{subfigure}[t]{0.48\linewidth}
      \centering
      \includegraphics[width=\linewidth]{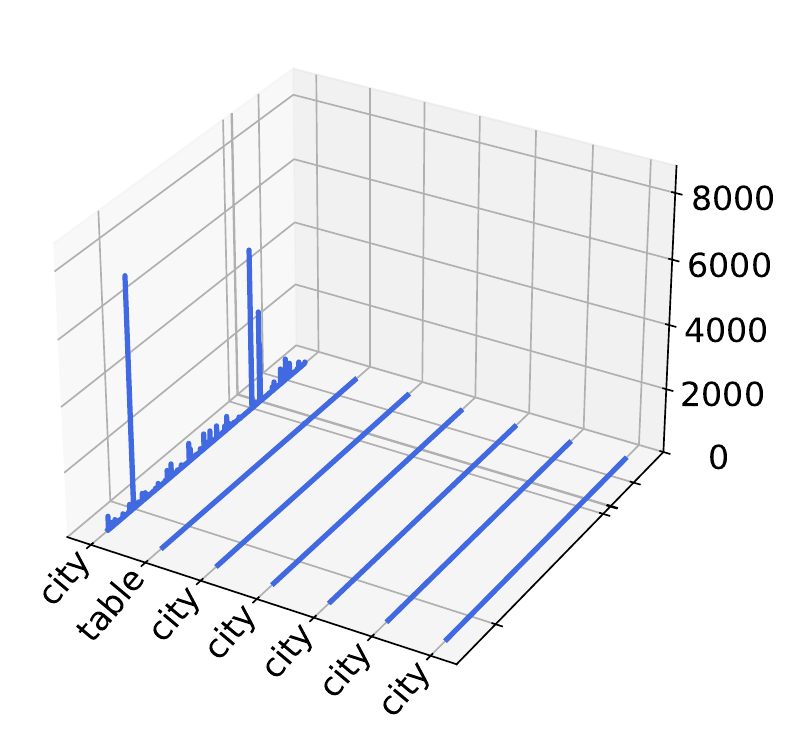}
      \caption{Activation: Distinct token at pos. 2}
      \label{fig:act_diff_second_qwen_s2}
  \end{subfigure}

  \begin{subfigure}[t]{0.48\linewidth}
      \centering
      \includegraphics[width=\linewidth]{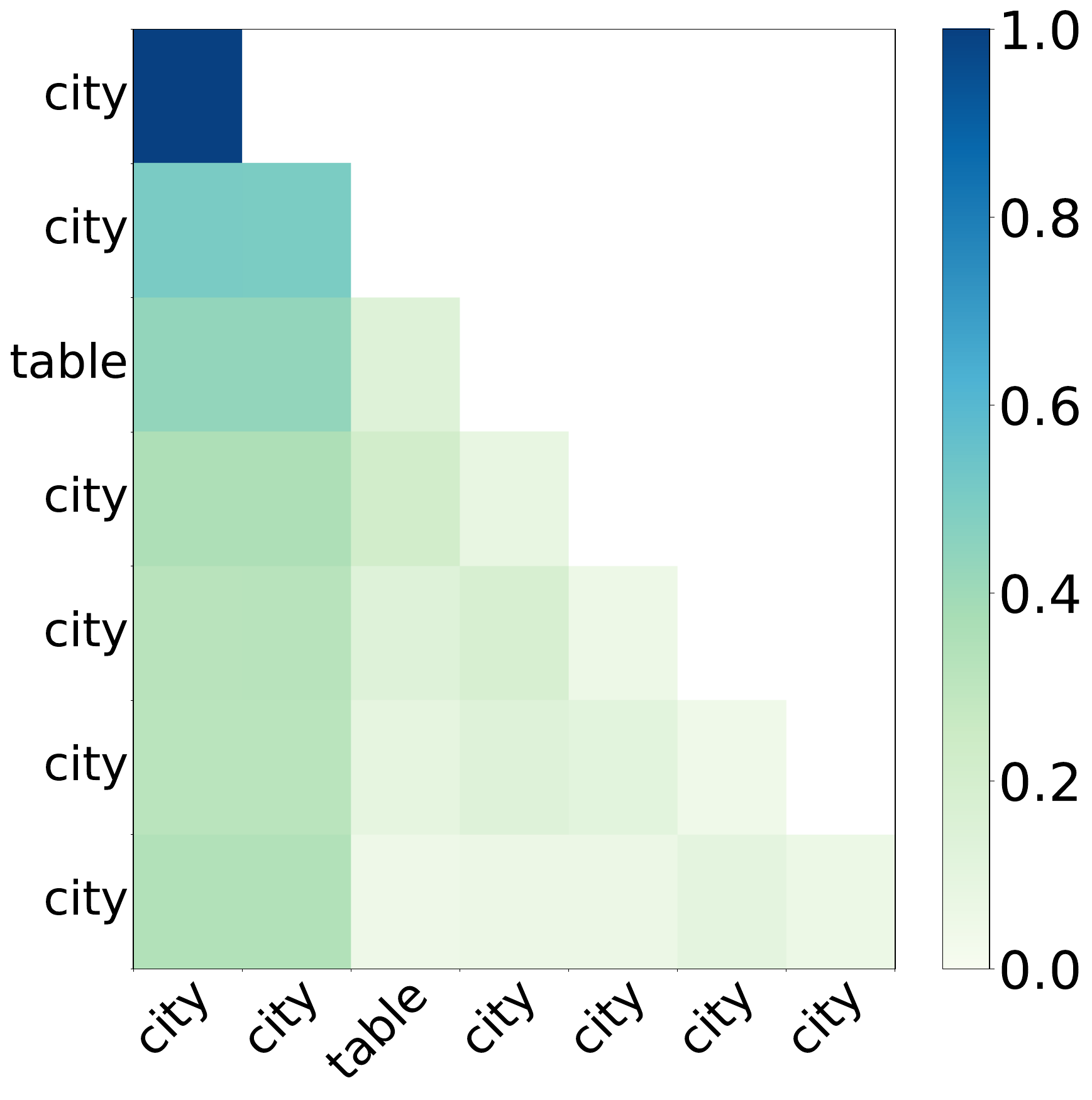}
      \caption{Heatmap: Distinct token at pos. 3}
      \label{fig:heatmap_diff_third_qwen_s2}
  \end{subfigure}
  \begin{subfigure}[t]{0.48\linewidth}
      \centering
      \includegraphics[width=\linewidth]{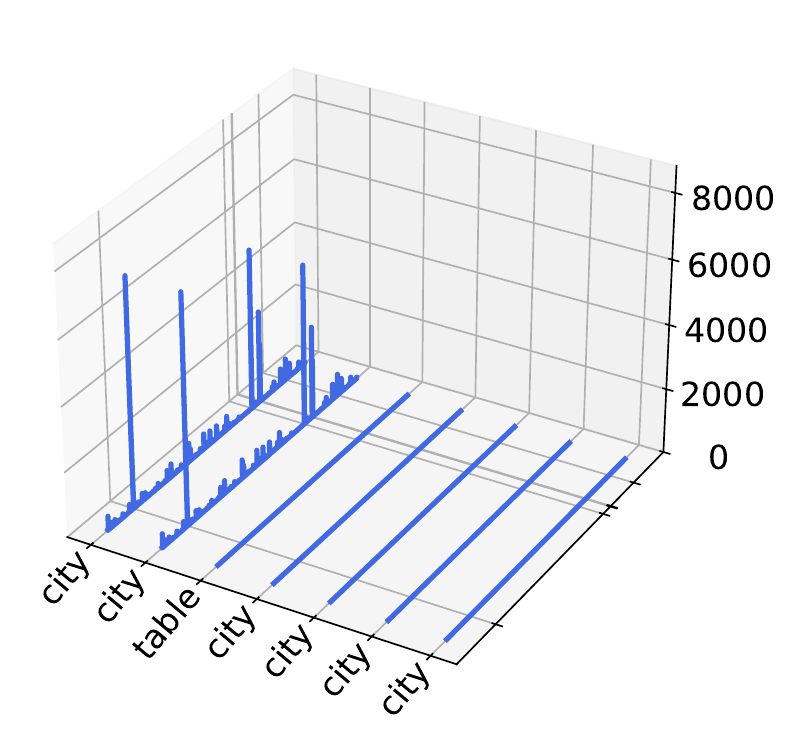}
      \caption{Activation: Distinct token at pos. 3}
      \label{fig:act_diff_third_qwen_s2}
  \end{subfigure}

  \caption{Results for sequences with repeated tokens.
  Each row compares the attention heatmap (left) and hidden-state activations (right) under the same token distribution pattern for Qwen2-7B (Sample 2).}
  \label{fig:repeated_tokens_results_qwen_s2}
\end{figure}

\begin{figure}[t]
  \centering
  \begin{subfigure}[t]{0.48\linewidth}
    \centering
    \includegraphics[width=\linewidth]{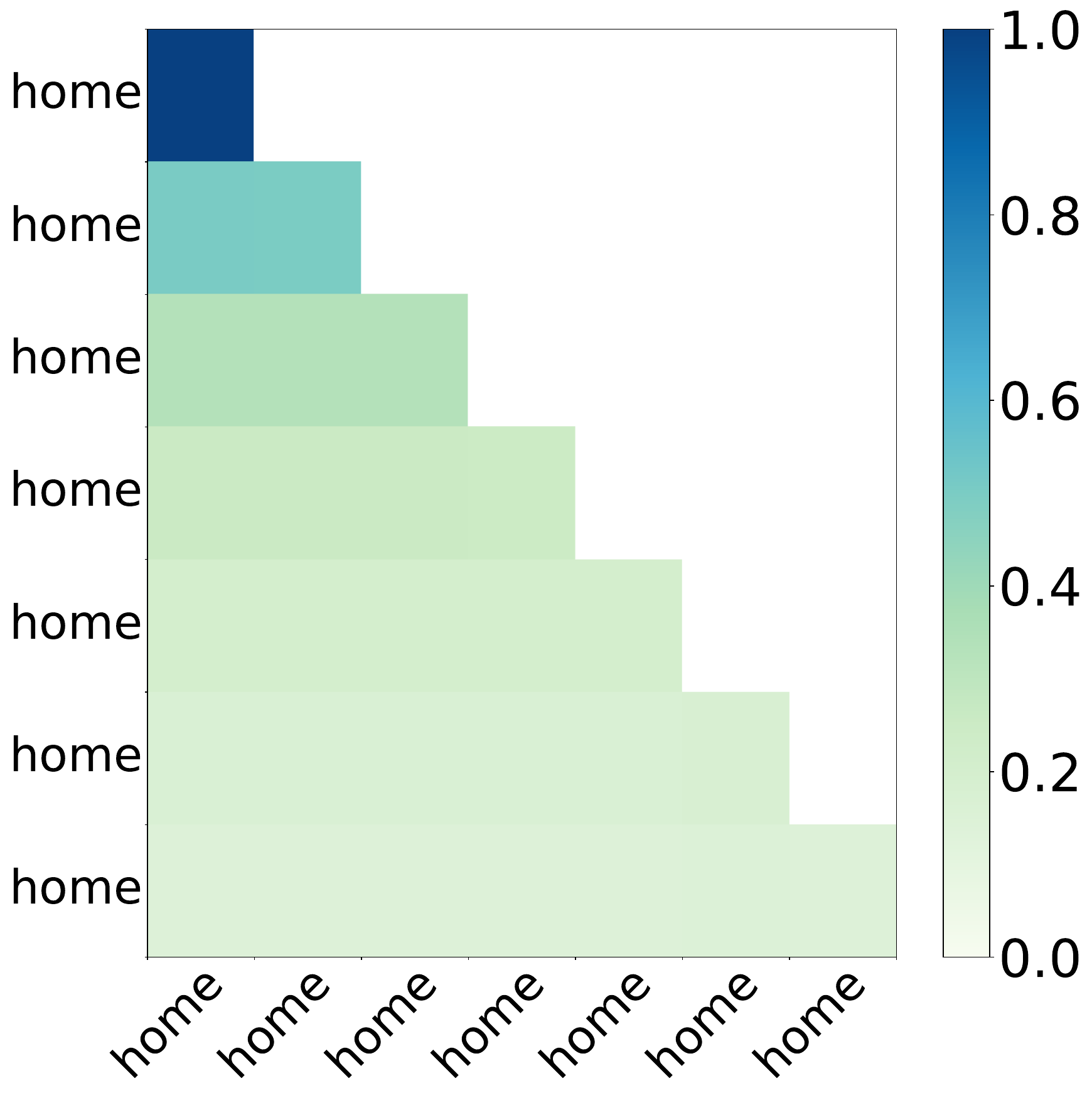}
    \caption{Heatmap: Uniform sequence}
    \label{fig:heatmap_all_same_qwen_s1}
  \end{subfigure}
  \begin{subfigure}[t]{0.48\linewidth}
      \centering
      \includegraphics[width=\linewidth]{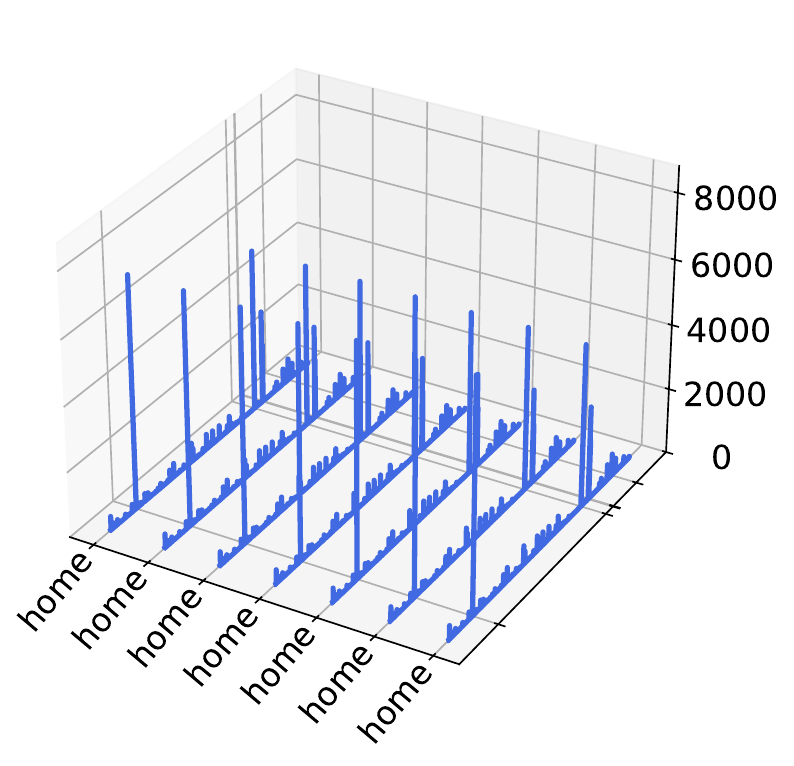}
      \caption{Activation: Uniform sequence}
      \label{fig:act_all_same_qwen_s1}
  \end{subfigure}

  \begin{subfigure}[t]{0.48\linewidth}
    \centering
    \includegraphics[width=\linewidth]{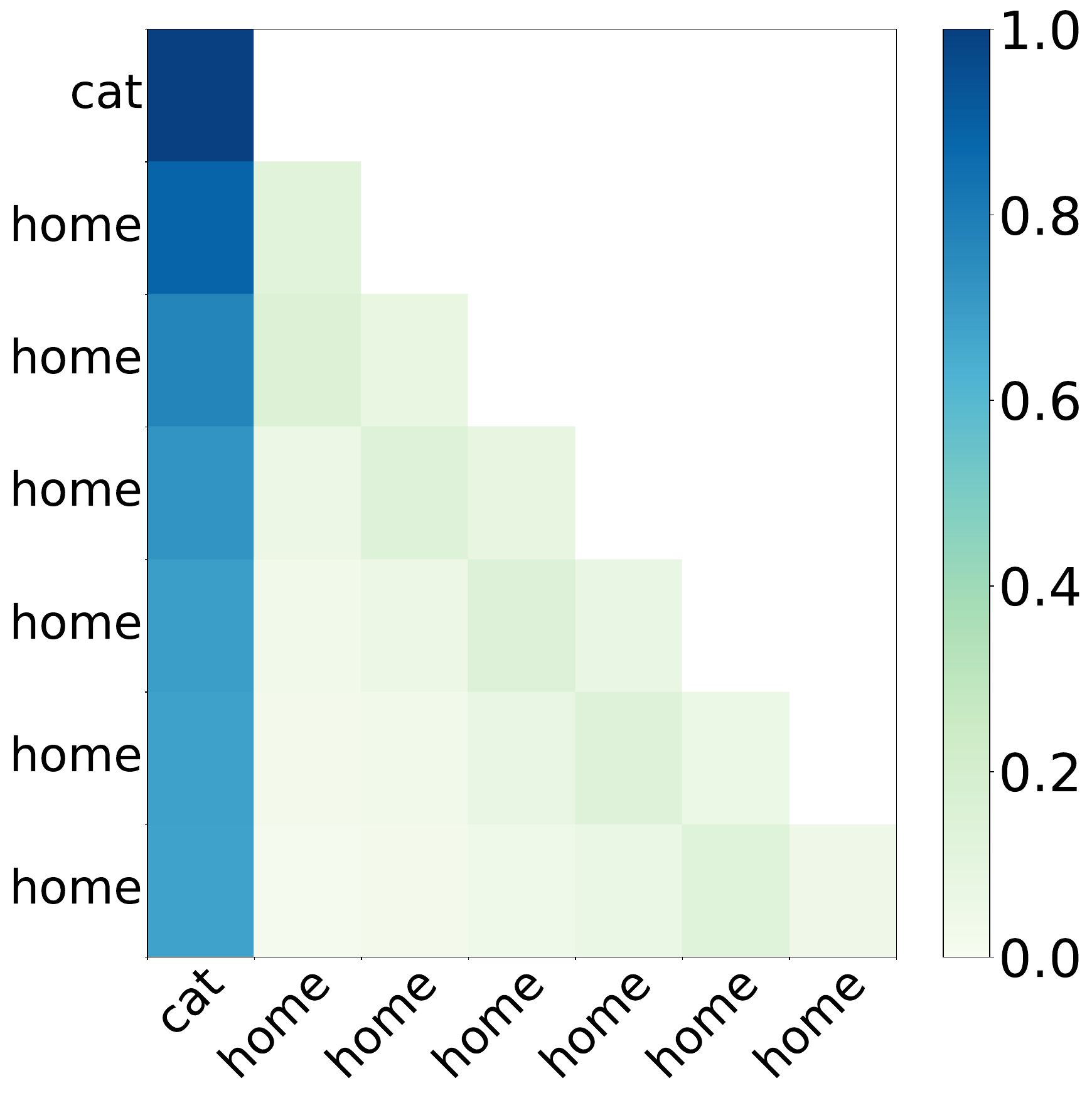}
    \caption{Heatmap: Distinct token at pos. 1}
    \label{fig:heatmap_diff_first_qwen_s1}
  \end{subfigure}
  \begin{subfigure}[t]{0.48\linewidth}
      \centering
      \includegraphics[width=\linewidth]{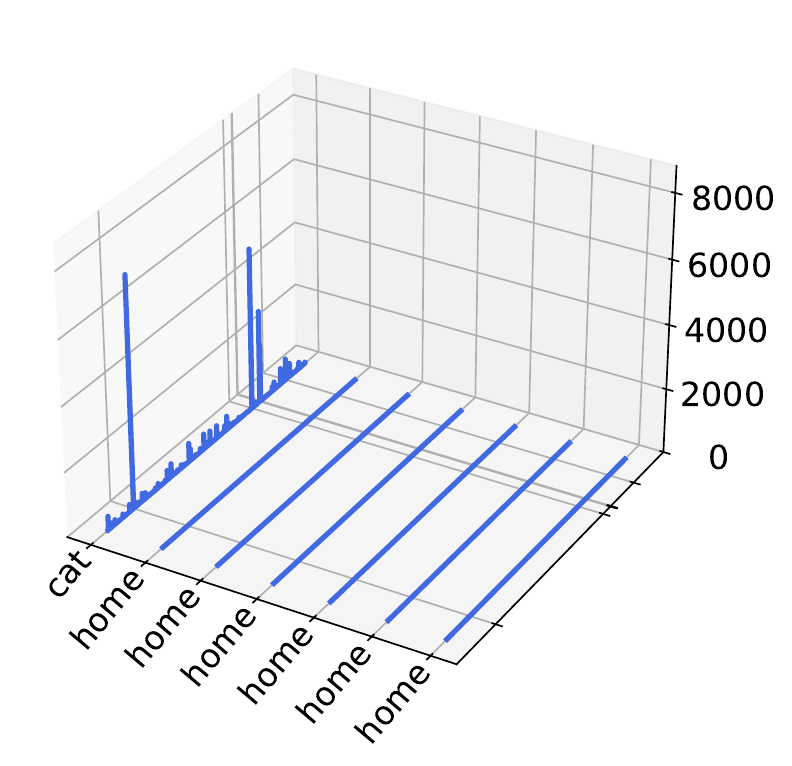}
      \caption{Activation: Distinct token at pos. 1}
      \label{fig:act_diff_first_qwen_s1}
  \end{subfigure}

  \begin{subfigure}[t]{0.48\linewidth}
    \centering
    \includegraphics[width=\linewidth]{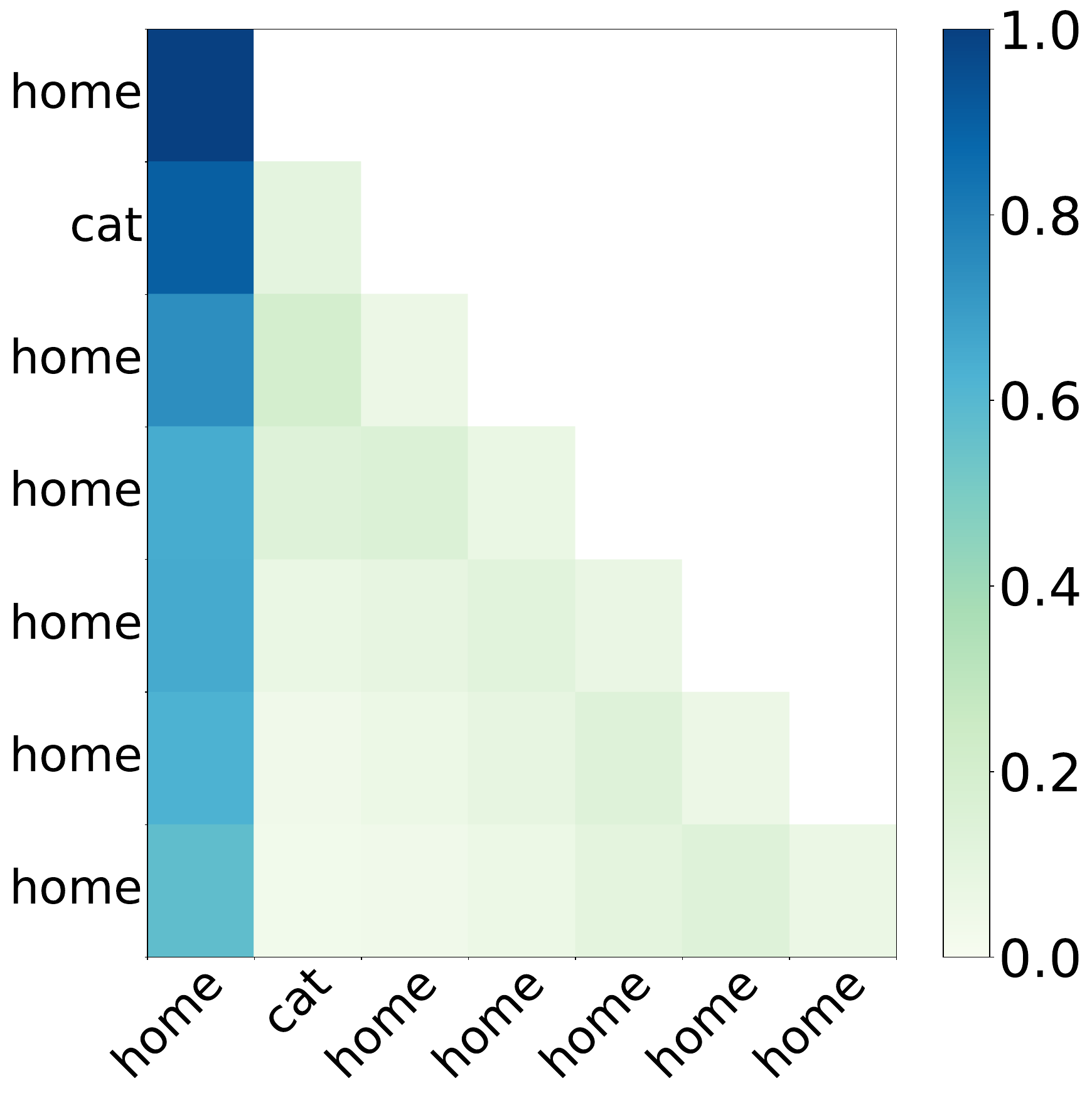}
    \caption{Heatmap: Distinct token at pos. 2}
    \label{fig:heatmap_diff_second_qwen_s1}
  \end{subfigure}
  \begin{subfigure}[t]{0.48\linewidth}
      \centering
      \includegraphics[width=\linewidth]{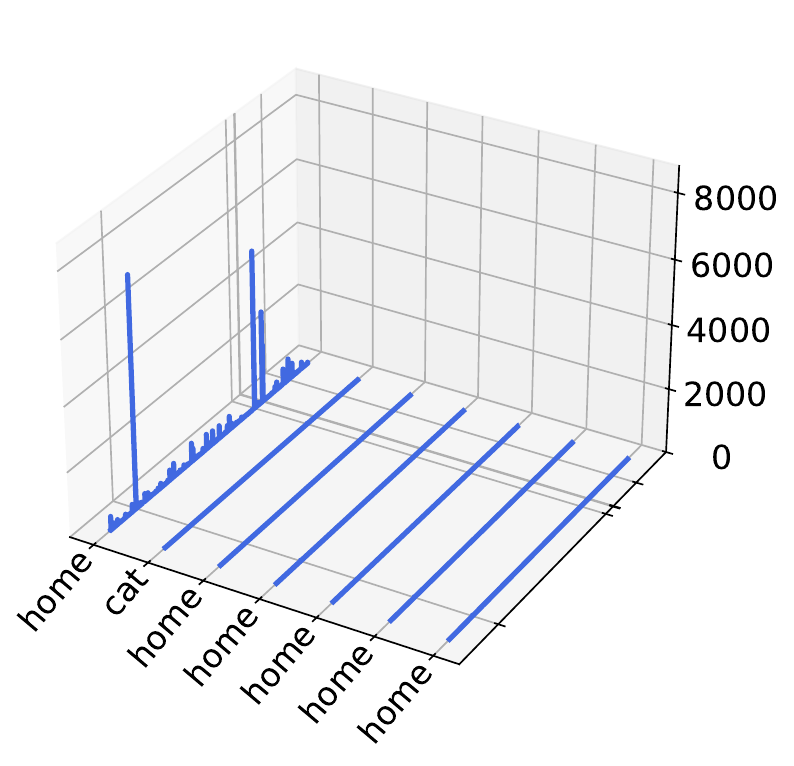}
      \caption{Activation: Distinct token at pos. 2}
      \label{fig:act_diff_second_qwen_s1}
  \end{subfigure}

  \begin{subfigure}[t]{0.48\linewidth}
      \centering
      \includegraphics[width=\linewidth]{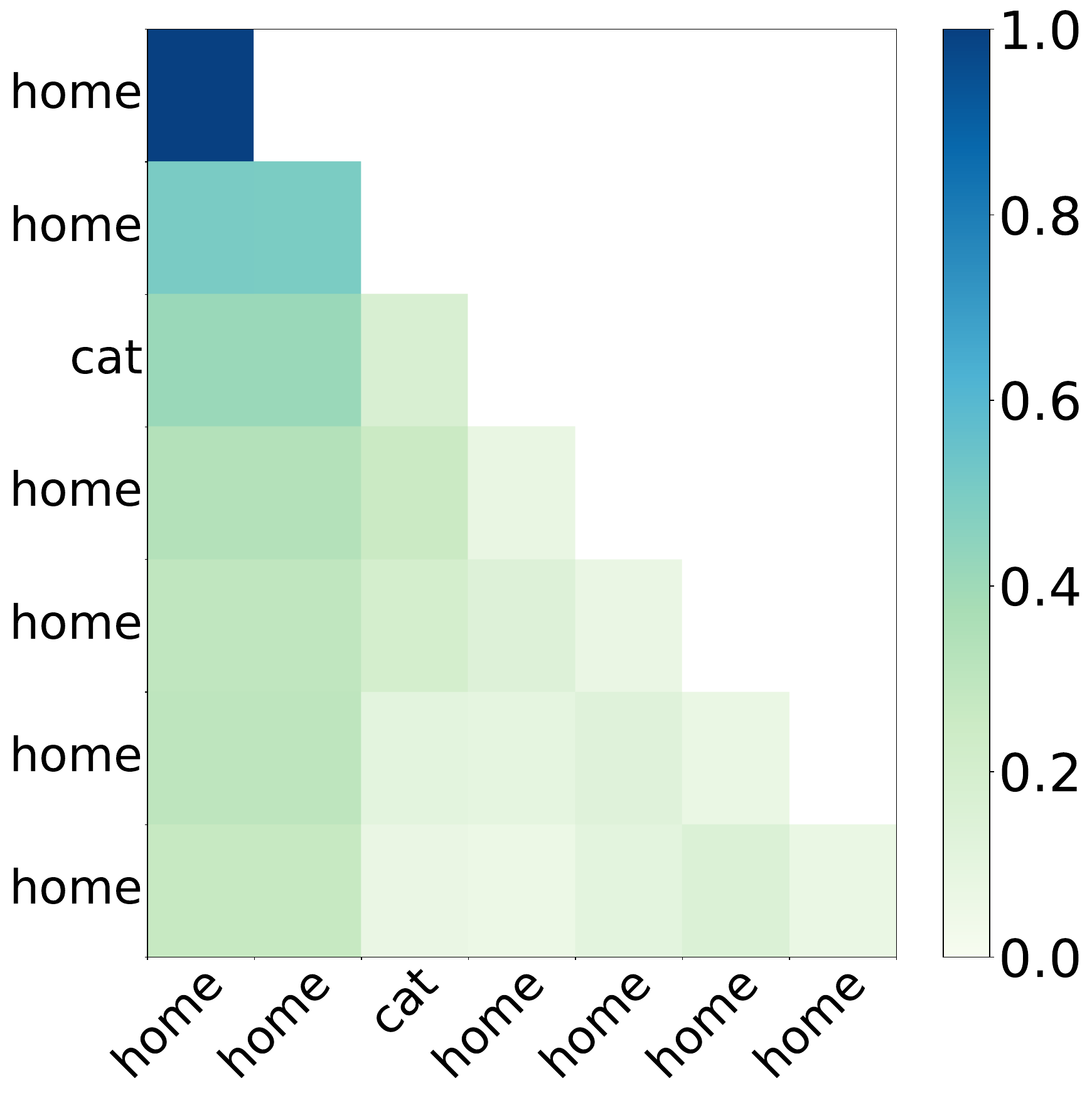}
      \caption{Heatmap: Distinct token at pos. 3}
      \label{fig:heatmap_diff_third_qwen_s1}
  \end{subfigure}
  \begin{subfigure}[t]{0.48\linewidth}
      \centering
      \includegraphics[width=\linewidth]{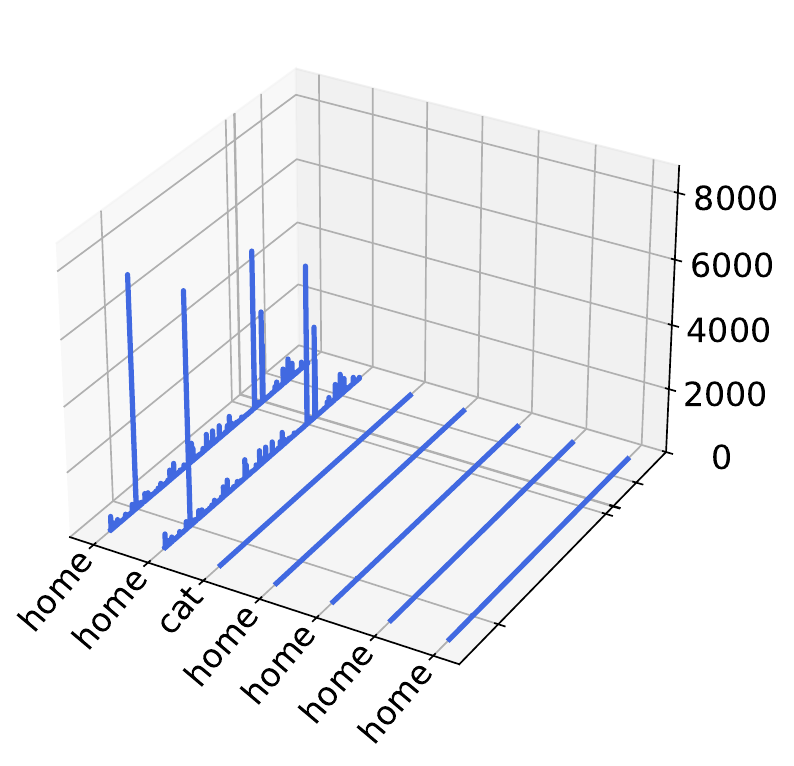}
      \caption{Activation: Distinct token at pos. 3}
      \label{fig:act_diff_third_qwen_s1}
  \end{subfigure}

  \caption{Results for sequences with repeated tokens.
  Each row compares the attention heatmap (left) and hidden-state activations (right) under the same token distribution pattern for Qwen2-7B (Sample 1).}
  \label{fig:repeated_tokens_results_qwen_s1}
\end{figure}

\begin{figure*}[t]
  \centering
  \begin{subfigure}[t]{0.48\linewidth}
    \centering
    \includegraphics[width=\linewidth]{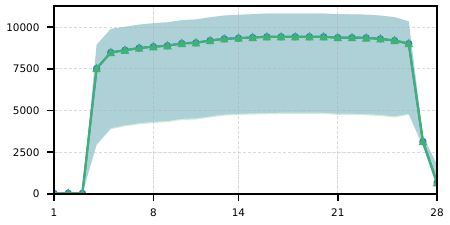}
    \caption{Norms: Uniform}
  \end{subfigure}
  \hfill
  \begin{subfigure}[t]{0.48\linewidth}
    \centering
    \includegraphics[width=\linewidth]{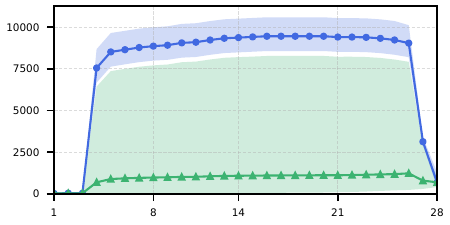}
    \caption{Norms: Pos. 1}
  \end{subfigure}

  \begin{subfigure}[t]{0.48\linewidth}
    \centering
    \includegraphics[width=\linewidth]{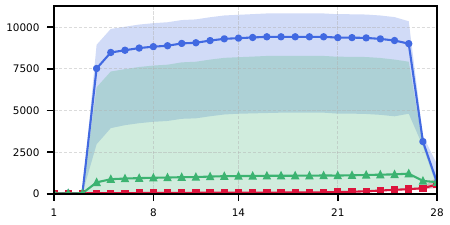}
    \caption{Norms: Pos. 2}
  \end{subfigure}
  \hfill
  \begin{subfigure}[t]{0.48\linewidth}
    \centering
    \includegraphics[width=\linewidth]{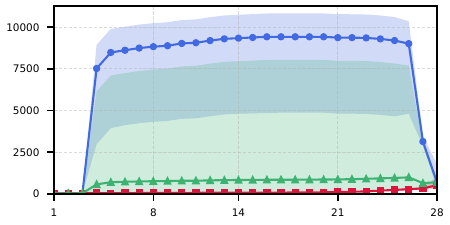}
    \caption{Norms: Pos. 3}
  \end{subfigure}
  \caption{Layer-wise averaged hidden-state norms in Qwen2-7B for repeated-token sequences under different distinct-token positions (colors and markers as in \Cref{fig:activation_norms_repeated_tokens_llama3}).
  The horizontal axis shows the layer index in every panel.}
  \label{fig:activation_norms_repeated_tokens_qwen}
\end{figure*}

\begin{figure}[t]
  \centering
  \begin{subfigure}[t]{0.48\linewidth}
    \centering
    \includegraphics[width=\linewidth]{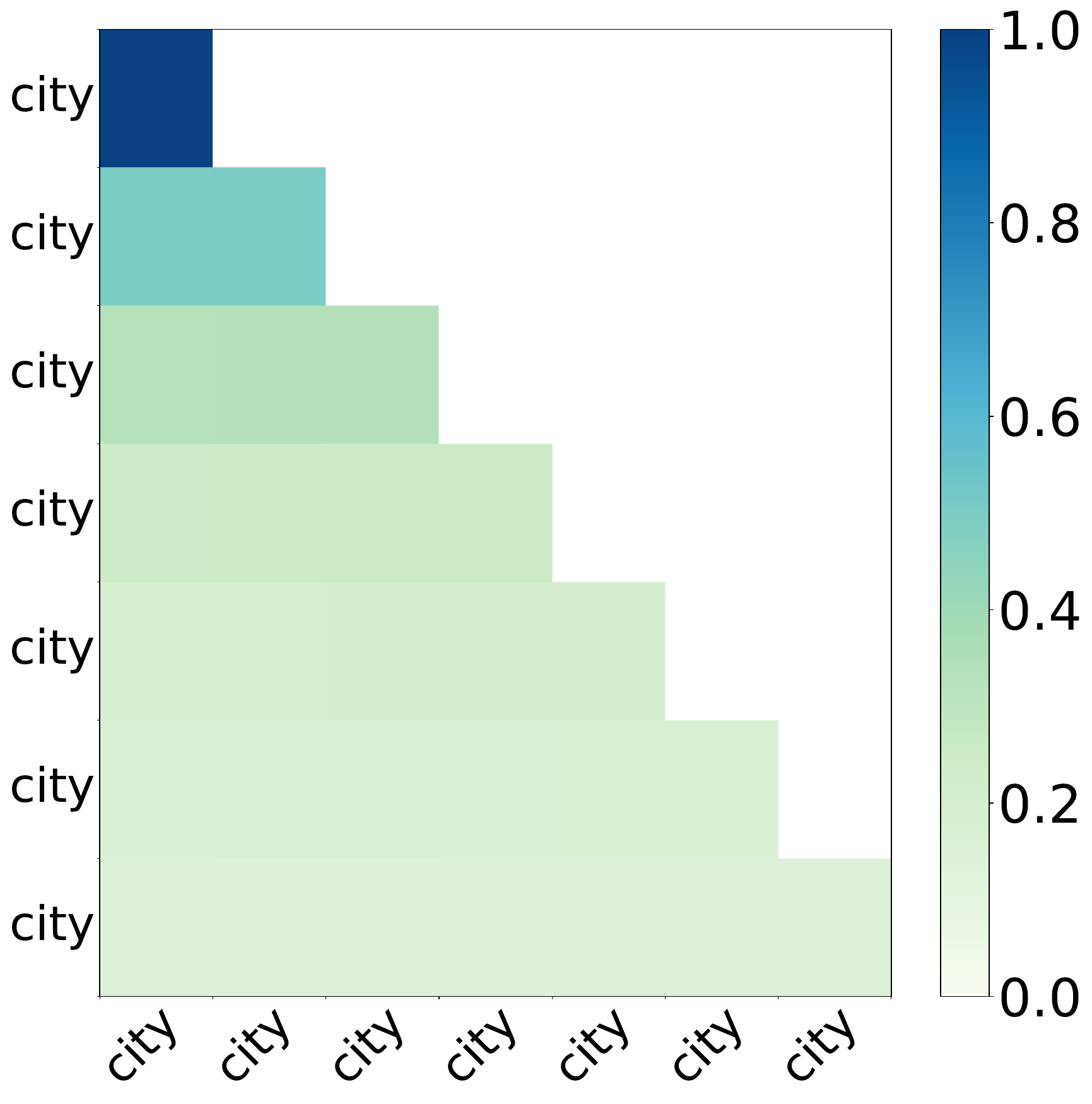}
    \caption{Heatmap: Uniform sequence}
    \label{fig:heatmap_all_same_pythia_s2}
  \end{subfigure}
  \begin{subfigure}[t]{0.48\linewidth}
      \centering
      \includegraphics[width=\linewidth]{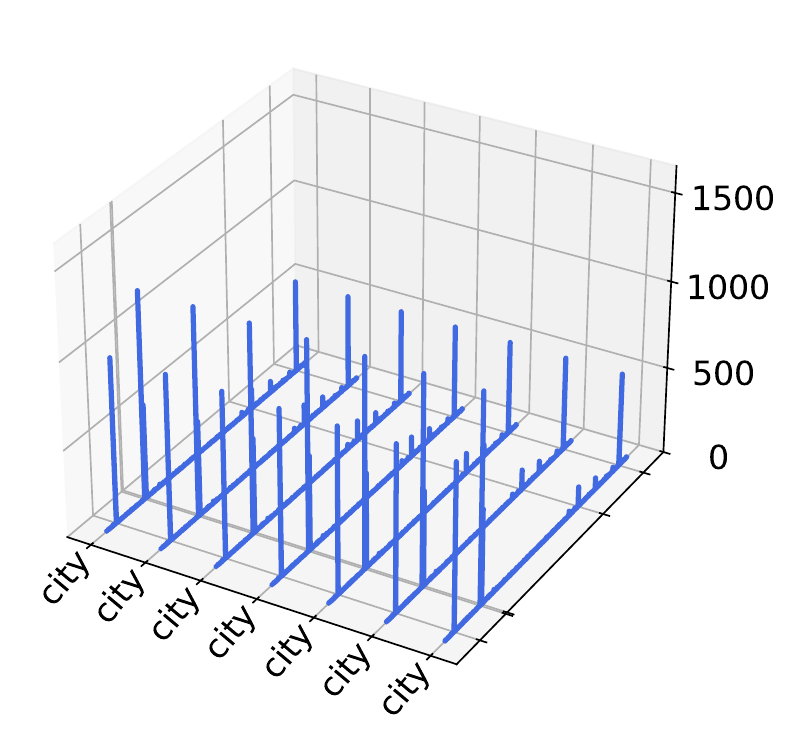}
      \caption{Activation: Uniform sequence}
      \label{fig:act_all_same_pythia_s2}
  \end{subfigure}

  \begin{subfigure}[t]{0.48\linewidth}
    \centering
    \includegraphics[width=\linewidth]{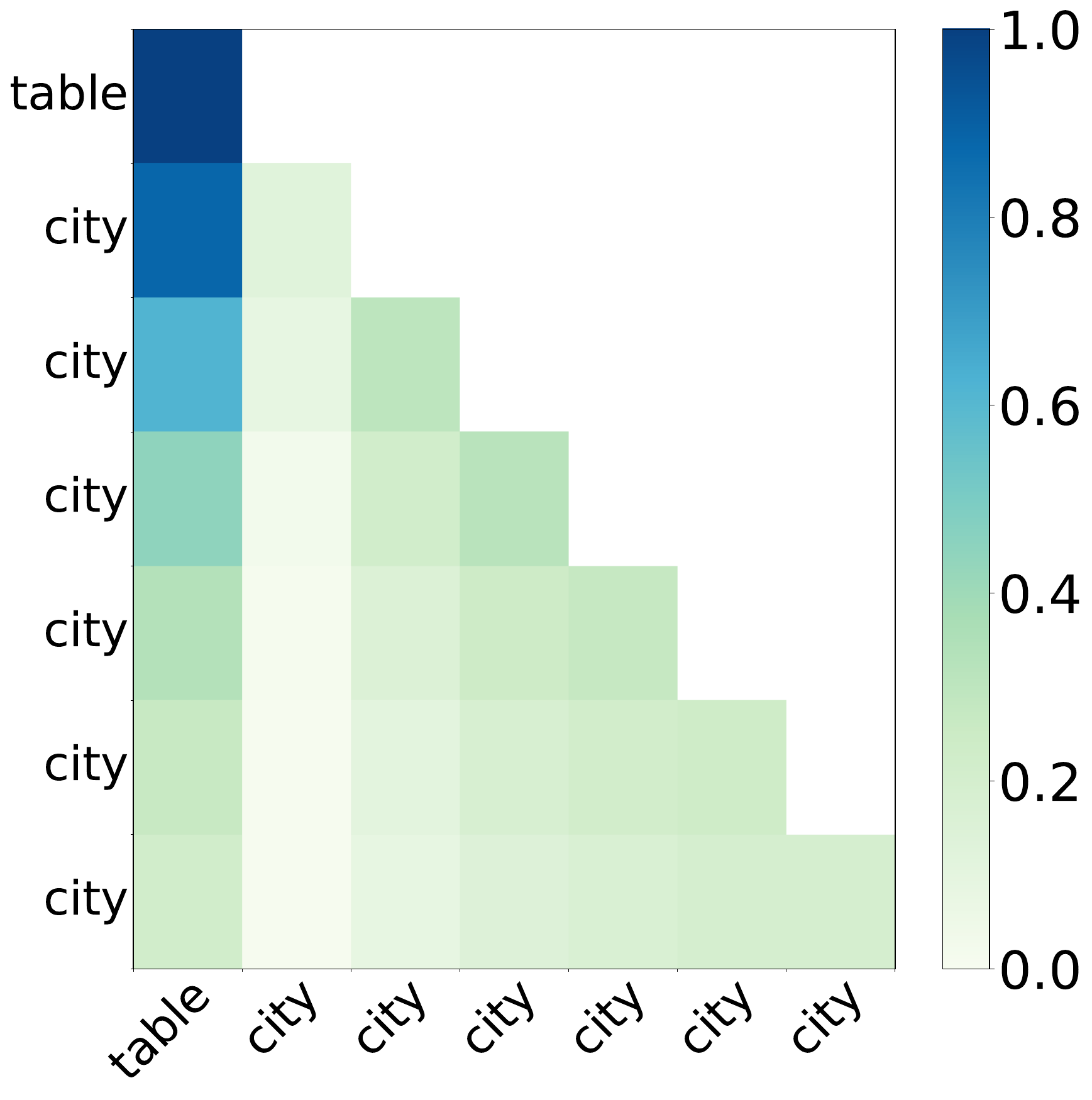}
    \caption{Heatmap: Distinct token at pos. 1}
    \label{fig:heatmap_diff_first_pythia_s2}
  \end{subfigure}
  \begin{subfigure}[t]{0.48\linewidth}
      \centering
      \includegraphics[width=\linewidth]{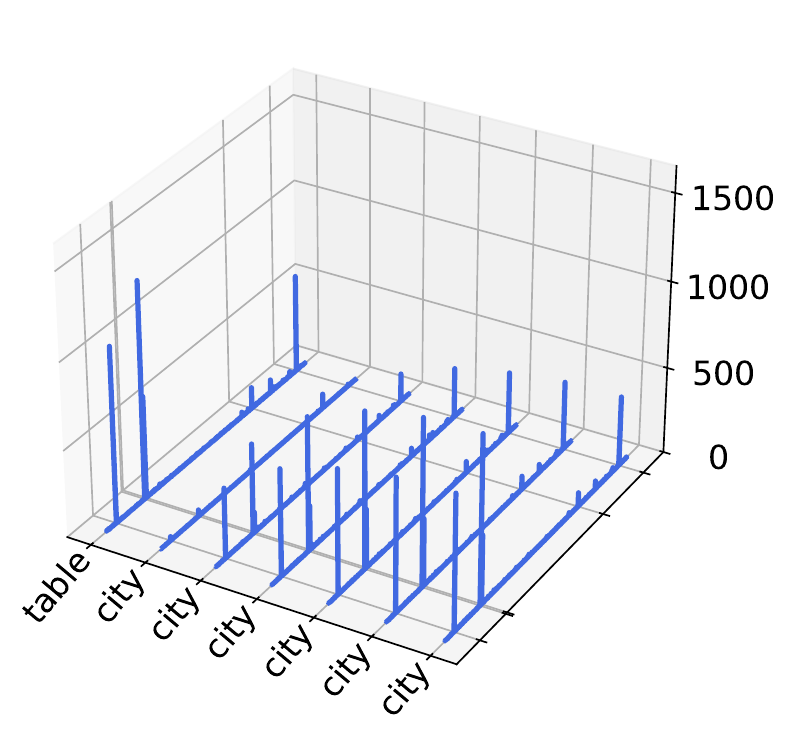}
      \caption{Activation: Distinct token at pos. 1}
      \label{fig:act_diff_first_pythia_s2}
  \end{subfigure}

  \begin{subfigure}[t]{0.48\linewidth}
    \centering
    \includegraphics[width=\linewidth]{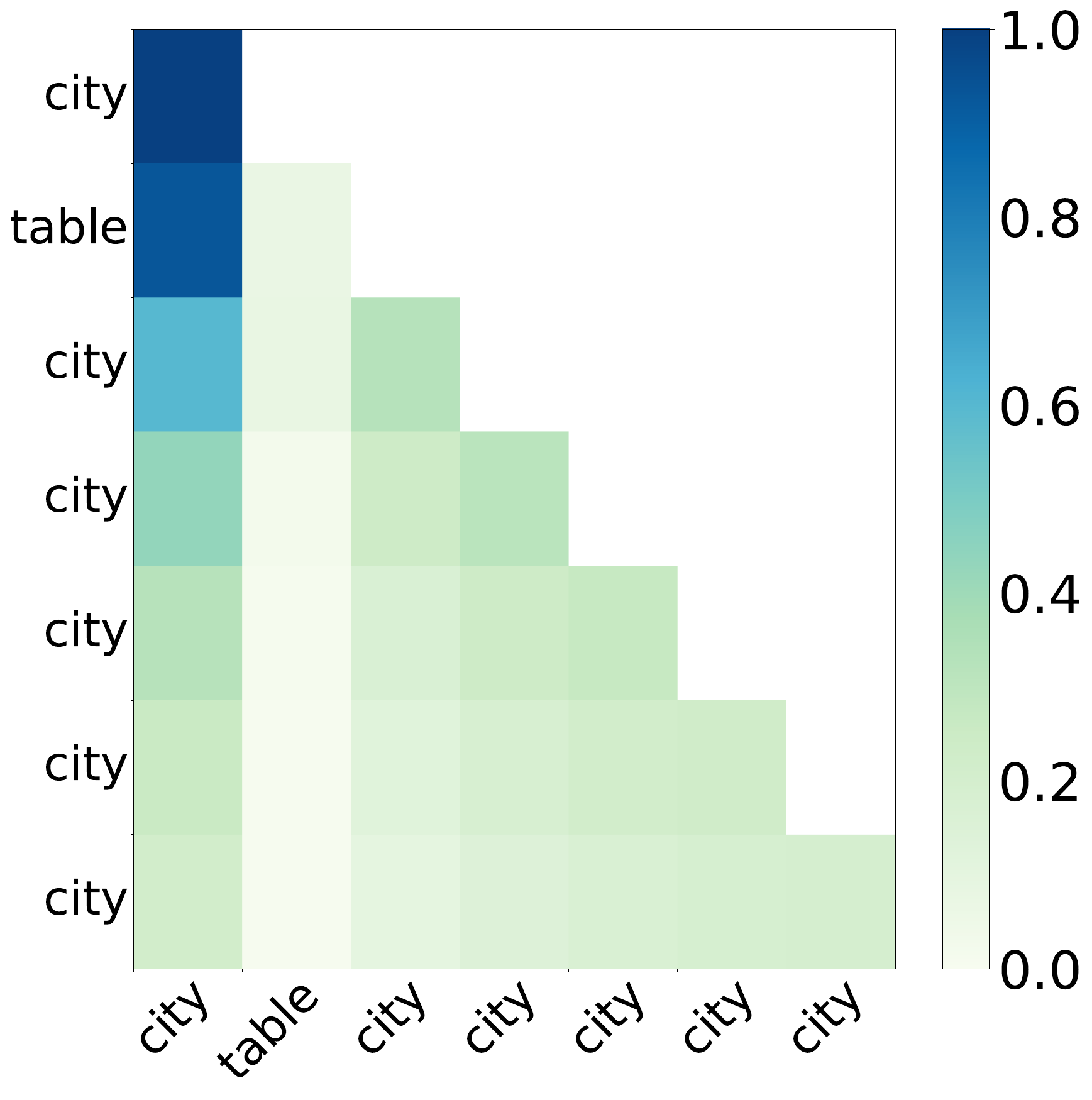}
    \caption{Heatmap: Distinct token at pos. 2}
    \label{fig:heatmap_diff_second_pythia_s2}
  \end{subfigure}
  \begin{subfigure}[t]{0.48\linewidth}
      \centering
      \includegraphics[width=\linewidth]{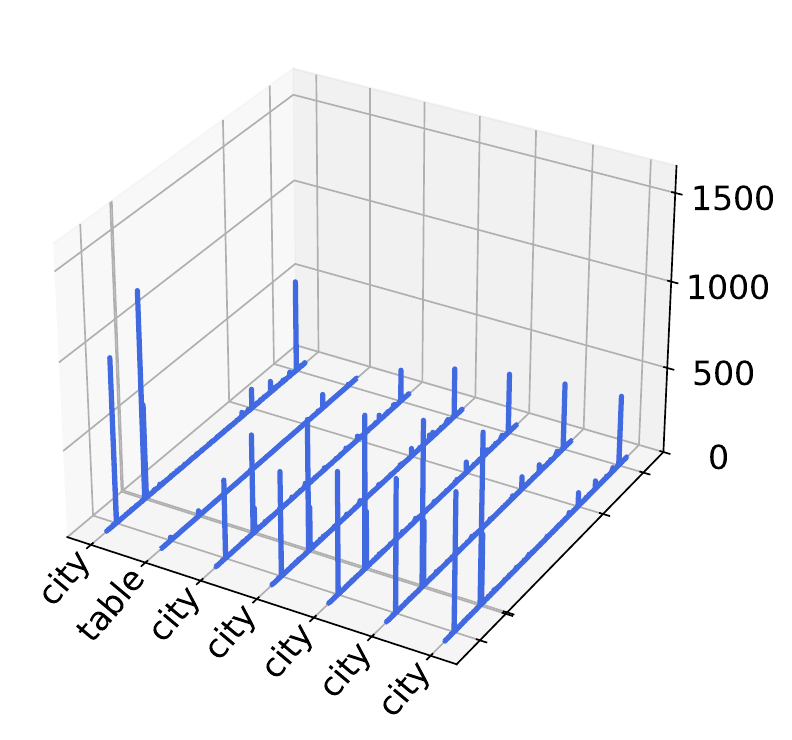}
      \caption{Activation: Distinct token at pos. 2}
      \label{fig:act_diff_second_pythia_s2}
  \end{subfigure}

  \begin{subfigure}[t]{0.48\linewidth}
      \centering
      \includegraphics[width=\linewidth]{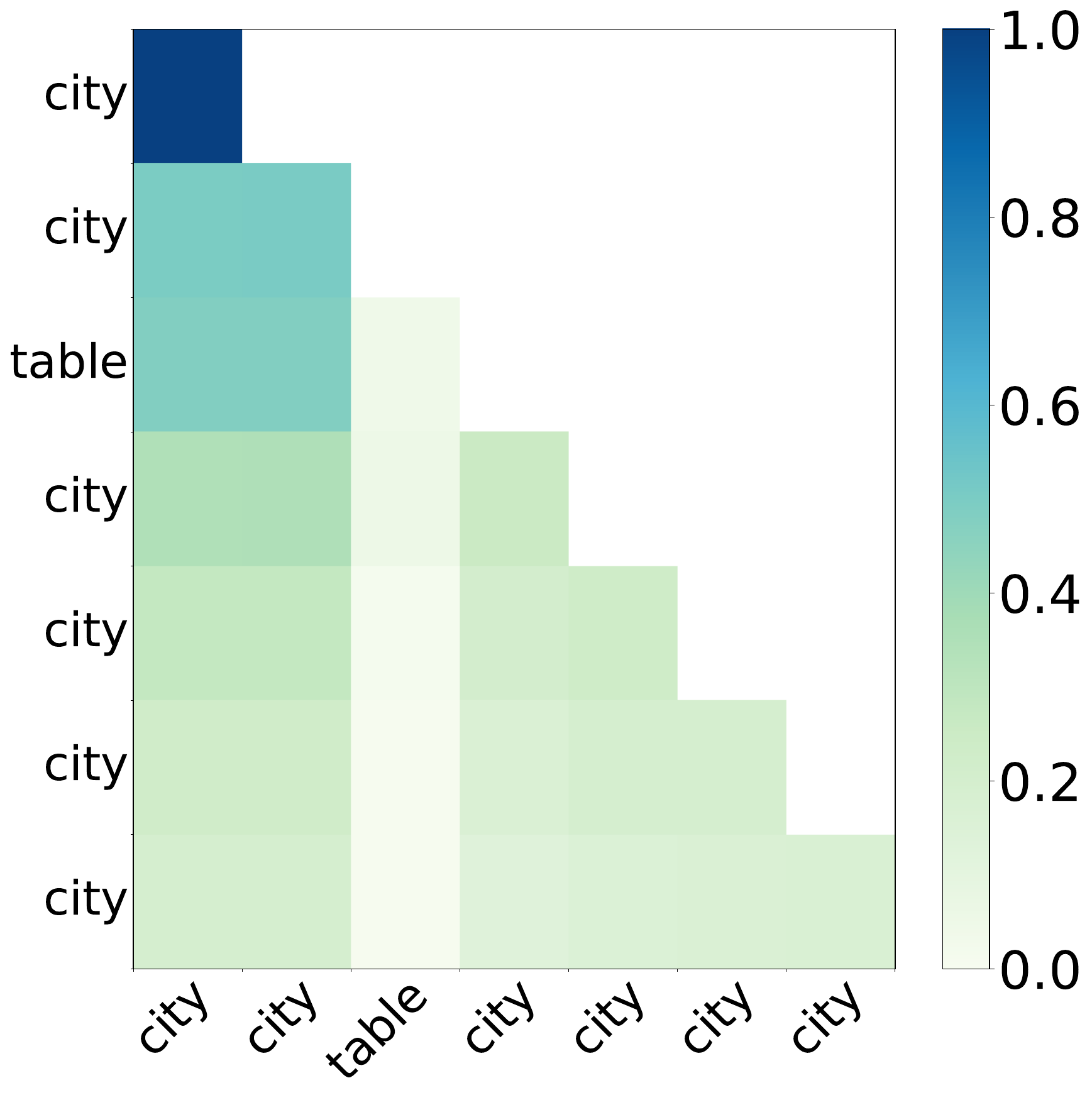}
      \caption{Heatmap: Distinct token at pos. 3}
      \label{fig:heatmap_diff_third_pythia_s2}
  \end{subfigure}
  \begin{subfigure}[t]{0.48\linewidth}
      \centering
      \includegraphics[width=\linewidth]{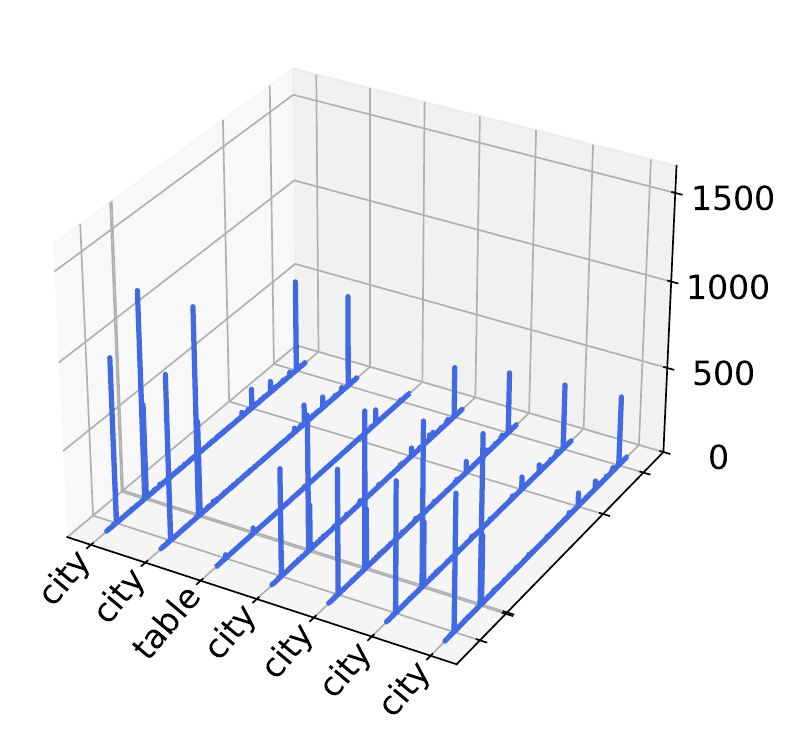}
      \caption{Activation: Distinct token at pos. 3}
      \label{fig:act_diff_third_pythia_s2}
  \end{subfigure}

  \caption{Results for sequences with repeated tokens.
  Each row compares the attention heatmap (left) and hidden-state activations (right) under the same token distribution pattern for pythia-1b (Sample 2).}
  \label{fig:repeated_tokens_results_pythia_s2}
\end{figure}

\begin{figure}[t]
  \centering
  \begin{subfigure}[t]{0.48\linewidth}
    \centering
    \includegraphics[width=\linewidth]{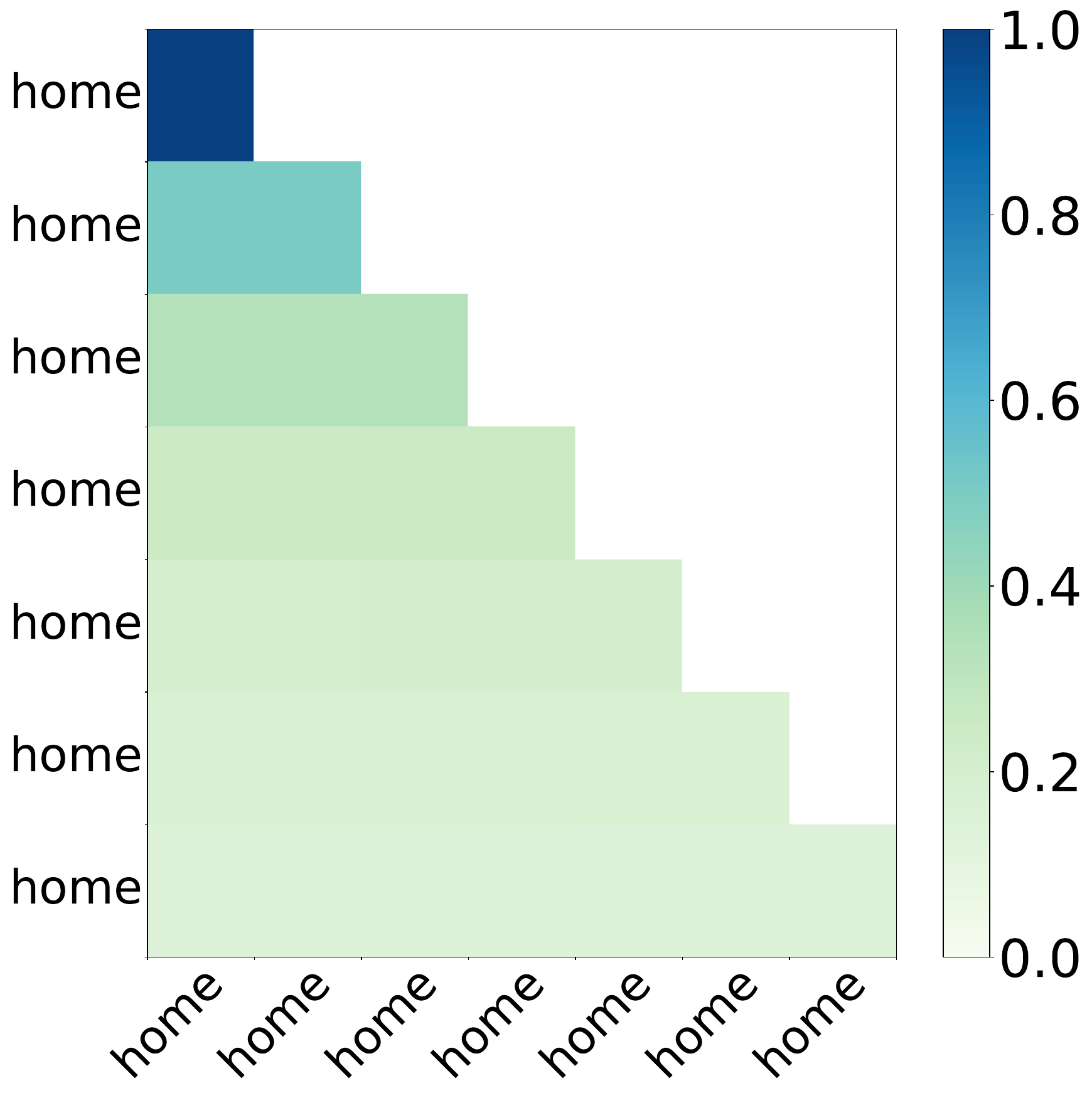}
    \caption{Heatmap: Uniform sequence}
    \label{fig:heatmap_all_same_pythia_s1}
  \end{subfigure}
  \begin{subfigure}[t]{0.48\linewidth}
      \centering
      \includegraphics[width=\linewidth]{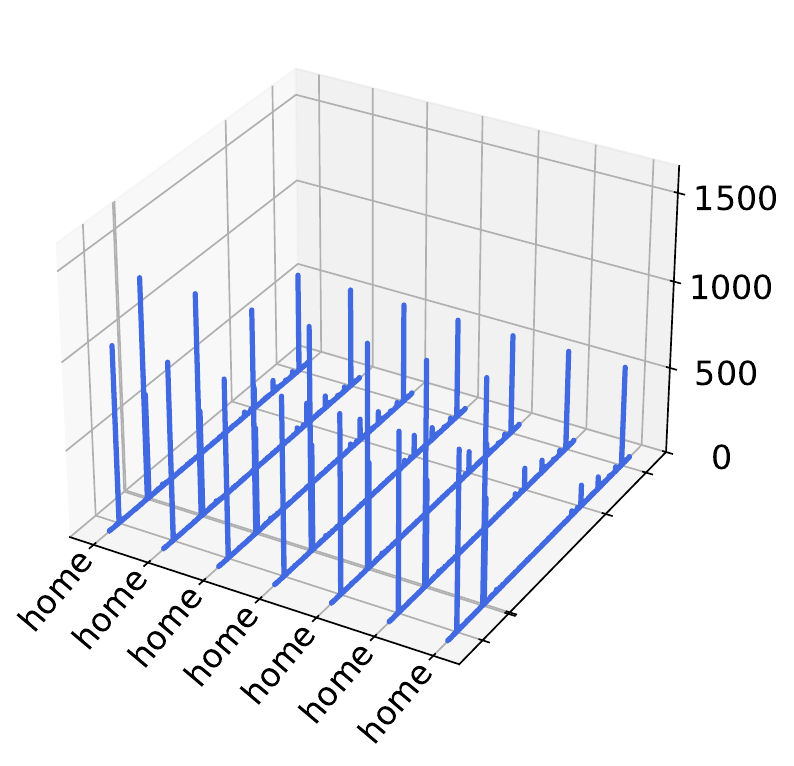}
      \caption{Activation: Uniform sequence}
      \label{fig:act_all_same_pythia_s1}
  \end{subfigure}

  \begin{subfigure}[t]{0.48\linewidth}
    \centering
    \includegraphics[width=\linewidth]{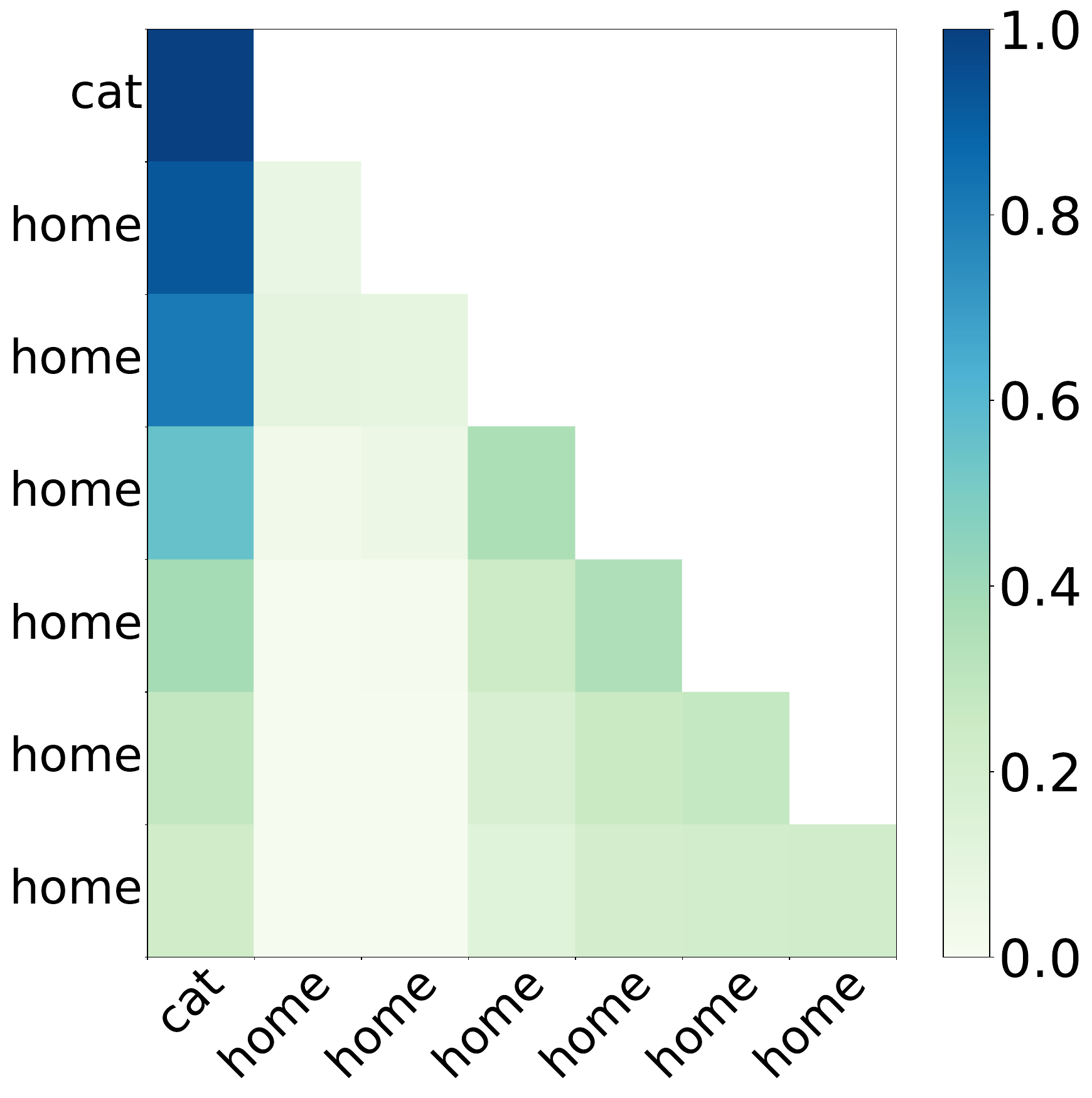}
    \caption{Heatmap: Distinct token at pos. 1}
    \label{fig:heatmap_diff_first_pythia_s1}
  \end{subfigure}
  \begin{subfigure}[t]{0.48\linewidth}
      \centering
      \includegraphics[width=\linewidth]{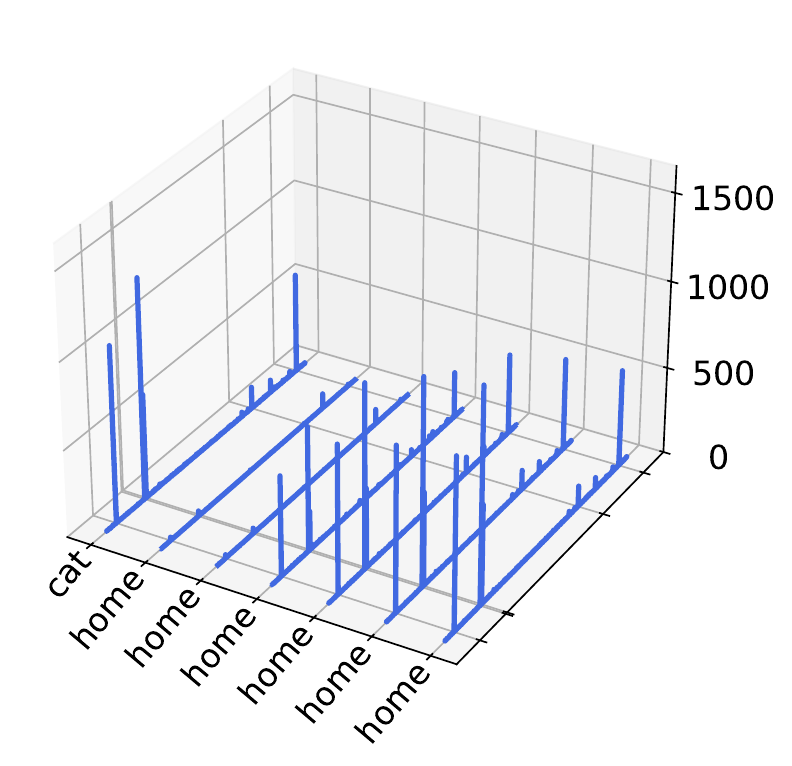}
      \caption{Activation: Distinct token at pos. 1}
      \label{fig:act_diff_first_pythia_s1}
  \end{subfigure}

  \begin{subfigure}[t]{0.48\linewidth}
    \centering
    \includegraphics[width=\linewidth]{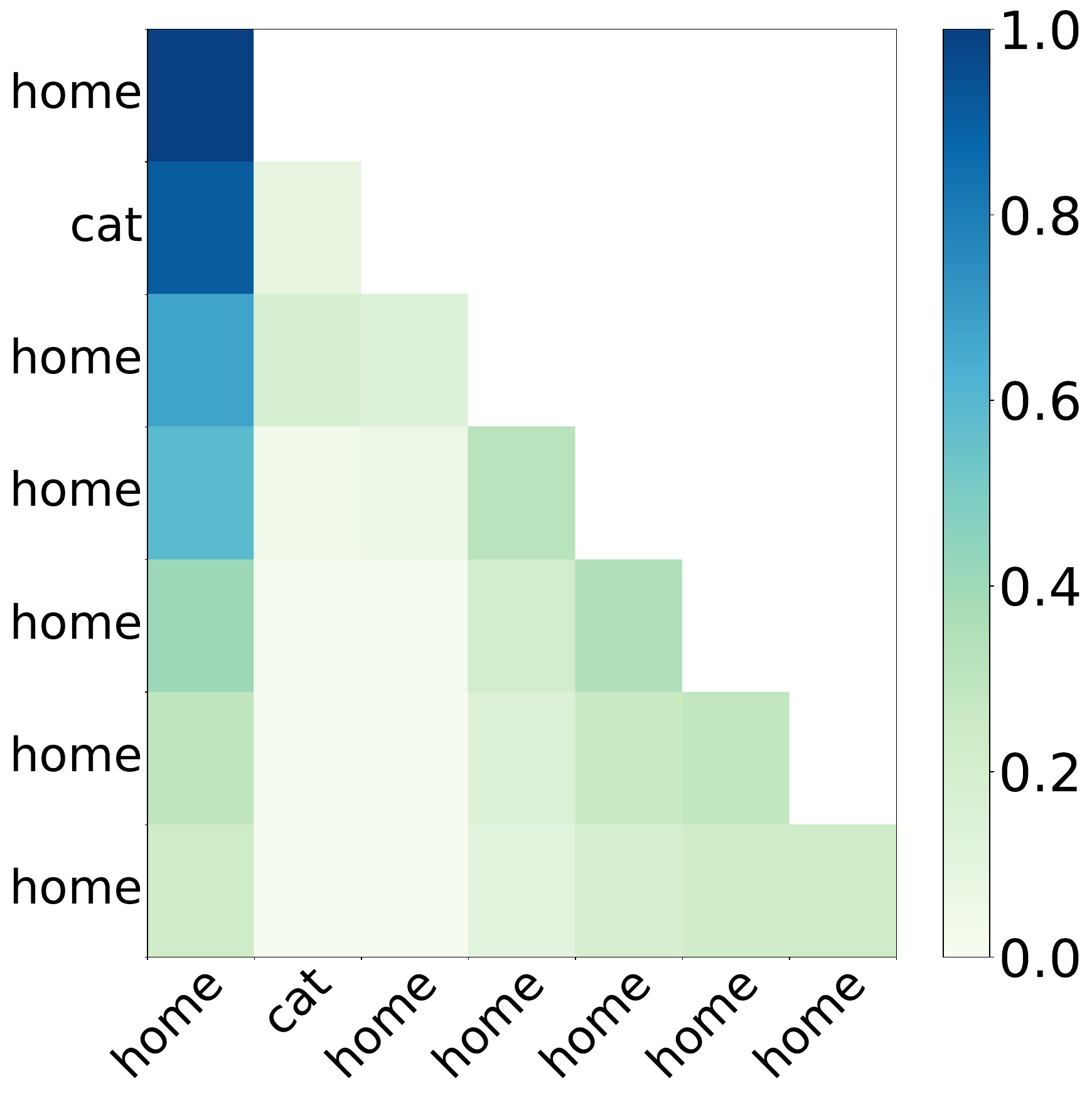}
    \caption{Heatmap: Distinct token at pos. 2}
    \label{fig:heatmap_diff_second_pythia_s1}
  \end{subfigure}
  \begin{subfigure}[t]{0.48\linewidth}
      \centering
      \includegraphics[width=\linewidth]{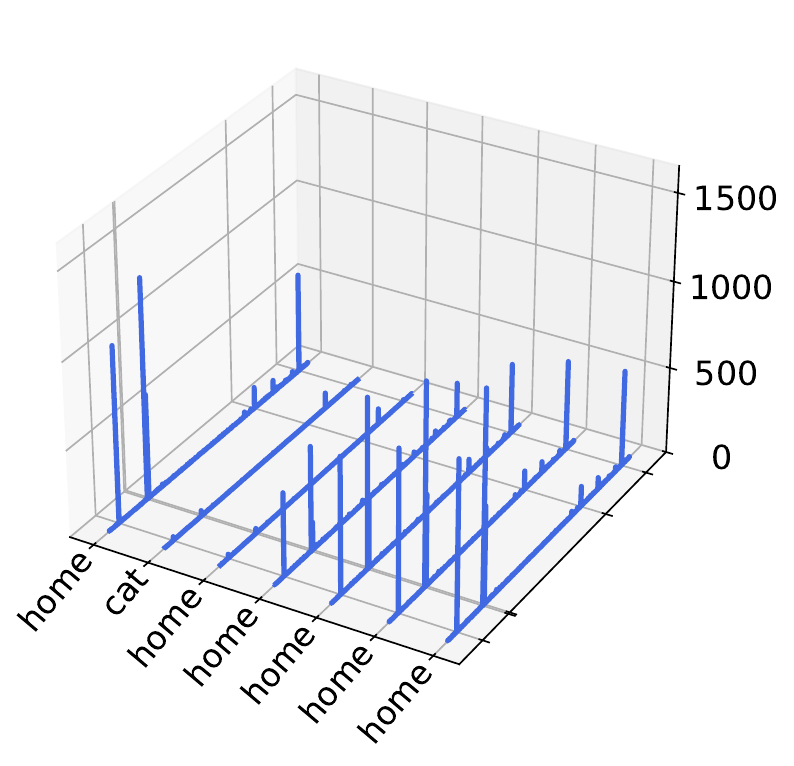}
      \caption{Activation: Distinct token at pos. 2}
      \label{fig:act_diff_second_pythia_s1}
  \end{subfigure}

  \begin{subfigure}[t]{0.48\linewidth}
      \centering
      \includegraphics[width=\linewidth]{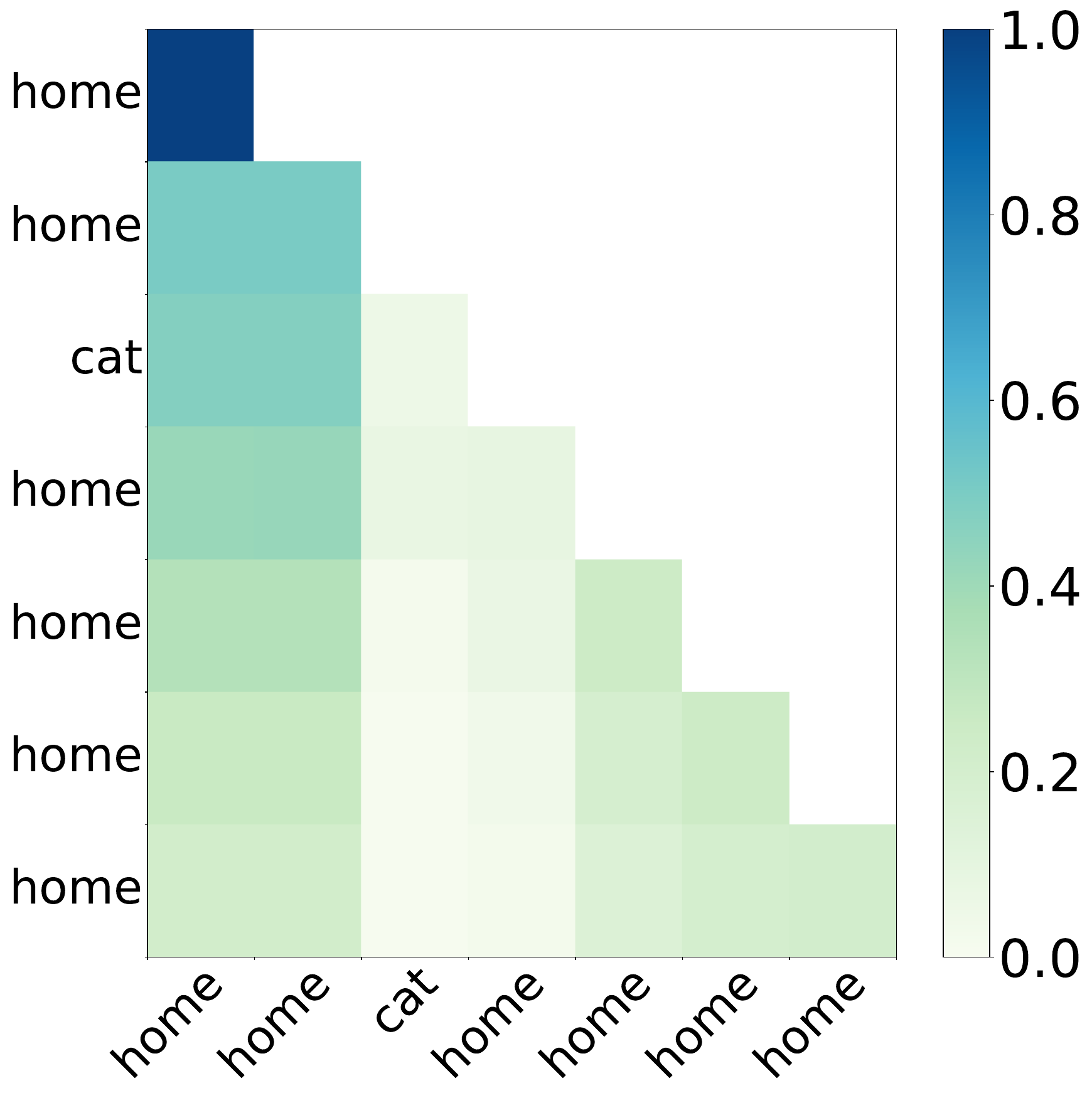}
      \caption{Heatmap: Distinct token at pos. 3}
      \label{fig:heatmap_diff_third_pythia_s1}
  \end{subfigure}
  \begin{subfigure}[t]{0.48\linewidth}
      \centering
      \includegraphics[width=\linewidth]{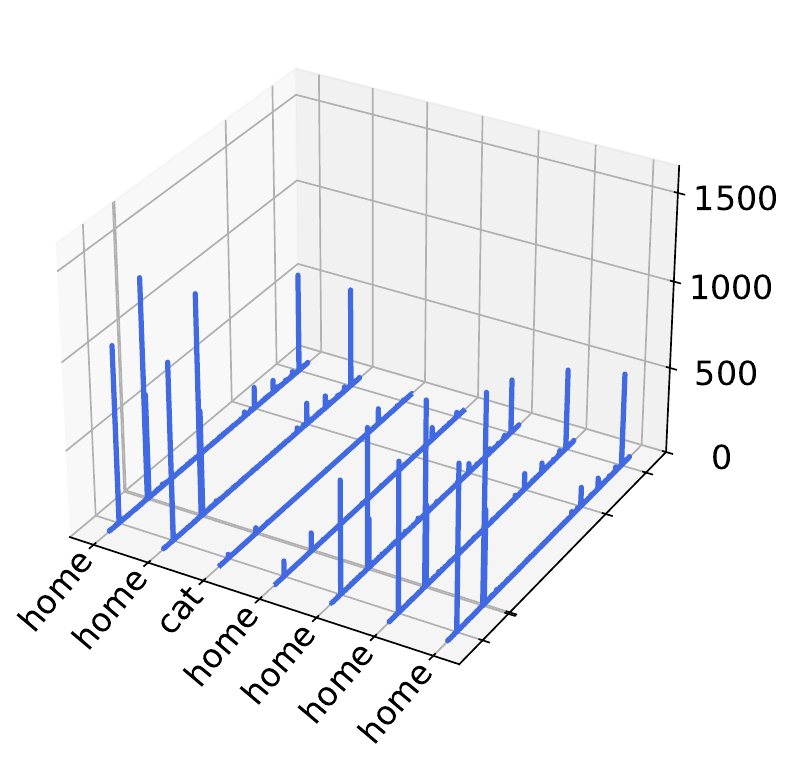}
      \caption{Activation: Distinct token at pos. 3}
      \label{fig:act_diff_third_pythia_s1}
  \end{subfigure}

  \caption{Results for sequences with repeated tokens.
  Each row compares the attention heatmap (left) and hidden-state activations (right) under the same token distribution pattern for pythia-1b (Sample 1).}
  \label{fig:repeated_tokens_results_pythia_s1}
\end{figure}

\begin{figure*}[t]
  \centering
  \begin{subfigure}[t]{0.48\linewidth}
    \centering
    \includegraphics[width=\linewidth]{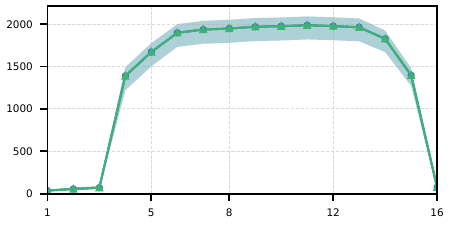}
    \caption{Norms: Uniform}
  \end{subfigure}
  \hfill
  \begin{subfigure}[t]{0.48\linewidth}
    \centering
    \includegraphics[width=\linewidth]{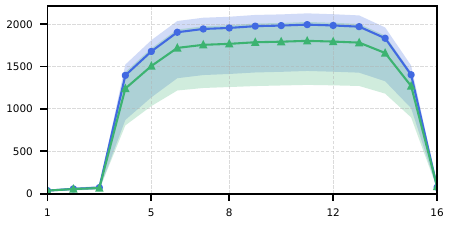}
    \caption{Norms: Pos. 1}
  \end{subfigure}

  \begin{subfigure}[t]{0.48\linewidth}
    \centering
    \includegraphics[width=\linewidth]{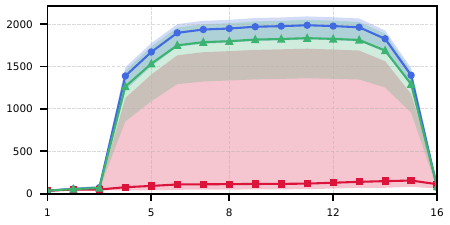}
    \caption{Norms: Pos. 2}
  \end{subfigure}
  \hfill
  \begin{subfigure}[t]{0.48\linewidth}
    \centering
    \includegraphics[width=\linewidth]{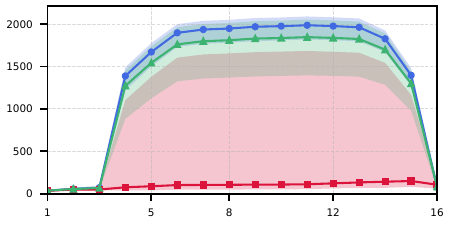}
    \caption{Norms: Pos. 3}
  \end{subfigure}
  \caption{Layer-wise averaged hidden-state norms in pythia-1b for repeated-token sequences under different distinct-token positions (colors and markers as in \Cref{fig:activation_norms_repeated_tokens_llama3}).
  The horizontal axis shows the layer index in every panel.}
  \label{fig:activation_norms_repeated_tokens_pythia}
\end{figure*}

\Crefrange{fig:repeated_tokens_results_llama2_s2}{fig:repeated_tokens_results_pythia_s1} below and \Cref{fig:repeated_tokens_results_llama3} in the main text show a clear trend supporting the ``Value-non-mixing'' hypothesis in Llama-3.2-3B, Llama-2-7b-hf, and Qwen2-7B, and the same trend in a weaker form in Mistral-7B-v0.3.
In pythia-1b, by contrast, \sinkRate{1} stays near zero for the first token, yet the attention pattern is not featureless.
The distinct token and the following token, whose representations mix into a distinct value, receive visibly less attention than the repeated tokens, in the direction ``Value-non-mixing'' predicts.
As in \Cref{fig:repeated_tokens_results_llama3}, these example figures use short sequences ($T=7$) with a fixed pair of tokens so that the per-token axis labels stay readable.
Sample 1 repeats ``home'' with ``cat'' as the distinct token, and Sample 2 repeats ``city'' with ``table''.
\Cref{fig:activation_norms_repeated_tokens_llama2,fig:activation_norms_repeated_tokens_mistral,fig:activation_norms_repeated_tokens_qwen,fig:activation_norms_repeated_tokens_pythia} show the corresponding layer-wise hidden-state norm trajectories for these models.

\Cref{tab:repeat_pattern_comparison_sink_metrics} summarizes \sinkRate{1} at the first token position across the four repeated-token patterns.
Uniform sequences (Pattern 0) yield \sinkRate[0.3]{1} of 0.000 for all five models, whereas inserting a single distinct token (Pattern 1--3) raises it to 0.68--0.80 for Llama-2-7b-hf, 0.69--0.80 for Llama-3.2-3B, and 0.42--0.55 for Qwen2-7B.
Mistral-7B-v0.3 (0.07--0.16) shows the same direction more weakly.
For pythia-1b the effect is far smaller: \sinkRate[0.3]{1} is 0.006 for Pattern 1 and rounds to 0.000 for Patterns 2--3.
The model still moves in the direction ``Value-non-mixing'' predicts.
A looser threshold of $\epsilon=0.2$ separates Pattern 1 (0.030) from Pattern 0 (0.000), and Patterns 1 and 2 remain above Pattern 3, as in the other four models.
This effect is nonetheless much smaller than in Llama-2-7b-hf or Llama-3.2-3B, and \sinkRate{1} stays at or below 0.03 at every threshold we test.

\input{tables/table_repeat_pattern_comparison}

\section{Additional Ablations under Varied Configurations}
\label{appendix:varied-configurations}

To assess the robustness of our findings, we additionally evaluate the self-concentration intervention of \Cref{sec:exp-self100} under varied experimental configurations: different natural-language datasets and input sequence lengths, different repeat lengths for the repeated-token setting introduced in \Cref{appendix:repeated-token-across-model}, restricting the intervention to subsets of layers, applying the intervention at additional intervened positions, and varying the threshold $\epsilon$ used to define \sinkRate{j}.

\subsection{Varying Dataset and Input Sequence Length}
\Cref{tab:sensitivity_analysis} reports \sinkRate{16} at the intervened token position ($t=16$) under the self-concentration intervention ($\alpha_{16,16}=1.0$, all layers, w/o BOS) for all five models across WikiText, GSM8K, and SlimPajama, and across input sequence lengths of 32, 64, 128, and (for WikiText) 256 tokens.
At $\epsilon=0.3$, the intervention yields a nonzero \sinkRate{16} for every model across all datasets and lengths, although the magnitude varies substantially across models and settings, particularly for Mistral-7B-v0.3.

\input{tables/table_sensitivity_analysis}

\begin{figure*}[t]
  \centering
  \begin{subfigure}[t]{0.32\linewidth}
    \centering
    \includegraphics[width=\linewidth]{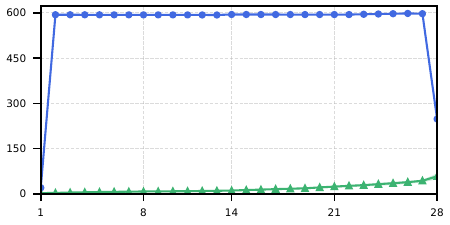}
    \caption{Norms: WikiText}
  \end{subfigure}
  \hfill
  \begin{subfigure}[t]{0.32\linewidth}
    \centering
    \includegraphics[width=\linewidth]{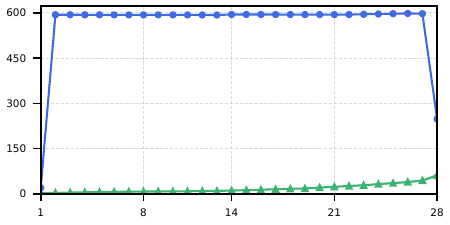}
    \caption{Norms: GSM8K}
  \end{subfigure}
  \hfill
  \begin{subfigure}[t]{0.32\linewidth}
    \centering
    \includegraphics[width=\linewidth]{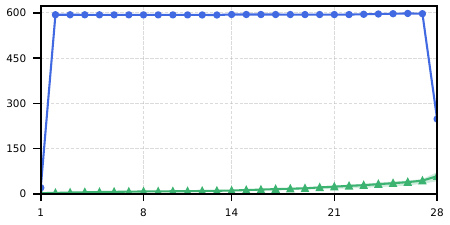}
    \caption{Norms: SlimPajama}
  \end{subfigure}
  \caption{Layer-wise averaged hidden-state norms in Llama-3.2-3B for an input length of 64 tokens across WikiText, GSM8K, and SlimPajama (baseline, no intervention).
  Blue circles and green triangles denote the first token and the mean over all other input tokens, respectively.
  The horizontal axis shows the layer index in every panel.
  Shaded bands show the sample min--max range.}
  \label{fig:activation_norms_sensitivity_llama3}
\end{figure*}

As a complementary baseline (no-intervention) check, \Cref{fig:activation_norms_sensitivity_llama3} shows layer-wise hidden-state norms for Llama-3.2-3B at an input length of 64 tokens across the three datasets, confirming that the large first-token norms reported in the main text are not specific to WikiText.

\subsection{Varying Repeat Length for Repeated-Token Sequences}
\label{appendix:repeat-length-varied}
\Cref{tab:repeat_input_sensitivity_analysis_part1,tab:repeat_input_sensitivity_analysis_part2,tab:repeat_input_sensitivity_analysis_part3} extend the repeated-token analysis of \Cref{appendix:repeated-token-across-model} to repeat lengths of 16, 32, 64, 128, and 256 tokens, confirming that the gap between the Uniform and Distinct-token patterns persists at every repeat length.
The only exception occurs at the shortest repeat length (16) with the lowest threshold ($\epsilon=0.2$), where even uniform sequences yield high \sinkRate{1} values.
At shorter lengths, baseline attention weights are inherently higher, so normal fluctuations easily exceed the fixed absolute threshold (see Limitations).

\input{tables/table_repeat_input_sensitivity}

\begin{figure*}[t]
  \centering
  \begin{subfigure}[t]{0.48\linewidth}
    \centering
    \includegraphics[width=\linewidth]{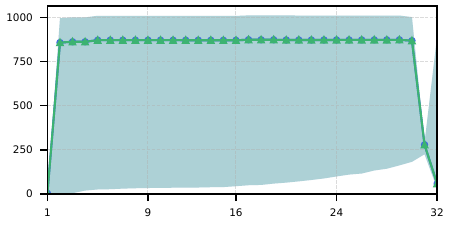}
    \caption{Norms: Uniform}
  \end{subfigure}
  \hfill
  \begin{subfigure}[t]{0.48\linewidth}
    \centering
    \includegraphics[width=\linewidth]{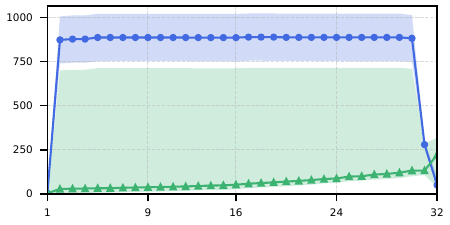}
    \caption{Norms: Pos. 1}
  \end{subfigure}

  \begin{subfigure}[t]{0.48\linewidth}
    \centering
    \includegraphics[width=\linewidth]{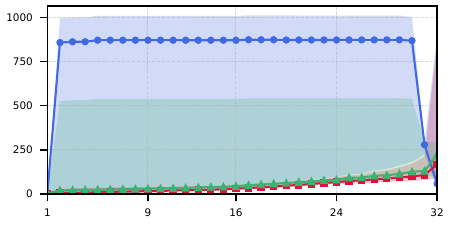}
    \caption{Norms: Pos. 2}
  \end{subfigure}
  \hfill
  \begin{subfigure}[t]{0.48\linewidth}
    \centering
    \includegraphics[width=\linewidth]{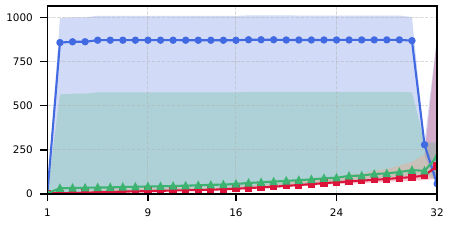}
    \caption{Norms: Pos. 3}
  \end{subfigure}
  \caption{Layer-wise averaged hidden-state norms in Llama-2-7b-hf for repeated-token sequences of length 128 under different distinct-token positions (colors and markers as in \Cref{fig:activation_norms_repeated_tokens_llama2}).
  The horizontal axis shows the layer index in every panel.}
  \label{fig:activation_norms_repeat_len128_llama2}
\end{figure*}

\Cref{fig:activation_norms_repeat_len128_llama2} shows the corresponding layer-wise hidden-state norms for Llama-2-7b-hf at a repeat length of 128 tokens, mirroring the trend already reported for length 64 in \Cref{fig:activation_norms_repeated_tokens_llama2}.

\subsection{Restricting the Self-Concentration Intervention to Layer Subsets}
\label{appendix:layer-restricted-intervention}
The self-concentration intervention in \Cref{sec:exp-self100} enforces $\alpha_{t,t}=1.0$ across all layers.
To identify which layers drive the resulting \as{}, we restrict the same intervention (at $t=16$, $T=64$, without a BOS token) to three-layer subsets: early (layers 1--3), middle (the three layers around the network midpoint), and late (the three layers preceding the final layer).
\Cref{tab:layer_specific_intervention_sink_metrics} reports \sinkRate{16} at the intervened position for each setting.
Intervening only on the early layers consistently yields \sinkRate[0.3]{16} values reaching 90--93\% of the all-layer intervention across all five models, whereas intervening on the middle or late layers leaves \sinkRate{16} near zero or at its vanilla level in all five models.
Intervening on layer 1 alone nearly reproduces the all-layer effect in Mistral-7B-v0.3 and pythia-1b, yields a partial increase in Llama-3.2-3B, and has minimal impact on Llama-2-7b-hf and Qwen2-7B, indicating that which early layers are responsible varies across models.
These results are consistent with the early-layer changes observed in the hidden-state norm trajectories.

\input{tables/table_layer_specific_intervention}

\subsection{Varying the Intervened Position}
\label{appendix:varying-intervened-position}
The self-concentration intervention in \Cref{sec:exp-self100} forces $\alpha_{t,t}=1.0$ at a single position, $t=16$.
To test whether the resulting increase in \as{} extends to other positions, we repeat the same intervention (all layers, w/o BOS, $T=64$) at $t=8$ and $t=32$.
\Cref{tab:intervention_position_sweep_sink_metrics} reports \sinkRate{t} at the intervened position for both settings.
The intervention increases \sinkRate{t} over the vanilla condition for all five models at both $t=8$ and $t=32$, showing that the observed increase is not unique to $t=16$ among the positions tested.
The increase is markedly smaller for Mistral-7B-v0.3, rising from 0.0660 to 0.1347 at $t=8$ and from 0.0000 to 0.0866 at $t=32$, several times below the intervened values of the other four models (0.4654--0.8542, \Cref{tab:intervention_position_sweep_sink_metrics}).

\input{tables/table_intervention_position_sweep}

\subsection{Varying Epsilon Thresholds}
\label{appendix:varying-epsilon-thresholds}
To verify that our main findings are not overly sensitive to the choice of the threshold $\epsilon$, we present the comparison of \sinkRate{j} across various thresholds ($\epsilon \in \{0.2, 0.3, 0.4, 0.5\}$) for the RoPE intervention experiment (\Cref{tab:sink_metrics_all_epsilons_standard}) and the self-concentration intervention experiment (\Cref{tab:sink_metrics_all_epsilons_interv}).
The overall trends remain consistent regardless of the specific threshold used.

\input{tables/table_set2_all_eps}
\input{tables/table_set4_intervention_all_eps}

%% file: tables/table_repeat_pattern_comparison.tex
\begin{table*}[t]
\centering
\small
\begin{tabular}{l l cccc}
\toprule
\textbf{Model} & \textbf{Pattern} & \textbf{$\epsilon=0.2$} & \textbf{$\epsilon=0.3$} & \textbf{$\epsilon=0.4$} & \textbf{$\epsilon=0.5$} \\
\midrule
\multirow{4}{*}{Llama-2-7b-hf} & Uniform (Pattern 0) & 0.0000 & 0.0000 & 0.0000 & 0.0000 \\
 & Distinct (Pattern 1) & 0.8454 & 0.7754 & 0.6898 & 0.6146 \\
 & Distinct (Pattern 2) & 0.8629 & 0.7955 & 0.7248 & 0.6529 \\
 & Distinct (Pattern 3) & 0.7934 & 0.6829 & 0.5166 & 0.0843 \\
\midrule
\multirow{4}{*}{Llama-3.2-3B} & Uniform (Pattern 0) & 0.0000 & 0.0000 & 0.0000 & 0.0000 \\
 & Distinct (Pattern 1) & 0.8197 & 0.7740 & 0.7177 & 0.6457 \\
 & Distinct (Pattern 2) & 0.8464 & 0.8042 & 0.7439 & 0.6665 \\
 & Distinct (Pattern 3) & 0.8039 & 0.6876 & 0.4526 & 0.0182 \\
\midrule
\multirow{4}{*}{Mistral-7B-v0.3} & Uniform (Pattern 0) & 0.0000 & 0.0000 & 0.0000 & 0.0000 \\
 & Distinct (Pattern 1) & 0.2344 & 0.1548 & 0.1104 & 0.0811 \\
 & Distinct (Pattern 2) & 0.2413 & 0.1568 & 0.1083 & 0.0784 \\
 & Distinct (Pattern 3) & 0.1435 & 0.0737 & 0.0311 & 0.0004 \\
\midrule
\multirow{4}{*}{Qwen2-7B} & Uniform (Pattern 0) & 0.0000 & 0.0000 & 0.0000 & 0.0000 \\
 & Distinct (Pattern 1) & 0.6475 & 0.5482 & 0.4688 & 0.3941 \\
 & Distinct (Pattern 2) & 0.6450 & 0.5404 & 0.4586 & 0.3803 \\
 & Distinct (Pattern 3) & 0.5829 & 0.4179 & 0.2264 & 0.0171 \\
\midrule
\multirow{4}{*}{pythia-1b} & Uniform (Pattern 0) & 0.0000 & 0.0000 & 0.0000 & 0.0000 \\
 & Distinct (Pattern 1) & 0.0302 & 0.0055 & 0.0035 & 0.0024 \\
 & Distinct (Pattern 2) & 0.0111 & 0.0000 & 0.0000 & 0.0000 \\
 & Distinct (Pattern 3) & 0.0001 & 0.0000 & 0.0000 & 0.0000 \\
\bottomrule
\end{tabular}
\caption{Comparison of \sinkRate{1} at the first token position between the uniform sequence (Pattern 0) and sequences with a single distinct token (Pattern 1--3) for repeated-token sequences with $T=64$.}
\label{tab:repeat_pattern_comparison_sink_metrics}
\end{table*}

%% file: tables/table_sensitivity_analysis.tex
\begin{table*}[t]
\centering
\small
\begin{tabular}{l l l c c c c}
\toprule
\textbf{Dataset} & \textbf{$T$} & \textbf{Model} & \textbf{$\epsilon=0.2$} & \textbf{$\epsilon=0.3$} & \textbf{$\epsilon=0.4$} & \textbf{$\epsilon=0.5$} \\
\midrule
\multirow{20}{*}{WikiText} & \multirow{5}{*}{32} & Llama-2-7b-hf & 0.9317 & 0.8932 & 0.6865 & 0.1487 \\
 &  & Llama-3.2-3B & 0.9329 & 0.8389 & 0.5787 & 0.1651 \\
 &  & Mistral-7B-v0.3 & 0.3077 & 0.1770 & 0.1109 & 0.0386 \\
 &  & Qwen2-7B & 0.8199 & 0.6346 & 0.3563 & 0.0735 \\
 &  & pythia-1b & 0.7342 & 0.6005 & 0.3301 & 0.0544 \\ \cmidrule(lr){2-7}
 & \multirow{5}{*}{64} & Llama-2-7b-hf & 0.9248 & 0.8397 & 0.5308 & 0.0095 \\
 &  & Llama-3.2-3B & 0.9121 & 0.7424 & 0.4527 & 0.0120 \\
 &  & Mistral-7B-v0.3 & 0.1897 & 0.1509 & 0.0868 & 0.0304 \\
 &  & Qwen2-7B & 0.7533 & 0.5085 & 0.2174 & 0.0182 \\
 &  & pythia-1b & 0.6975 & 0.4703 & 0.1703 & 0.0054 \\ \cmidrule(lr){2-7}
 & \multirow{5}{*}{128} & Llama-2-7b-hf & 0.9176 & 0.7870 & 0.4259 & 0.0013 \\
 &  & Llama-3.2-3B & 0.8888 & 0.6888 & 0.3911 & 0.0024 \\
 &  & Mistral-7B-v0.3 & 0.1830 & 0.1414 & 0.0762 & 0.0284 \\
 &  & Qwen2-7B & 0.7122 & 0.4401 & 0.1464 & 0.0113 \\
 &  & pythia-1b & 0.6729 & 0.3746 & 0.1013 & 0.0005 \\ \cmidrule(lr){2-7}
 & \multirow{5}{*}{256} & Llama-2-7b-hf & 0.9125 & 0.7543 & 0.3730 & 0.0009 \\
 &  & Llama-3.2-3B & 0.8767 & 0.6696 & 0.3717 & 0.0017 \\
 &  & Mistral-7B-v0.3 & 0.1810 & 0.1382 & 0.0726 & 0.0275 \\
 &  & Qwen2-7B & 0.6952 & 0.4119 & 0.1256 & 0.0102 \\
 &  & pythia-1b & 0.6570 & 0.3329 & 0.0805 & 0.0005 \\
\midrule
\multirow{15}{*}{GSM8K} & \multirow{5}{*}{32} & Llama-2-7b-hf & 0.9196 & 0.7548 & 0.2460 & 0.0349 \\
 &  & Llama-3.2-3B & 0.9216 & 0.7736 & 0.4897 & 0.1090 \\
 &  & Mistral-7B-v0.3 & 0.3357 & 0.2053 & 0.1008 & 0.0051 \\
 &  & Qwen2-7B & 0.8138 & 0.6294 & 0.3222 & 0.0519 \\
 &  & pythia-1b & 0.7115 & 0.4763 & 0.1953 & 0.0103 \\ \cmidrule(lr){2-7}
 & \multirow{5}{*}{64} & Llama-2-7b-hf & 0.8880 & 0.5561 & 0.0577 & 0.0002 \\
 &  & Llama-3.2-3B & 0.8922 & 0.6769 & 0.3905 & 0.0207 \\
 &  & Mistral-7B-v0.3 & 0.2479 & 0.1705 & 0.0624 & 0.0009 \\
 &  & Qwen2-7B & 0.7682 & 0.5200 & 0.1943 & 0.0206 \\
 &  & pythia-1b & 0.6488 & 0.3407 & 0.0941 & 0.0000 \\ \cmidrule(lr){2-7}
 & \multirow{5}{*}{128} & Llama-2-7b-hf & 0.8789 & 0.5132 & 0.0438 & 0.0002 \\
 &  & Llama-3.2-3B & 0.8871 & 0.6656 & 0.3800 & 0.0186 \\
 &  & Mistral-7B-v0.3 & 0.2462 & 0.1659 & 0.0563 & 0.0006 \\
 &  & Qwen2-7B & 0.7632 & 0.5053 & 0.1809 & 0.0186 \\
 &  & pythia-1b & 0.6416 & 0.3180 & 0.0811 & 0.0000 \\
\midrule
\multirow{15}{*}{SlimPajama} & \multirow{5}{*}{32} & Llama-2-7b-hf & 0.9291 & 0.8330 & 0.3042 & 0.0575 \\
 &  & Llama-3.2-3B & 0.9328 & 0.8576 & 0.6092 & 0.1773 \\
 &  & Mistral-7B-v0.3 & 0.3432 & 0.2173 & 0.1108 & 0.0249 \\
 &  & Qwen2-7B & 0.8243 & 0.6595 & 0.3841 & 0.0821 \\
 &  & pythia-1b & 0.7344 & 0.5374 & 0.2180 & 0.0462 \\ \cmidrule(lr){2-7}
 & \multirow{5}{*}{64} & Llama-2-7b-hf & 0.9144 & 0.6415 & 0.0995 & 0.0002 \\
 &  & Llama-3.2-3B & 0.9161 & 0.7680 & 0.4761 & 0.0143 \\
 &  & Mistral-7B-v0.3 & 0.2329 & 0.1644 & 0.0649 & 0.0153 \\
 &  & Qwen2-7B & 0.7640 & 0.5403 & 0.2449 & 0.0185 \\
 &  & pythia-1b & 0.6631 & 0.3365 & 0.0972 & 0.0038 \\ \cmidrule(lr){2-7}
 & \multirow{5}{*}{128} & Llama-2-7b-hf & 0.8908 & 0.4642 & 0.0178 & 0.0000 \\
 &  & Llama-3.2-3B & 0.8935 & 0.7009 & 0.4078 & 0.0024 \\
 &  & Mistral-7B-v0.3 & 0.2126 & 0.1419 & 0.0429 & 0.0134 \\
 &  & Qwen2-7B & 0.7221 & 0.4605 & 0.1671 & 0.0105 \\
 &  & pythia-1b & 0.5809 & 0.2037 & 0.0381 & 0.0004 \\
\bottomrule
\end{tabular}
\caption{\sinkRate{16} at the intervened token position ($t=16$) under the self-concentration intervention ($\alpha_{16,16}=1.0$, all layers, w/o BOS) across natural-language datasets and input sequence lengths $T$.}
\label{tab:sensitivity_analysis}
\end{table*}

%% file: tables/table_repeat_input_sensitivity.tex
\begin{table*}[t]
\centering
\footnotesize
\begin{tabular}{l l l c c c c}
\toprule
\textbf{Repeat Length} & \textbf{Model} & \textbf{Pattern} & \textbf{$\epsilon=0.2$} & \textbf{$\epsilon=0.3$} & \textbf{$\epsilon=0.4$} & \textbf{$\epsilon=0.5$} \\
\midrule
\multirow{20}{*}{16} & \multirow{4}{*}{Llama-2-7b-hf} & Uniform (Pattern 0) & 0.9593 & 0.0001 & 0.0000 & 0.0000 \\
 &  & Distinct (Pattern 1) & 0.9551 & 0.8931 & 0.8340 & 0.7284 \\
 &  & Distinct (Pattern 2) & 0.9469 & 0.8996 & 0.8479 & 0.7545 \\
 &  & Distinct (Pattern 3) & 0.9419 & 0.8283 & 0.6307 & 0.3001 \\ \cmidrule(lr){2-7}
 & \multirow{4}{*}{Llama-3.2-3B} & Uniform (Pattern 0) & 0.9426 & 0.0035 & 0.0001 & 0.0000 \\
 &  & Distinct (Pattern 1) & 0.9328 & 0.8489 & 0.7883 & 0.7330 \\
 &  & Distinct (Pattern 2) & 0.9353 & 0.8634 & 0.8224 & 0.7545 \\
 &  & Distinct (Pattern 3) & 0.9205 & 0.8111 & 0.6421 & 0.1864 \\ \cmidrule(lr){2-7}
 & \multirow{4}{*}{Mistral-7B-v0.3} & Uniform (Pattern 0) & 0.7456 & 0.0000 & 0.0000 & 0.0000 \\
 &  & Distinct (Pattern 1) & 0.9019 & 0.6684 & 0.3816 & 0.2671 \\
 &  & Distinct (Pattern 2) & 0.9348 & 0.7320 & 0.4111 & 0.2808 \\
 &  & Distinct (Pattern 3) & 0.9101 & 0.3667 & 0.1499 & 0.0177 \\ \cmidrule(lr){2-7}
 & \multirow{4}{*}{Qwen2-7B} & Uniform (Pattern 0) & 0.9087 & 0.0063 & 0.0025 & 0.0006 \\
 &  & Distinct (Pattern 1) & 0.8847 & 0.7919 & 0.6725 & 0.5873 \\
 &  & Distinct (Pattern 2) & 0.8933 & 0.7882 & 0.6714 & 0.5754 \\
 &  & Distinct (Pattern 3) & 0.8490 & 0.6448 & 0.4252 & 0.1036 \\ \cmidrule(lr){2-7}
 & \multirow{4}{*}{pythia-1b} & Uniform (Pattern 0) & 0.7830 & 0.0067 & 0.0000 & 0.0000 \\
 &  & Distinct (Pattern 1) & 0.8373 & 0.6219 & 0.2639 & 0.1123 \\
 &  & Distinct (Pattern 2) & 0.8245 & 0.6010 & 0.2709 & 0.0927 \\
 &  & Distinct (Pattern 3) & 0.8059 & 0.2901 & 0.0246 & 0.0000 \\
\midrule
\multirow{20}{*}{32} & \multirow{4}{*}{Llama-2-7b-hf} & Uniform (Pattern 0) & 0.0000 & 0.0000 & 0.0000 & 0.0000 \\
 &  & Distinct (Pattern 1) & 0.8917 & 0.8243 & 0.7371 & 0.6524 \\
 &  & Distinct (Pattern 2) & 0.8923 & 0.8449 & 0.7622 & 0.6809 \\
 &  & Distinct (Pattern 3) & 0.8502 & 0.7205 & 0.5495 & 0.1633 \\ \cmidrule(lr){2-7}
 & \multirow{4}{*}{Llama-3.2-3B} & Uniform (Pattern 0) & 0.0007 & 0.0000 & 0.0000 & 0.0000 \\
 &  & Distinct (Pattern 1) & 0.8474 & 0.8017 & 0.7498 & 0.6868 \\
 &  & Distinct (Pattern 2) & 0.8625 & 0.8310 & 0.7784 & 0.7034 \\
 &  & Distinct (Pattern 3) & 0.8291 & 0.7405 & 0.5247 & 0.0630 \\ \cmidrule(lr){2-7}
 & \multirow{4}{*}{Mistral-7B-v0.3} & Uniform (Pattern 0) & 0.0000 & 0.0000 & 0.0000 & 0.0000 \\
 &  & Distinct (Pattern 1) & 0.5238 & 0.2821 & 0.1919 & 0.1446 \\
 &  & Distinct (Pattern 2) & 0.5833 & 0.2995 & 0.1927 & 0.1447 \\
 &  & Distinct (Pattern 3) & 0.3187 & 0.1443 & 0.0638 & 0.0017 \\ \cmidrule(lr){2-7}
 & \multirow{4}{*}{Qwen2-7B} & Uniform (Pattern 0) & 0.0028 & 0.0004 & 0.0000 & 0.0000 \\
 &  & Distinct (Pattern 1) & 0.7825 & 0.6552 & 0.5579 & 0.4807 \\
 &  & Distinct (Pattern 2) & 0.7781 & 0.6526 & 0.5556 & 0.4688 \\
 &  & Distinct (Pattern 3) & 0.6813 & 0.5158 & 0.3090 & 0.0410 \\ \cmidrule(lr){2-7}
 & \multirow{4}{*}{pythia-1b} & Uniform (Pattern 0) & 0.0013 & 0.0000 & 0.0000 & 0.0000 \\
 &  & Distinct (Pattern 1) & 0.4490 & 0.0870 & 0.0164 & 0.0055 \\
 &  & Distinct (Pattern 2) & 0.4200 & 0.0674 & 0.0045 & 0.0001 \\
 &  & Distinct (Pattern 3) & 0.1306 & 0.0002 & 0.0000 & 0.0000 \\
\bottomrule
\end{tabular}
\caption{\sinkRate{1} for repeated-token sequences across repeat lengths, without intervention (Part 1: Length 16, 32).}
\label{tab:repeat_input_sensitivity_analysis_part1}
\end{table*}

\begin{table*}[t]
\centering
\footnotesize
\begin{tabular}{l l l c c c c}
\toprule
\textbf{Repeat Length} & \textbf{Model} & \textbf{Pattern} & \textbf{$\epsilon=0.2$} & \textbf{$\epsilon=0.3$} & \textbf{$\epsilon=0.4$} & \textbf{$\epsilon=0.5$} \\
\midrule
\multirow{20}{*}{64} & \multirow{4}{*}{Llama-2-7b-hf} & Uniform (Pattern 0) & 0.0000 & 0.0000 & 0.0000 & 0.0000 \\
 &  & Distinct (Pattern 1) & 0.8454 & 0.7754 & 0.6898 & 0.6146 \\
 &  & Distinct (Pattern 2) & 0.8629 & 0.7955 & 0.7248 & 0.6529 \\
 &  & Distinct (Pattern 3) & 0.7934 & 0.6829 & 0.5166 & 0.0843 \\ \cmidrule(lr){2-7}
 & \multirow{4}{*}{Llama-3.2-3B} & Uniform (Pattern 0) & 0.0000 & 0.0000 & 0.0000 & 0.0000 \\
 &  & Distinct (Pattern 1) & 0.8197 & 0.7740 & 0.7177 & 0.6457 \\
 &  & Distinct (Pattern 2) & 0.8464 & 0.8042 & 0.7439 & 0.6665 \\
 &  & Distinct (Pattern 3) & 0.8039 & 0.6876 & 0.4526 & 0.0182 \\ \cmidrule(lr){2-7}
 & \multirow{4}{*}{Mistral-7B-v0.3} & Uniform (Pattern 0) & 0.0000 & 0.0000 & 0.0000 & 0.0000 \\
 &  & Distinct (Pattern 1) & 0.2344 & 0.1548 & 0.1104 & 0.0811 \\
 &  & Distinct (Pattern 2) & 0.2413 & 0.1568 & 0.1083 & 0.0784 \\
 &  & Distinct (Pattern 3) & 0.1435 & 0.0737 & 0.0311 & 0.0004 \\ \cmidrule(lr){2-7}
 & \multirow{4}{*}{Qwen2-7B} & Uniform (Pattern 0) & 0.0000 & 0.0000 & 0.0000 & 0.0000 \\
 &  & Distinct (Pattern 1) & 0.6475 & 0.5482 & 0.4688 & 0.3941 \\
 &  & Distinct (Pattern 2) & 0.6450 & 0.5404 & 0.4586 & 0.3803 \\
 &  & Distinct (Pattern 3) & 0.5829 & 0.4179 & 0.2264 & 0.0171 \\ \cmidrule(lr){2-7}
 & \multirow{4}{*}{pythia-1b} & Uniform (Pattern 0) & 0.0000 & 0.0000 & 0.0000 & 0.0000 \\
 &  & Distinct (Pattern 1) & 0.0302 & 0.0055 & 0.0035 & 0.0024 \\
 &  & Distinct (Pattern 2) & 0.0111 & 0.0000 & 0.0000 & 0.0000 \\
 &  & Distinct (Pattern 3) & 0.0001 & 0.0000 & 0.0000 & 0.0000 \\
\midrule
\multirow{20}{*}{128} & \multirow{4}{*}{Llama-2-7b-hf} & Uniform (Pattern 0) & 0.0000 & 0.0000 & 0.0000 & 0.0000 \\
 &  & Distinct (Pattern 1) & 0.8125 & 0.7484 & 0.6691 & 0.6022 \\
 &  & Distinct (Pattern 2) & 0.8289 & 0.7752 & 0.7064 & 0.6411 \\
 &  & Distinct (Pattern 3) & 0.7708 & 0.6646 & 0.5060 & 0.0408 \\ \cmidrule(lr){2-7}
 & \multirow{4}{*}{Llama-3.2-3B} & Uniform (Pattern 0) & 0.0000 & 0.0000 & 0.0000 & 0.0000 \\
 &  & Distinct (Pattern 1) & 0.7987 & 0.7472 & 0.6803 & 0.6042 \\
 &  & Distinct (Pattern 2) & 0.8304 & 0.7848 & 0.7171 & 0.6371 \\
 &  & Distinct (Pattern 3) & 0.7766 & 0.6452 & 0.4095 & 0.0032 \\ \cmidrule(lr){2-7}
 & \multirow{4}{*}{Mistral-7B-v0.3} & Uniform (Pattern 0) & 0.0000 & 0.0000 & 0.0000 & 0.0000 \\
 &  & Distinct (Pattern 1) & 0.1295 & 0.0828 & 0.0584 & 0.0410 \\
 &  & Distinct (Pattern 2) & 0.1294 & 0.0815 & 0.0529 & 0.0326 \\
 &  & Distinct (Pattern 3) & 0.0721 & 0.0322 & 0.0129 & 0.0001 \\ \cmidrule(lr){2-7}
 & \multirow{4}{*}{Qwen2-7B} & Uniform (Pattern 0) & 0.0000 & 0.0000 & 0.0000 & 0.0000 \\
 &  & Distinct (Pattern 1) & 0.5524 & 0.4658 & 0.3957 & 0.3326 \\
 &  & Distinct (Pattern 2) & 0.5207 & 0.4278 & 0.3519 & 0.2928 \\
 &  & Distinct (Pattern 3) & 0.4756 & 0.3338 & 0.1788 & 0.0073 \\ \cmidrule(lr){2-7}
 & \multirow{4}{*}{pythia-1b} & Uniform (Pattern 0) & 0.0000 & 0.0000 & 0.0000 & 0.0000 \\
 &  & Distinct (Pattern 1) & 0.0070 & 0.0045 & 0.0023 & 0.0011 \\
 &  & Distinct (Pattern 2) & 0.0000 & 0.0000 & 0.0000 & 0.0000 \\
 &  & Distinct (Pattern 3) & 0.0000 & 0.0000 & 0.0000 & 0.0000 \\
\bottomrule
\end{tabular}
\caption{\sinkRate{1} for repeated-token sequences across repeat lengths, without intervention (Part 2: Length 64, 128).}
\label{tab:repeat_input_sensitivity_analysis_part2}
\end{table*}

\begin{table*}[t]
\centering
\footnotesize
\begin{tabular}{l l l c c c c}
\toprule
\textbf{Repeat Length} & \textbf{Model} & \textbf{Pattern} & \textbf{$\epsilon=0.2$} & \textbf{$\epsilon=0.3$} & \textbf{$\epsilon=0.4$} & \textbf{$\epsilon=0.5$} \\
\midrule
\multirow{20}{*}{256} & \multirow{4}{*}{Llama-2-7b-hf} & Uniform (Pattern 0) & 0.0000 & 0.0000 & 0.0000 & 0.0000 \\
 &  & Distinct (Pattern 1) & 0.7957 & 0.7306 & 0.6570 & 0.5928 \\
 &  & Distinct (Pattern 2) & 0.8138 & 0.7608 & 0.6949 & 0.6342 \\
 &  & Distinct (Pattern 3) & 0.7488 & 0.6524 & 0.5011 & 0.0130 \\ \cmidrule(lr){2-7}
 & \multirow{4}{*}{Llama-3.2-3B} & Uniform (Pattern 0) & 0.0000 & 0.0000 & 0.0000 & 0.0000 \\
 &  & Distinct (Pattern 1) & 0.7739 & 0.7121 & 0.6482 & 0.5734 \\
 &  & Distinct (Pattern 2) & 0.8101 & 0.7543 & 0.6868 & 0.6057 \\
 &  & Distinct (Pattern 3) & 0.7456 & 0.6120 & 0.3770 & 0.0002 \\ \cmidrule(lr){2-7}
 & \multirow{4}{*}{Mistral-7B-v0.3} & Uniform (Pattern 0) & 0.0000 & 0.0000 & 0.0000 & 0.0000 \\
 &  & Distinct (Pattern 1) & 0.0665 & 0.0385 & 0.0255 & 0.0184 \\
 &  & Distinct (Pattern 2) & 0.0616 & 0.0303 & 0.0209 & 0.0148 \\
 &  & Distinct (Pattern 3) & 0.0290 & 0.0152 & 0.0061 & 0.0000 \\ \cmidrule(lr){2-7}
 & \multirow{4}{*}{Qwen2-7B} & Uniform (Pattern 0) & 0.0000 & 0.0000 & 0.0000 & 0.0000 \\
 &  & Distinct (Pattern 1) & 0.4623 & 0.3806 & 0.3147 & 0.2558 \\
 &  & Distinct (Pattern 2) & 0.4389 & 0.3761 & 0.3215 & 0.2667 \\
 &  & Distinct (Pattern 3) & 0.4187 & 0.2900 & 0.1526 & 0.0030 \\ \cmidrule(lr){2-7}
 & \multirow{4}{*}{pythia-1b} & Uniform (Pattern 0) & 0.0000 & 0.0000 & 0.0000 & 0.0000 \\
 &  & Distinct (Pattern 1) & 0.0044 & 0.0019 & 0.0009 & 0.0004 \\
 &  & Distinct (Pattern 2) & 0.0000 & 0.0000 & 0.0000 & 0.0000 \\
 &  & Distinct (Pattern 3) & 0.0000 & 0.0000 & 0.0000 & 0.0000 \\
\bottomrule
\end{tabular}
\caption{\sinkRate{1} for repeated-token sequences across repeat lengths, without intervention (Part 3: Length 256).}
\label{tab:repeat_input_sensitivity_analysis_part3}
\end{table*}

%% file: tables/table_layer_specific_intervention.tex
\begin{table*}[t]
\centering
\small
\begin{tabular}{l c c c c c c }
\toprule
\textbf{Model} & \textbf{Vanilla} & \textbf{All layers} & \textbf{Early} & \textbf{Middle} & \textbf{Late} & \textbf{Layer 1} \\
\midrule
Llama-2-7b-hf & 0.0073 & 0.8397 & 0.7724 & 0.0074 & 0.0074 & 0.0241 \\
\midrule
Llama-3.2-3B & 0.0012 & 0.7424 & 0.6670 & 0.0013 & 0.0012 & 0.1310 \\
\midrule
Mistral-7B-v0.3 & 0.0373 & 0.1509 & 0.1396 & 0.0374 & 0.0374 & 0.1385 \\
\midrule
Qwen2-7B & 0.0000 & 0.5085 & 0.4737 & 0.0000 & 0.0000 & 0.0036 \\
\midrule
pythia-1b & 0.0000 & 0.4703 & 0.4296 & 0.0000 & 0.0000 & 0.4443 \\
\bottomrule
\end{tabular}
\caption{\sinkRate{16} at the intervened token position ($t=16$) when the self-concentration intervention ($\alpha_{t,t}=1.0$) is applied to all layers, restricted to the early, middle, or late three layers, or to layer 1 alone, with $T=64$ (all w/o BOS).}
\label{tab:layer_specific_intervention_sink_metrics}
\end{table*}

%% file: tables/table_intervention_position_sweep.tex
\begin{table}[t]
\centering
\small
\begin{tabular}{l l c c }
\toprule
\textbf{Model} & \textbf{Position} & \textbf{Vanilla} & \textbf{Interv} \\
\midrule
\multirow{2}{*}{Llama-2-7b-hf} & 8 & 0.0000 & \textbf{0.8542} \\
 & 32 & 0.0072 & \textbf{0.8217} \\
\midrule
\multirow{2}{*}{Llama-3.2-3B} & 8 & 0.0026 & \textbf{0.7560} \\
 & 32 & 0.0006 & \textbf{0.7406} \\
\midrule
\multirow{2}{*}{Mistral-7B-v0.3} & 8 & 0.0660 & \textbf{0.1347} \\
 & 32 & 0.0000 & \textbf{0.0866} \\
\midrule
\multirow{2}{*}{Qwen2-7B} & 8 & 0.0001 & \textbf{0.5324} \\
 & 32 & 0.0001 & \textbf{0.4946} \\
\midrule
\multirow{2}{*}{pythia-1b} & 8 & 0.0000 & \textbf{0.4885} \\
 & 32 & 0.0000 & \textbf{0.4654} \\
\bottomrule
\end{tabular}
\caption{\sinkRate{t} at the intervened token position for $t=8$ and $t=32$, comparing vanilla attention to the self-concentration intervention ($\alpha_{t,t}=1.0$, all layers, w/o BOS), with $T=64$.}
\label{tab:intervention_position_sweep_sink_metrics}
\end{table}

%% file: tables/table_set2_all_eps.tex
\begin{table*}[t]
\centering
\small
\begin{tabular}{lll cccc}
\toprule
\textbf{Model} & \textbf{Condition} & \textbf{Position} & \textbf{$\epsilon=0.2$} & \textbf{$\epsilon=0.3$} & \textbf{$\epsilon=0.4$} & \textbf{$\epsilon=0.5$} \\
\midrule
\multirow{5}{*}{Llama-2-7b-hf} & \multirow{4}{*}{Vanilla} & \textbf{1} & 0.9334 & 0.9305 & 0.9187 & 0.8819 \\
 &  & 15 & 0.0005 & 0.0001 & 0.0000 & 0.0000 \\
 &  & 16 & 0.0090 & 0.0073 & 0.0028 & 0.0001 \\
 &  & 17 & 0.0091 & 0.0071 & 0.0023 & 0.0000 \\ \cmidrule(lr){2-7}
 & \multirow{1}{*}{RoPE Interv} & \textbf{1} & 0.9341 & 0.9273 & 0.9100 & 0.8663 \\
\midrule
\multirow{5}{*}{Llama-3.2-3B} & \multirow{4}{*}{Vanilla} & \textbf{1} & 0.9329 & 0.9218 & 0.8933 & 0.8118 \\
 &  & 15 & 0.0011 & 0.0008 & 0.0006 & 0.0004 \\
 &  & 16 & 0.0014 & 0.0012 & 0.0007 & 0.0004 \\
 &  & 17 & 0.0004 & 0.0003 & 0.0003 & 0.0002 \\ \cmidrule(lr){2-7}
 & \multirow{1}{*}{RoPE Interv} & \textbf{1} & 0.9317 & 0.9133 & 0.8713 & 0.7749 \\
\midrule
\multirow{5}{*}{Mistral-7B-v0.3} & \multirow{4}{*}{Vanilla} & \textbf{1} & 0.2258 & 0.0343 & 0.0033 & 0.0001 \\
 &  & 15 & 0.0195 & 0.0187 & 0.0172 & 0.0143 \\
 &  & 16 & 0.0386 & 0.0373 & 0.0344 & 0.0285 \\
 &  & 17 & 0.0194 & 0.0187 & 0.0172 & 0.0146 \\ \cmidrule(lr){2-7}
 & \multirow{1}{*}{RoPE Interv} & \textbf{1} & 0.3655 & 0.1525 & 0.1129 & 0.0960 \\
\midrule
\multirow{5}{*}{Qwen2-7B} & \multirow{4}{*}{Vanilla} & \textbf{1} & 0.8677 & 0.8173 & 0.7439 & 0.6113 \\
 &  & 15 & 0.0005 & 0.0002 & 0.0000 & 0.0000 \\
 &  & 16 & 0.0004 & 0.0000 & 0.0000 & 0.0000 \\
 &  & 17 & 0.0004 & 0.0001 & 0.0000 & 0.0000 \\ \cmidrule(lr){2-7}
 & \multirow{1}{*}{RoPE Interv} & \textbf{1} & 0.8671 & 0.7931 & 0.6810 & 0.5290 \\
\midrule
\multirow{5}{*}{pythia-1b} & \multirow{4}{*}{Vanilla} & \textbf{1} & 0.7433 & 0.7337 & 0.6889 & 0.5659 \\
 &  & 15 & 0.0002 & 0.0000 & 0.0000 & 0.0000 \\
 &  & 16 & 0.0001 & 0.0000 & 0.0000 & 0.0000 \\
 &  & 17 & 0.0001 & 0.0000 & 0.0000 & 0.0000 \\ \cmidrule(lr){2-7}
 & \multirow{1}{*}{RoPE Interv} & \textbf{1} & 0.7955 & 0.7547 & 0.7056 & 0.5749 \\
\bottomrule
\end{tabular}
\caption{Comparison of \sinkRate{j} across various thresholds $\epsilon$ for RoPE ablation (both w/o BOS).}
\label{tab:sink_metrics_all_epsilons_standard}
\end{table*}

%% file: tables/table_set4_intervention_all_eps.tex
\begin{table*}[t]
\centering
\small
\begin{tabular}{lll cccc}
\toprule
\textbf{Model} & \textbf{Condition} & \textbf{Position} & \textbf{$\epsilon=0.2$} & \textbf{$\epsilon=0.3$} & \textbf{$\epsilon=0.4$} & \textbf{$\epsilon=0.5$} \\
\midrule
\multirow{8}{*}{Llama-2-7b-hf} & \multirow{4}{*}{Vanilla} & \textbf{1 (First)} & 0.9334 & 0.9305 & 0.9187 & 0.8819 \\
 &  & 15 & 0.0005 & 0.0001 & 0.0000 & 0.0000 \\
 &  & 16 (Intervened) & 0.0090 & 0.0073 & 0.0028 & 0.0001 \\
 &  & 17 & 0.0091 & 0.0071 & 0.0023 & 0.0000 \\ \cmidrule(lr){2-7}
 & \multirow{4}{*}{$\alpha_{16,16}=1.0$} & \textbf{1 (First)} & 0.9323 & 0.9164 & 0.8248 & 0.5356 \\
 &  & 15 & 0.0002 & 0.0000 & 0.0000 & 0.0000 \\
 &  & 16 (Intervened) & 0.9248 & 0.8397 & 0.5308 & 0.0095 \\
 &  & 17 & 0.0072 & 0.0006 & 0.0000 & 0.0000 \\
\midrule
\multirow{8}{*}{Llama-3.2-3B} & \multirow{4}{*}{Vanilla} & \textbf{1 (First)} & 0.9329 & 0.9218 & 0.8933 & 0.8118 \\
 &  & 15 & 0.0011 & 0.0008 & 0.0006 & 0.0004 \\
 &  & 16 (Intervened) & 0.0014 & 0.0012 & 0.0007 & 0.0004 \\
 &  & 17 & 0.0004 & 0.0003 & 0.0003 & 0.0002 \\ \cmidrule(lr){2-7}
 & \multirow{4}{*}{$\alpha_{16,16}=1.0$} & \textbf{1 (First)} & 0.9291 & 0.8785 & 0.6972 & 0.4172 \\
 &  & 15 & 0.0011 & 0.0008 & 0.0006 & 0.0003 \\
 &  & 16 (Intervened) & 0.9121 & 0.7424 & 0.4527 & 0.0120 \\
 &  & 17 & 0.0004 & 0.0003 & 0.0003 & 0.0002 \\
\midrule
\multirow{8}{*}{Mistral-7B-v0.3} & \multirow{4}{*}{Vanilla} & \textbf{1 (First)} & 0.2258 & 0.0343 & 0.0033 & 0.0001 \\
 &  & 15 & 0.0195 & 0.0187 & 0.0172 & 0.0143 \\
 &  & 16 (Intervened) & 0.0386 & 0.0373 & 0.0344 & 0.0285 \\
 &  & 17 & 0.0194 & 0.0187 & 0.0172 & 0.0146 \\ \cmidrule(lr){2-7}
 & \multirow{4}{*}{$\alpha_{16,16}=1.0$} & \textbf{1 (First)} & 0.1662 & 0.0198 & 0.0003 & 0.0000 \\
 &  & 15 & 0.0195 & 0.0182 & 0.0152 & 0.0117 \\
 &  & 16 (Intervened) & 0.1897 & 0.1509 & 0.0868 & 0.0304 \\
 &  & 17 & 0.0194 & 0.0184 & 0.0161 & 0.0129 \\
\midrule
\multirow{8}{*}{Qwen2-7B} & \multirow{4}{*}{Vanilla} & \textbf{1 (First)} & 0.8677 & 0.8173 & 0.7439 & 0.6113 \\
 &  & 15 & 0.0005 & 0.0002 & 0.0000 & 0.0000 \\
 &  & 16 (Intervened) & 0.0004 & 0.0000 & 0.0000 & 0.0000 \\
 &  & 17 & 0.0004 & 0.0001 & 0.0000 & 0.0000 \\ \cmidrule(lr){2-7}
 & \multirow{4}{*}{$\alpha_{16,16}=1.0$} & \textbf{1 (First)} & 0.8485 & 0.7278 & 0.4770 & 0.2108 \\
 &  & 15 & 0.0004 & 0.0001 & 0.0000 & 0.0000 \\
 &  & 16 (Intervened) & 0.7533 & 0.5085 & 0.2174 & 0.0182 \\
 &  & 17 & 0.0005 & 0.0001 & 0.0000 & 0.0000 \\
\midrule
\multirow{8}{*}{pythia-1b} & \multirow{4}{*}{Vanilla} & \textbf{1 (First)} & 0.7433 & 0.7337 & 0.6889 & 0.5659 \\
 &  & 15 & 0.0002 & 0.0000 & 0.0000 & 0.0000 \\
 &  & 16 (Intervened) & 0.0001 & 0.0000 & 0.0000 & 0.0000 \\
 &  & 17 & 0.0001 & 0.0000 & 0.0000 & 0.0000 \\ \cmidrule(lr){2-7}
 & \multirow{4}{*}{$\alpha_{16,16}=1.0$} & \textbf{1 (First)} & 0.7392 & 0.6702 & 0.4391 & 0.1769 \\
 &  & 15 & 0.0001 & 0.0000 & 0.0000 & 0.0000 \\
 &  & 16 (Intervened) & 0.6975 & 0.4703 & 0.1703 & 0.0054 \\
 &  & 17 & 0.0001 & 0.0000 & 0.0000 & 0.0000 \\
\bottomrule
\end{tabular}
\caption{Comparison of \sinkRate{j} across various thresholds $\epsilon$ when intervention forces self-concentration of attention at position 16 (both w/o BOS).}
\label{tab:sink_metrics_all_epsilons_interv}
\end{table*}